\documentclass[pdflatex,sn-mathphys-num]{sn-jnl}

\usepackage{graphicx}%
\usepackage{multirow}%
\usepackage{amsmath,amssymb,amsfonts}%
\usepackage{amsthm}%
\usepackage{mathrsfs}%
\usepackage[title]{appendix}%
\usepackage{xcolor}%
\usepackage{textcomp}%
\usepackage{manyfoot}%
\usepackage{booktabs}%
\usepackage{algorithm}%
\usepackage{algorithmicx}%
\usepackage{algpseudocode}%
\usepackage{listings}%
\usepackage{mathtools}
\usepackage{tabularx}
\usepackage{makecell} 
\usepackage{ragged2e}
\usepackage{multirow}
\usepackage{xcolor}
\usepackage{subcaption}

\theoremstyle{thmstyleone}%
\theoremstyle{thmstyletwo}%

\theoremstyle{thmstylethree}%

\begin{document}

\title[Article Title]{Toward Seamless Physical Human-Humanoid Interaction: Insights from Control, Intent, and Modeling with a Vision for What Comes Next}


\author{\fnm{Gustavo A.} \sur{Cardona}}\email{cardona@udel.edu}

\author{\fnm{Shubham S.} \sur{Kumbhar}}\email{shubhamk@udel.edu}

\author*{\fnm{Panagiotis} \sur{Artemiadis}}\email{partem@udel.udel.edu}

\affil{\orgdiv{Department of Mechanical Engineering}, \orgname{University of Delaware}, \orgaddress{\street{210 S College Ave}, \city{Newark}, \postcode{19716}, \state{DE}, \country{USA}}}




\abstract{
Physical Human-Humanoid Interaction (pHHI) is a rapidly advancing field with significant implications for deploying robots in unstructured, human-centric environments. 
In this review, we examine the current state of the art in pHHI through three core pillars: (i) humanoid modeling and control, (ii) human intent estimation, and (iii) computational human models. 
For each pillar, we survey representative approaches, identify open challenges, and analyze current limitations that hinder robust, scalable, and adaptive interaction. 
These include the need for whole-body control strategies capable of handling uncertain human dynamics, real-time intent inference under limited sensing, and modeling techniques that account for variability in human physical states.  
Although significant progress has been made within each domain, integration across pillars remains limited. We propose pathways for unifying methods across these areas to enable cohesive interaction frameworks. 
This structure enables us not only to map the current landscape but also to propose concrete directions for future research that aim to bridge these domains.
Additionally, we introduce a unified taxonomy of interaction types based on modality, distinguishing between direct interactions (e.g., physical contact) and indirect interactions (e.g., object-mediated), and on the level of robot engagement, ranging from assistance to cooperation and collaboration. 
For each category in this taxonomy, we provide the three core pillars that highlight opportunities for cross-pillar unification. 
Our goal is to suggest avenues to advance robust, safe, and intuitive physical interaction, providing a roadmap for future research that will allow humanoid systems to effectively understand, anticipate, and collaborate with human partners in diverse real-world settings.
}




\maketitle

\section{Introduction}
\label{sec:introduction}

In recent years, advances in artificial intelligence and the decreasing cost of technological components have led to an unprecedented surge in the development and deployment of humanoid robots. 
Market forecasts predict remarkable growth, with some estimates suggesting that billions of humanoid robots could be deployed around the world by 2040~\cite{neumann2025humanoid}.
Their anthropomorphic form and dexterous movements make them especially well-suited to navigate, manipulate, and interact within environments designed for people, which motivates an urgent push to expand their capabilities and versatility. 
Cutting-edge platforms, such as Honda's ASIMO~\cite{sakagami2002intelligent}, Boston Dynamics' Atlas~\cite{guizzo2019leaps}, Unitree's G1~\cite{unitreeG1}, and Agility Robotics' Digit~\cite{agilityroboticsHomepage}, with the help of multiple researchers from industry and academia, have shown the promise of these robots for performing locomotion tasks.
Experts foresee them moving beyond the limited scope of past industrial applications and becoming a fundamental part of daily life.
In areas such as healthcare~\cite{mukherjee2022humanoid}, humanoid robots can assist with tasks including lifting and transporting patients, guiding rehabilitation exercises, and serving as agile assistants in clinics and hospitals, thereby helping to reduce the workload of medical personnel.
In domestic and service roles~\cite{mcginn2014towards}, humanoids could perform tasks of cleaning, fetching, and education, while also providing essential support to the elderly and people with disabilities. 
Furthermore, in dangerous and inaccessible environments, such as disaster zones~\cite{settimi2014modular} or outer space~\cite{diftler2011robonaut}, humanoid robots could carry out critical missions, keeping humans safe from harm, among many other applications.

From factory floors to disaster relief efforts, from hospital wards to home living rooms, humanoid robots are on the brink of becoming indispensable partners, reshaping how we live, work, and interact with others~\cite{tong2024advancements}.
These diverse applications highlight that the technical challenge is not just achieving isolated competence in walking~\cite{mikolajczyk2022recent}, balancing~\cite{zhang2025review}, or grasping~\cite{laschi2000grasping}, but enabling safe, intuitive, and effective \emph{physical Human–Humanoid Interaction} (pHHI), which is the primary focus of this review. 

PHHI refers to the bidirectional exchange of forces, motion, and information between a human and a humanoid robot through intentional physical contact to achieve a shared task. 
In contrast to robot-alone tasks, pHHI embeds the human as a dynamic, uncertain, and co-equal element of the control loop. 
This introduces complexities beyond those already present in humanoid control for locomotion, manipulation, or balance, as human responses to robot actions, as well as subtle variations in contact, timing, or posture, can substantially influence task outcomes.

Unlike purely virtual or verbal interaction, pHHI requires continuous physical coupling in which both partners adapt their behaviors in response to real-time variations in force, posture, and intent. 
These couplings, also referred to as \emph{interaction modalities}~\cite{farajtabar2024path}, can be broadly classified into two types. 
In \emph{direct interaction}, such as body-to-body contact (e.g., assisting a person to stand~\cite{lefevre2024humanoid}), the humanoid must exhibit whole-body compliance, reliable contact detection, and human-safe impedance control. 
In \emph{indirect, object-mediated interaction}, such as co-manipulation~\cite{agravante2019human} or teleoperation via a haptic interface~\cite{darvish2023teleoperation}, effective coordination depends on shared object models, force/torque coupling, and accurate task state estimation. 
In both cases, the physical coupling serves not only as a means to accomplish the task but also as a communication channel: the human and the humanoid both convey their intents through applied forces, while adapting to reach a common consensus. 
Any mismatch in intent reduces efficiency and can compromise the stability of the coupled system. 
Consequently, humanoids must explicitly integrate safety mechanisms such as compliance modulation and impedance control, as their actions directly affect human safety during interaction.
Therefore, in pHHI these couplings inherently combine dynamic stability, compliance, intent inference, and safety assurance into a unified control problem. 

The existence of multiple autonomous entities often puts these control problems under the constraints of \emph{shared autonomy} and human-centered ergonomics. 
The concept of shared autonomy~\cite{losey2018review} introduces an additional layer of autonomy that spans a spectrum from pure teleoperation to fully proactive assistance, with cooperative modes in which roles, timing, and force-sharing change dynamically as the task phases and human state evolve. 
Some researchers also refer to this as arbitration.
This perspective emphasizes that effective pHHI requires not only robust physical coupling and control but also adaptive role allocation, ensuring both efficiency and safety in human–robot partnerships.

These defining aspects of pHHI distinguish it from conventional humanoid control, making it far more than a straightforward application of existing methods. 
pHHI requires dedicated consideration of the coupled human–robot system. 
The classical kinematic, dynamic, and trajectory-centric paradigms must be extended to account for human biomechanics, latent intent, comfort, social expectations, and safety attributes, which are noisy, time-varying, and person-specific.

\begin{figure}[t]
    \centering
    \includegraphics[width=0.7\linewidth]{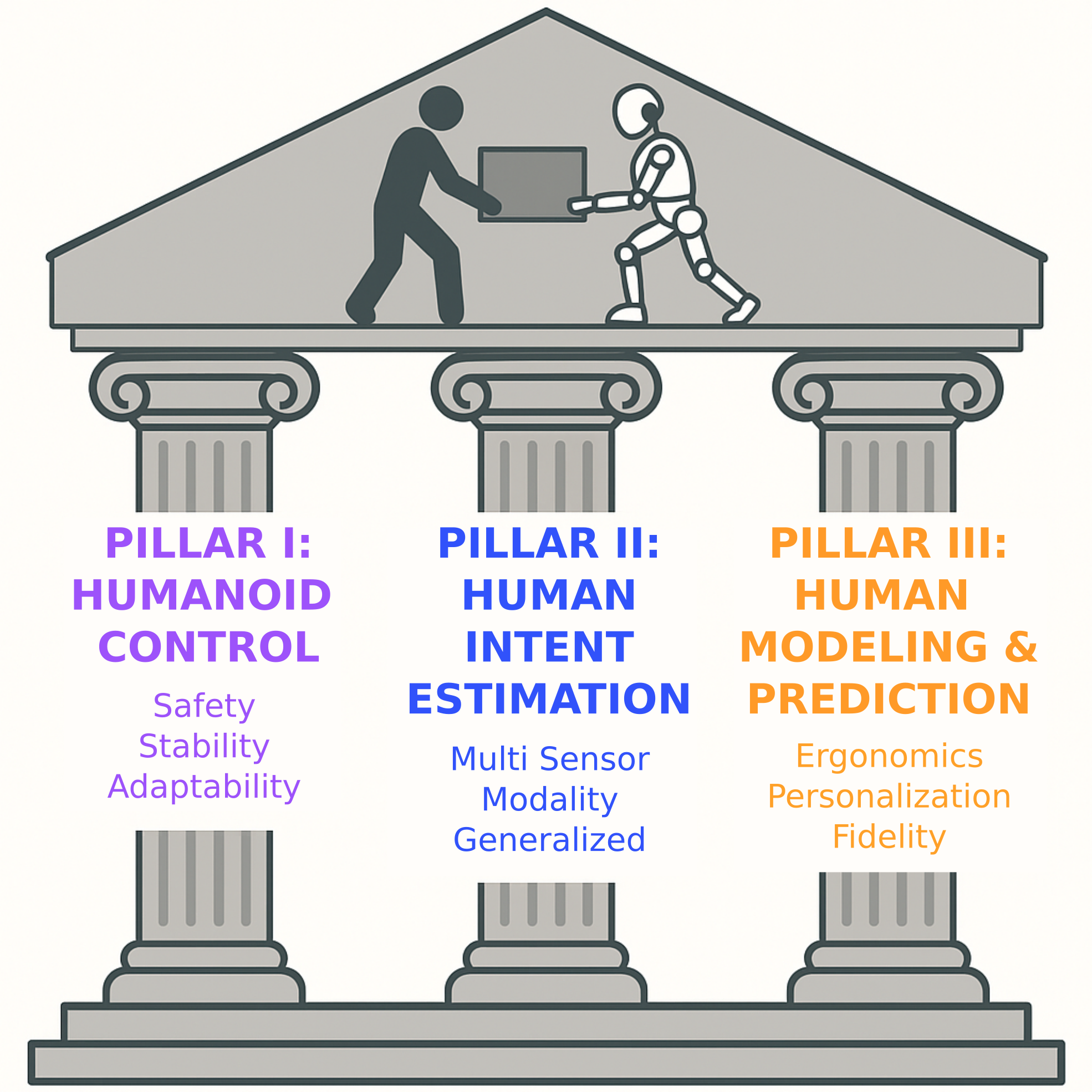}
    \caption{Conceptual structure of pHHI with three algorithmic pillars: Humanoid Control (Stability, Safety, Adaptability), Human Intent estimation (Multi-Sensor, Modality, Generalization), and Human Models (Ergonomics, Fidelity, Personalization).
    }
    \label{fig:pillars_intro}
\end{figure}

Given all the challenges of pHHI, we believe there are four key components that must be considered: hardware, humanoid modeling and control, human intent estimation, and computational human models. 
The foundational layer is the hardware~\cite{ficht2021bipedal}. 
This establishes the upper limit of system performance. 
It determines the robot's sensor and control bandwidth and its resilience to impact.
Hardware also determines the extent to which authority can be safely transferred to a human partner and the system's ability to handle unexpected events. 
Although hardware is recognized as a prerequisite, we do not provide a comprehensive review of mechanisms and actuators here. 
Instead, this review concentrates on the three algorithmic pillars that build upon this foundation to enable robust pHHI, as shown in Figure~\ref{fig:pillars_intro}. 

The first of these pillars centers on humanoid modeling and control. 
Here, we consider the principal features that the model and controller should have, such as stability, safety, and adaptability. 
Controllers must maintain balance and feasibility when facing unpredictable human-imposed forces (stability). 
They must protect both partners during contact events and transitions (safety). 
They should also intelligently reallocate momentum and authority as the task, environment, or partner's behavior changes (adaptability). 
These demands shape the design of control architectures, ranging from single locomotion activities to loco-manipulation. 
These requirements establish the foundation on which the other pillars can reliably build.

The second pillar centers on human intent estimation. 
Raw sensory observations are transformed into actionable beliefs and short-horizon forecasts under three critical themes. 
First, multi-sensor fusion is used to create low-latency, fault-tolerant cues by combining data from pose, wrench, physiological, and gaze sensors. 
Next, the choice of modality involves matching appropriate probabilistic, decision-theoretic, or learning-based models to each signal stream, ensuring timely and meaningful outputs. 
Finally, generalization ensures performance across diverse partners, tasks, and scenarios, allowing the system to resist distribution shifts and explicitly represent uncertainty that downstream controllers can utilize for safer decisions.

The third and final algorithmic pillar centers on the development of computational human models. 
Here, the robot moves beyond simple collision avoidance; it must consider what is comfortable, feasible, and fair for the person it is assisting. 
This pillar is defined by ergonomics, fidelity, and personalization.
These models must capture metrics of comfort, effort, and physical limits (ergonomics). 
They must be reliable enough to support predictions and constraint-setting. This can range from biomechanical to neuromechanical levels (fidelity). 
They must also adapt online to user-specific parameters and preferences (personalization). 
Exported as constraints, costs, or targets, these models allow the robot's controller to share effort and respect human limits intelligently. 
In doing so, these models enable the robot to move from simply preventing failure to actively promoting safe, efficient, and human-centered collaboration.

A primary challenge in the field is not the absence of sophisticated methods within each domain. 
Rather, there is a fundamental lack of integration among them. 
The existing literature often presents solutions that optimize individual aspects such as control, prediction, or human modeling in isolation. 
However, effective pHHI in real-world scenarios relies on the composition of these elements. 
For example, intent estimation must communicate calibrated uncertainty to the controller. 
Human models must define ergonomic limits as explicit constraints for the planner. 
The controller must provide stability assurances for higher-level processes. 
This review is motivated by the need for a unified framework with well-defined interfaces. 
Such a framework should connect sensing and prediction to whole-body action, while maintaining human comfort and safety.

The structure of this review paper begins with 
Section~\ref{sec:pHHI} that defines pHHI, outlines representative scenarios, and introduces a two-axis taxonomy (i.e., interaction modality and shared autonomy). 
Sections~\ref{sec:pillar1:mod_cont}–\ref{sec:pillar3:human_model} develop the three pillars: 
Section~\ref{sec:pillar1:mod_cont} covers humanoid dynamics and control. 
Section~\ref{sec:pillar2: intent} addresses human intent estimation and prediction. 
Section~\ref{sec:pillar3:human_model} reviews computational human models.  
In each pillar, we synthesize foundational and recent work, assess relevance to pHHI, and identify remaining challenges. 
In Section~\ref{sec:discuss_future}, we then transition to a broader discussion of the field's primary challenges and outline an architecture for a unified framework that integrates these proposed pillars. 
Additionally, we identify key future research directions.
Finally, Section~\ref{sec:conclusions} presents the conclusions of this comprehensive analysis.

\section{Review of Physical Humanoid-Human Interaction}
\label{sec:pHHI}

Advancements in physical human-humanoid interaction are deeply rooted in the evolution of the robot's ability to control its complex dynamics while safely managing external forces from a human partner. 
The literature reveals a progression from fundamental safety principles to highly integrated, task-specific applications. 
A comprehensive review of human-humanoid interaction and cooperation is presented in \cite{vianello2021human}, which provides a broad overview of this landscape. 
Our synthesis identifies three core themes that consider only physical interaction:
i) control strategies that seek to ensure safe and compliant behavior, 
ii) task-specific implementations for key pHHI scenarios such as collaborative transport and physical assistance, and
iii) emerging approaches that incorporate proactive and learned behaviors for more intuitive interaction.


\subsection{Control Strategies}
At the core of pHHI is the requirement that the robot must be physically compliant and safe to humans. 
Early and foundational work focused on establishing control laws that enable the robot to yield to external forces predictably, a principle essential for managing the uncertainties associated with physical contact.
Some of the control strategies found in the literature are as follows.

\subsubsection{Compliance and Force Control} 
A fundamental aspect of safe interaction in robotics is the ability to effectively modulate interaction forces.
This regulation is often achieved through a technique called \emph{impedance control}~\cite{song2019tutorial}, in which the controller establishes a desired dynamic relationship, similar to a virtual spring-damper system, between the robot's movements and the external forces it experiences~\cite{li2024variable}. 
This approach allows the robot to be ``soft" when pushed and ``stiff" when it needs to apply force. 
This principle is essential and has been successfully applied in various applications, such as helping individuals stand up~\cite{lopez2014compliant}, where compliant control enables the robot to assist the person without creating a rigid or uncomfortable motion. 
However, a significant challenge is maintaining whole-body balance while being compliant. 
In~\cite{hyon2007full}, this issue was addressed by developing a full-body compliant controller that can maintain balance even in the presence of large and unpredictable external forces exerted by a human partner.
In~\cite{KumbharArtemiadis2025MPCQP}, an \emph{admittance controller} is used, which, instead of regulating the dynamic relationship between position and force, finds a position for the Center of Mass (CoM) of the robot based on external forces generated by the expected interaction with an object jointly transported with a human.

\subsubsection{Whole-Body Control via Optimization}
Optimization-based whole-body control treats motion and force generation as a constrained optimization problem that enforces dynamics and contact feasibility while balancing task objectives in real time. 
In practice, two frameworks dominate: hierarchical formulations (cascaded Quadratic Programming (QP) techniques with strict priorities) and weighted-sum formulations (a single QP with tunable objective trade-offs).
In \emph{hierarchical optimization}, also known as \emph{Stack-of-Tasks}~\cite{bussy2012human, agravante2014collaborative, stasse2009fast, bussy2012proactive}, Whole-Body Control (WBC) is formulated as a sequence of strictly prioritized QPs. 
At the highest level, balance, contact feasibility, and actuation limits are enforced. 
Lower-priority tasks, such as interaction, end-effector positioning, and posture, are projected into the nullspace of higher-priority tasks. 
This approach is particularly practical for pHHI because it ensures that human–robot contact remains within stability and friction limits, even during disturbances caused by the human partner. 
However, this structure limits the ability to balance soft objectives, such as comfort and speed, across priorities, and it may introduce task-switching artifacts.
In contrast, weighted-sum optimization addresses WBC by solving a single QP that minimizes a weighted sum of task costs, such as tracking accuracy, effort, and interaction-wrench regulation, subject to dynamics and contact constraints~\cite{KumbharArtemiadis2025MPCQP, otani2018generating, romano2017codyco, agravante2019human}. 
This approach enables continuous, real-time trade-offs that are advantageous in pHHI. 
However, the effectiveness of this method depends on the selection of appropriate weights and the design of the cost function. 
Critical safety requirements should be enforced as hard constraints or protected through explicit feasibility checks.



\subsubsection{Learning Based Methods}
Learning-based approaches complement optimization by acquiring whole-body control policies or residual adaptations directly from data. 
These methods enhance robustness and flexibility in environments where frequent physical contact is present. 
Recent studies have demonstrated the scalability of skill acquisition on humanoid robots~\cite{bethala2025h2}. 
In addition to policy learning, two further research directions are considered. 
The first is Learning from Human Demonstration (LfD), which enables data-efficient imitation of expert behaviors. 
The second is Learning and Adapting Physical Interaction, which focuses on online adjustment of assistance, impedance, and force sharing during pHHI.
\paragraph{Learning from Human Demonstration (LfD)}
Instead of being explicitly programmed, robots can learn collaborative tasks by observing humans or being physically guided by them. 
This method is effective in teaching complex physical interactions. 
In~\cite{evrard2009teaching}, it was demonstrated that a humanoid robot could learn to lift objects through physical guidance. 
This approach, commonly referred to as imitation learning, enables the robot to capture the subtleties of the interaction, such as timing and force profiles, directly from an expert human~\cite{lee2011physical}.

\paragraph{Learning and Adapting Physical Interaction}
Beyond one-shot learning, some approaches enable robots to adapt their behavior continuously during interactions. 
The authors in~\cite{ikemoto2009physical} developed a framework for ``physical interaction learning," enabling robots to adjust their behavior in real-time to better coordinate with a human partner during cooperative tasks. 
Another strategy employs finite state machines to track the progress of a collaborative task effectively. 
This method involves passing through distinct phases, such as ``approaching," ``grasping," and ``lifting," while providing appropriate assistance at each stage~\cite{brecelj2023utilizing}.

\subsection{Task-Specific Implementations of pHHI}
Building on these foundational control strategies, researchers have demonstrated increasingly complex pHHI capabilities by applying them to specific, challenging interaction scenarios.
Some of these examples are as follows.

\subsubsection{Collaborative Transportation and Co-manipulation}
\begin{figure}[htbp]
    \centering
    \begin{subfigure}[b]{0.51\columnwidth}
        \centering
        \includegraphics[width=0.95\textwidth]{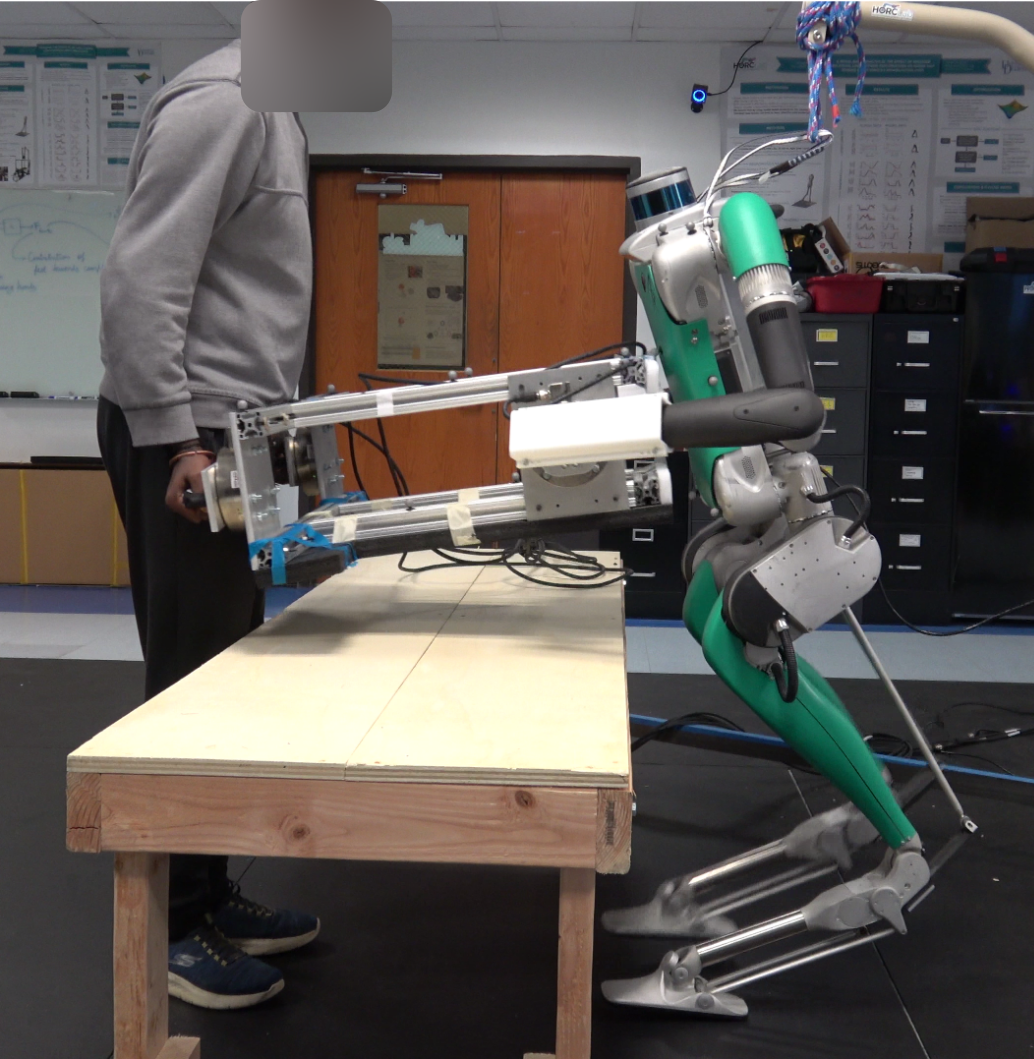}
        \caption{Indirect comanipulation}
        \label{fig:indirect}
    \end{subfigure}
    \hfill
    \begin{subfigure}[b]{0.48\columnwidth}
        \centering
        \includegraphics[width=\textwidth]{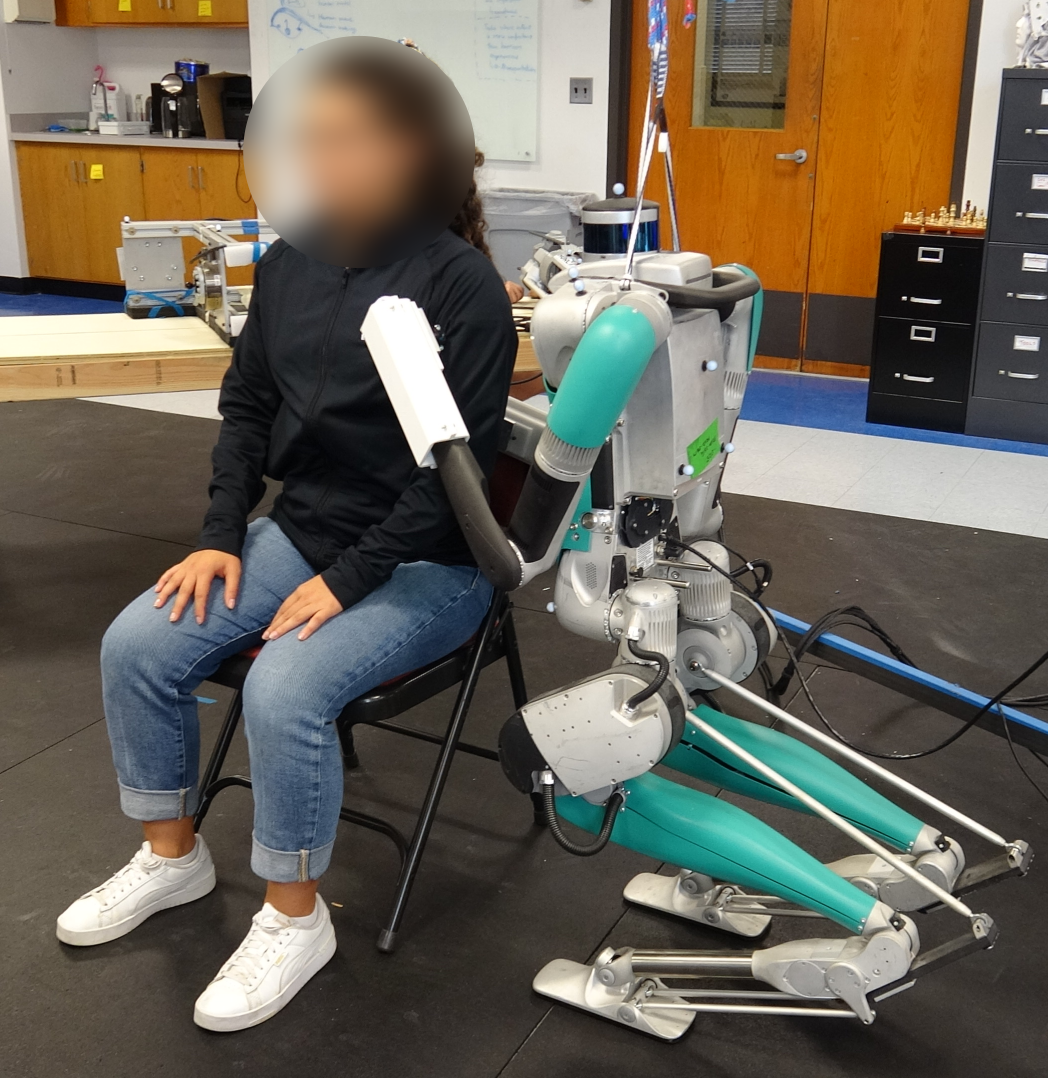}
        \caption{Direct interaction}
        \label{fig:direct}
    \end{subfigure}
    \caption{Comparison of interaction modalities in physical human–humanoid interaction. (a) Indirect, object-mediated co-manipulation (e.g., collaborative carrying), where coordination is mediated through a shared object and relies on shared object models and interaction-wrench regulation. (b) Direct, body-to-body assistance (e.g., emulated rehabilitation), which requires whole-body balance, reliable contact detection, and safe impedance/admittance control.}
    \label{fig:comparison}
\end{figure}

A substantial amount of research has concentrated on the fundamental task of two partners collaboratively carrying an object, as shown in Figure~\ref{fig:indirect}.
This task requires precise coordination of both locomotion and manipulation~\cite{monje2011new}. 
Early studies introduced homotopy-based controllers~\cite{evrard2009homotopy} designed for this purpose and examined the problem from a haptic feedback perspective. 
Notably, the work in~\cite{agravante2016walking, stasse2009fast} addressed the challenge using specialized walking pattern generators that adapt to each partner's influence. 
Then, in~\cite{agravante2014collaborative}, the approach incorporates both vision and haptic sensing to assess each partner's contribution, ultimately developing a comprehensive framework for human-humanoid collaborative carrying~\cite{agravante2019human}. 
Other researchers have investigated model-based control methods for managing load distribution~\cite{rahem2022human} or utilized vision systems to allow a robot to follow a human's lead while carrying a large object~\cite{stuckler2011following}. 
In~\cite{KumbharArtemiadis2025MPCQP}, an object was jointly transported by finding the robot's CoM targets based on an interaction admittance model.
A critical challenge in this field is making the interaction proactive rather than merely reactive. 
In~\cite{bussy2012proactive,maroger2022study}, the authors examined how a humanoid robot can anticipate a human's intentions, facilitating a smoother and more efficient joint transport task.

\subsubsection{Assistive Care and Ergonomics}
Another important application area is providing direct physical assistance in healthcare and daily living activities as shown in Figure~\ref{fig:direct}. 
The sit-to-stand and stand-to-sit motions are key benchmark tasks that require the robot to provide enough support without being excessive. 
Recent advances have demonstrated the effectiveness of robust controllers designed for this purpose, utilizing force sensing to synchronize with human movements~\cite{lefevre2024humanoid,lopez2014compliant}.
In addition to basic assistance, there is a new focus on ergonomic physical human-robot interaction (pHRI), where the robot's actions are optimized to minimize physical strain and reduce injury risk for the human partner. 
For instance, in~\cite{rapetti2023control}, a control approach is explicitly developed for ergonomic payload lifting, where the robot's controller takes into account a model of human ergonomics. 
This represents a significant shift from simply completing a task to doing so in a manner that prioritizes the well-being of the human user.
Other innovative forms of assistance include centaur-like systems that help individuals carry heavy loads over long distances~\cite{yang2022centaur}.

\begin{table*}[htbp]
\centering
\footnotesize
\setlength{\tabcolsep}{3
pt} 
\renewcommand{\arraystretch}{1.1}

\caption{Comparison of primary control strategies for physical human-humanoid interaction (pHHI). This table outlines their key features, advantages, limitations, typical applications, and critical safety implications.}
\label{tab:phhi_ctrl_strategies}

\begin{tabular}{%
    >{\RaggedRight\arraybackslash}p{0.11\textwidth} 
    >{\RaggedRight\arraybackslash}p{0.13\textwidth} 
    >{\RaggedRight\arraybackslash}p{0.14\textwidth} 
    >{\RaggedRight\arraybackslash}p{0.14\textwidth} 
    >{\RaggedRight\arraybackslash}p{0.13\textwidth} 
    >{\RaggedRight\arraybackslash}p{0.16\textwidth} 
    >{\RaggedRight\arraybackslash}p{0.07\textwidth} 
}
\toprule
\textbf{Control Strategy} &
\textbf{Key Features} &
\textbf{Advantages for pHHI} &
\textbf{Limitations} &
\textbf{Typical Use Cases} &
\textbf{Safety Implications} &
\textbf{Rep. Ref.} \\
\midrule

Compliance / Force Control &
Regulates end-effector/task-space dynamics; explicit force–motion blending. &
Safe and forgiving contact; intuitive feel; predictably yields to human wrenches. &
Trade-off between softness and tracking authority; heuristic contact handling. &
Hand-guiding, sit-to-stand assistance, and object co-transport. &
\emph{Inherently passive and stable}; provides a fundamental layer of safety. &
\cite{lopez2014compliant, hyon2007full, KumbharArtemiadis2025MPCQP} \\
\addlinespace

\multirow{2}{=}{Whole-Body Control (WBC) via Optimization} &
\textit{(Hierarchical QP)} Cascaded QPs with strict priorities; hard constraints. &
Feasible-by-construction; coordinates the entire body under contact. &
Rigid priorities; discontinuities at task/priority switches; needs reliable estimates. &
Balance during interaction (e.g., stabilizing the forearm while walking). &
Safety via hard constraints (joint limits, friction), but \emph{sensitive to unexpected forces}. &
\cite{bussy2012human, agravante2014collaborative, stasse2009fast, bussy2012proactive} \\

& \textit{(Weighted-Sum QP)} Single QP trading off multiple costs. &
Smooth real-time trade-offs (e.g., comfort vs. speed). &
Sensitive to weight tuning; safety must be explicitly encoded as constraints. &
Co-manipulation, sharing tasks based on vision and haptics. &
Relies on cost function design; no inherent safety guarantees without explicit constraints. &
\cite{KumbharArtemiadis2025MPCQP, agravante2019human, romano2017codyco, otani2018generating} \\
\addlinespace

Learning-based Methods &
Policies / residuals learned from data; imitation from demonstrations. &
Personalization to partner; captures nuances of timing/grip; complements physics models. &
Distribution shift, weak formal guarantees, and data/annotation burden. &
Teaching collaborative skills; adjusting assistance or impedance online. &
\emph{Provides no formal guarantees}; requires a safety-critical supervisor (e.g., CBF). &
\cite{bethala2025h2, evrard2009teaching, lee2011physical, ikemoto2009physical, brecelj2023utilizing} \\
\bottomrule
\end{tabular}
\end{table*}

\subsection{Insights and Discussion}
Table~\ref{tab:phhi_ctrl_strategies}, summarizes the control strategies mentioned in Section~\ref{sec:pHHI} into three groups and it outlines their key features, advantages, limitations, typical applications, and critical safety implications. 
As summarized in the table, compliance/force control delivers intuitive, ``forgiving” contact by shaping task-space dynamics. 
Its softness trades away authority and often relies on a lower layer to maintain balance and limits. 
Optimization-based WBC enforces multi-contact feasibility. 
Hierarchical QPs ensure feasibility but can show discontinuities at task or priority switches. 
Weighted-sum QPs allow smooth trade-offs but require explicit safety encoding. 
Learning-based methods add personalization and fill modeling gaps, however, without a constraint-aware wrapper, they are non-robust to shifts and rare events. 
In practice, the learning-based methods work best as residuals or reference generators under a certified WBC/MPC layer.

In most of the approaches reviewed in this section, robotic initiative is limited to reactive behaviors~\cite{lefevre2024humanoid, lopez2014compliant, KumbharArtemiadis2025MPCQP, stasse2009fast, monje2011new, evrard2009homotopy, rahem2022human, stuckler2011following, yang2022centaur}. 
Robots respond to measured interaction forces, follow the motion of their human partners, and redistribute loads to maintain compliance and adhere to safety standards. 
Recent studies have examined predictive and proactive strategies. 
In predictive approaches~\cite{agravante2019human, agravante2014collaborative, rapetti2023control}, controllers are adapted based on forecasts of ergonomic risks or anticipated human contributions. 
However, the robot does not initiate task goals. 
In contrast, proactive strategies~\cite{bussy2012proactive, maroger2022study} involve anticipating human partner movements or inferring intent through subtle actions, followed by redistributing momentum or adjusting footstep timing. 
Sensing modalities for reactive strategies primarily use haptic feedback, while predictive and proactive approaches incorporate kinematic and visual data.

Despite impressive progress in modeling and control strategies for pHHI tasks, research still tends to focus on isolated yet critical subproblems.
This fragmentation is understandable, as safe, compliant controllers, balancing under contact, intent inference, and human variability are each challenging problems on their own. 
However, this conservative approach leaves a gap between lab demonstrations and deployable interaction.
A common and cohesive framework is needed: one that first \emph{models and controls the humanoid} at the whole-body, multi-contact level, while guaranteeing safety in an efficient and tractable way. 
Second, it \emph{estimates and forecasts human intent} from noisy multimodal signals. 
Third, it embeds \emph{human biomechanical/behavioral models} so that the robot optimizes not only the task success but also comfort and effort. 
In this review paper, we use this lens to survey the landscape and highlight bridging scenarios where currently separate lines of work can be stitched together into an actionable framework.
We approach this by framing the field around three key pillars: humanoid modeling and control, human intent estimation, and computational human models. 
This approach helps reveal the requirements, challenges, and opportunities for building safe, adaptive, and human-centered systems.

\section{Pillar I: Dynamics Models and Control Strategies for Humanoid Robots}
\label{sec:pillar1:mod_cont}
\begin{figure*}
    \centering
    \includegraphics[width=\linewidth]{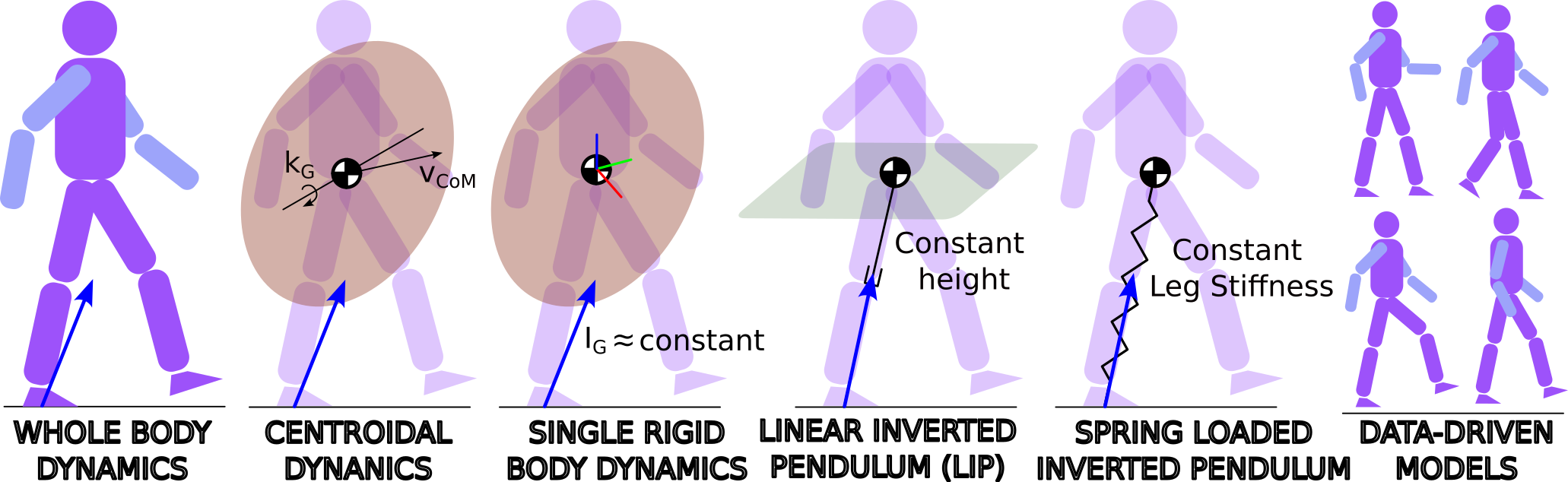}
    \caption{Modeling methodologies surveyed in this paper, arranged from high-fidelity to reduced-order and learned abstractions: whole-body multibody dynamics (WB), centroidal dynamics, single-rigid-body (SRB) approximations, Linear Inverted Pendulum (LIP), Spring-Loaded Inverted Pendulum (SLIP), and data-driven models. 
    }
    \label{fig:models}
\end{figure*}

Humanoid control operates at the intersection of free-floating, underactuated, hybrid, and multi-contact dynamics, making even seemingly simple motions complex. 
This complexity transforms physical interactions with people into a challenge of balancing feasibility, responsiveness, and safety. 
In this section, we explore this challenge by surveying the main modeling abstractions, ranging from full-order whole-body and centroidal dynamics to reduced-order templates and learned residual models, as shown in Figure~\ref{fig:models}. 
Additionally, we mention their corresponding control paradigms, including classical control techniques, optimal control formulations with safety certificates, and learning-based controllers.
For each method, we analyze its benefits (e.g., fast balance reflexes, force-aware coordination, comfort, and predictability), its limitations (e.g., model mismatch and latency), and how well it scales across different interaction modalities and levels of shared autonomy. 
Our goal is not to create an exhaustive catalog but to provide a coherent overview of fit-for-purpose tools and their complementary roles, setting the foundation for developing a safe, intuitive, and robust control stack for pHHI.

\subsection{Dynamic Models for Humanoids}
\subsubsection{Whole-body dynamics}
At the core of this discussion are whole-body models of humanoid robots, which consist of numerous rigid links, closed contact constraints, and a free-floating base, making the entire system underactuated. 
These models, whether expressed in the \emph{Lagrangian or Newton-Euler} form, serve as the foundation for modern control strategies, including Resolved Momentum Control (RMC) and optimization-based WBC systems~\cite{kajita2003resolved, orin2013centroidal, kuindersma2016optimization, dai2014whole, ponton2016convex}.
In the context of pHHI, the accuracy of these models is crucial. 
When a person interacts with the robot by stabilizing it at the forearm or sharing a load through their hands, the forces involved result in motion effects at the base and ground contacts. 
Even minor inaccuracies in mass, inertia, or contact modeling can lead to noticeable issues, such as stiffness spikes or drift during interactions.
Using full models enables us to analyze balance, friction, and actuation limits simultaneously, an essential factor for practical direct assistance and co-manipulation. 
However, a challenge in pHHI arises from the combination of parametric uncertainty, the time-varying nature of the human impedance, and the need for real-time fidelity models. 
Without additional layers of compliance and predictive control, achieving perfect model matching is often unrealistic.

\subsubsection{Simplified abstracted models}
To respond quickly to human actions, controllers utilize reduced-order models that consider different levels of abstraction. 
For instance, the Single Rigid Body (SRB)~\cite{dai2014whole} and \emph{Linear Inverted Pendulum Model (LIPM)} and its cart-table interpretation simplify balance to Center of Mass (CoM) motion and a feasibility region for Zero Moment Point (ZMP) or Center of Pressure (CoP). 
This simplification provides efficient guidelines for foot placement and CoM adjustment~\cite{wieber2006trajectory, kajita1991study,caron2018capturabilitybased, caron2018balance, hirukawa2006universal}. 
Another well-used model is the \emph{Spring-Loaded Inverted Pendulum (SLIP)}~\cite{shahbazi2016unified,geyer2019gait}, which incorporates a compliant leg and better captures the bounce dynamics compared to the rigid-leg assumption of the LIPM model. 
This feature allows it to buffer impact forces and exhibit self-stabilizing properties. 
The \emph{capture point} and \emph{Divergent Component of Motion (DCM)} extend these concepts into smooth stepping and push-recovery strategies that are effective even when faced with sudden human movements~\cite{koolen2012capturability, englsberger2011bipedal, englsberger2017smooth}.
In pHHI, these are the foundation of concepts for the robot's balance framework: when a partner misjudges the timing of a handover or shifts a shared object, the robot can execute reactive actions, such as stepping, adjusting the CoM, or adjusting its gait. 
Actions can be computed within milliseconds, rather than engaging in the full complexity of the whole-body dynamics. 
However, these abstractions prioritize computational speed over model fidelity. 
They typically assume coplanar and rigid contacts, and negligible angular momentum.  
Additionally, in the case of LIPM, they often neglect swing and impact dynamics as well as hand or object contacts and consider nearly constant CoM height. 
As a result, ZMP or capture margins may be inaccurately estimated during large disturbances or when operating on uneven terrain. 
In pHHI, where a human applies time-varying forces and introduces individualized constraints, this modeling gap becomes more notorious. 
Anticipatory control and personalization remain limited unless the model incorporates external wrench estimation, non-coplanar contacts, and regulation of variable height and momentum.

\emph{Centroidal dynamics} 
serve as a compact model of CoM motion and global angular momentum, sitting between full fidelity and point-mass models. 
This approach captures how all joints contribute to maintaining whole-body balance~\cite{orin2013centroidal, dai2014whole, ponton2016convex}. 
Centroidal planners schedule feasible contact phases and momentum profiles. 
In practice, a whole body controller with inverse dynamics maps the centroidal momentum schedules to torque commands that respect the friction, contact, and actuation limits on the hardware, as demonstrated on Atlas-class platforms~\cite{kuindersma2016optimization, dai2014whole, ponton2016convex}.
When viewed through a pHHI lens, this middle layer becomes crucial. 
Direct assistance tasks, such as sitting-to-standing movements and standing during gait, require precise regulation of the interaction forces in the hands while ensuring global balance in the feet. 

However, centroidal models often neglect limb kinematics and generally assume predefined contact schedules, rigid contacts, and precise CoM or momentum estimates. 
Translating momentum plans into whole-body torques may encounter joint or velocity constraints, self-collisions, or reachability limitations, particularly when significant hand forces or non-coplanar grasp conditions are present. 
In pHHI, unmodeled partner impedance and sensing delays can compromise external-wrench regulation. 
Finally, inaccurate centroidal estimates or friction cone assumptions may result in infeasible or excessively aggressive momentum commands.

\subsubsection{Data-driven modeling}
Due to the challenges of obtaining precise models for complex humanoids and the unpredictability of multi contact scenarios, \emph{data-driven modeling} has become an essential tool. 
Data-driven dynamics, including learned residuals and models, complement physics in situations where uncertainty is a significant factor, such as partner-specific stiffness, unknown payloads, and unmodeled compliance. 
Neural policies and residuals are increasingly integrated into whole-body control stacks and loco-manipulation planners~\cite{radosavovic2023learning, lai2023sim, dugar2024learning, paredes2022resolved, dao2024sim, xie2023hierarchical, cheng2024expressive, ji2024exbody2, liu2024opt2skill, fu2024humanplus}. 
In the context of pHHI, these data-driven methods prove most useful as enhancements to the overall control architecture. 
For instance, a residual could compensate for an individual's grip stiffness, or a learned predictor might adjust expectations based on a partner's characteristic timing.
However, learned models are sensitive to distribution shifts (e.g., new partners, surfaces, payloads) and provide weak guarantees of safety and stability. 


\subsection{Control Strategies for Humanoids}
\subsubsection{Classical control}
\paragraph{Whole-Body Control (WBC) via inverse dynamics} 
A common approach to WBC on humanoids is operational-space inverse dynamics, which prioritizes tasks and regulates internal forces. 
This method builds on torque control at key coordinates~\cite{khatib1987operational}, organizing constraint tasks, motion tasks, and posture in a hierarchy. 
Lower-priority tasks are projected into the null space of higher-priority ones, enabling compliant interaction at the torque level~\cite{sentis2005freefloating,sentis2006wholebody,park2008multiplecontact}. 
Internal-force control has also been introduced, allowing unified regulation of the center of mass, operational tasks, and internal wrenches~\cite{sentis2010compliant}. 
Momentum-based methods, such as Resolved Momentum Control, handle both linear and angular momentum as a whole-body task~\cite{kajita2003resolved}, thereby helping to maintain balance and manage interaction dynamics within inverse-dynamics controllers.
When interacting with unknown objects, WBC can utilize online estimation of operational forces to adapt motion and force allocation as payloads change, thereby maintaining balance and contact~\cite{nozawa2010full}.
Whole-body impedance control at the torque level has been demonstrated on robots like DLR Justin, enabling self-collision avoidance and compliant multi-task manipulation~\cite{dietrich2011dynamic,dietrich2012reactive}.

In pHHI, torque-level whole-body control (WBC) is advantageous because it provides direct access to interaction ports, meaning the points of contact, such as the hands, forearms, and torso. 
This approach incorporates impedance control (which resists motion in response to force) and internal force or momentum regulation, allowing for transparent and adjustable contact forces. 
As a result, the system can adapt to unknown forces from partners or objects.

The primary limitations of this approach in pHHI would be the management of inequality constraints, such as friction cones (the allowable directions of force to avoid slipping) and torque or CoP bounds, as well as managing model uncertainty, such as payload shifts and soft contacts.
Additionally, the method depends on precise torque sensing or estimation.
In practice, many control architectures combine inverse-dynamics whole-body control with supervisory safety monitors, while comprehensive inequality constraint handling is typically addressed in the optimization section discussed later.

Promising directions for improvement include embedding contact and wrench observers directly within the WBC loop, and leveraging explicit centroidal momentum tasks to absorb partner-induced twists with the whole body rather than relying solely on ankle torques~\cite{orin2013centroidal,dai2014whole,ponton2016convex}. 
Additional enhancements involve adding safety layers that bound the CoM, ZMP, and friction regardless of the nominal task stack~\cite{hsu2015control}, incorporating learned residuals to address model discrepancies without neglecting physical principles~\cite{dugar2024learning,paredes2022resolved}, and implementing automatic gain tuning to reduce the need for manual adjustments when switching between humans or tasks~\cite{sartore2024automatic}.

\paragraph{Impedance, admittance, and hybrid force/position control}
Impedance, admittance, and hybrid force/position control are crucial concepts in humanoid loco-manipulation. 
If WBC determines the forces to apply, impedance and admittance govern how these forces are exchanged. 
In practice, humanoid manipulation controllers surround their interaction ports, such as hands and forearms, with whole-body impedance. 
This design allows for compliant motion when faced with unexpected human wrenches, ensuring that stance contacts remain feasible~\cite{murooka2021humanoid, audren2014model, harada2007real, KumbharArtemiadis2025MPCQP}. 
This is the layer that individuals experience, it influences whether interaction feels soft, whether handovers are forgiving, and whether teleoperation remains safely energetic despite operator variations and uncertainties.

For indirect interactions, such as carrying a tray, hybrid force/position control separates the axes, allowing vertical force to be regulated while maintaining precise horizontal motion.
This helps prevent sudden changes in load sharing. 
The challenges here involve classic trade-offs: if the system is too soft, the robot loses control, especially with heavy loads. 
If too stiff, the human may feel uncomfortable surges in force, and high admittance can endanger balance without proper constraints.

Improvements for pHHI could include variable impedance schedules that adapt to different phases of a task, for example, a softer approach, a firm lift, and a gentle handoff. 
Comfort can also be enhanced by limiting wrench rate and jerk, and by modulating admittance gains based on ZMP and CoP margins together with Control Barrier Function (CBF) constraints~\cite{fahmi2019passive,audren2014model}. 
Finally, co-designing with the whole-body controller enables impedance to shape the interaction ports while ensuring that the WBC preserves overall feasibility~\cite{harada2007real}.

\subsubsection{Optimal control and MPC/MPPI}
Optimal control frames humanoid behavior as a constrained optimization problem, with Model Predictive Control (MPC) offering anticipatory control by optimizing footsteps, contacts, and force sharing over a receding horizon, allowing for replanning as conditions change. 
\paragraph{Modular Approaches}
The high complexity and dimensionality of humanoids motivate a modular framework consisting of a high-level planner and a low-level controller. 
The high-level planner performs predictive planning of COM motion, footstep locations, forces, etc., while the low-level controller executes these plans on the physical robot. 
Both the high-level and the low-level are often formulated using optimization schemes. 
Model Predictive Control (MPC) has been widely used for footstep planing in humanoid locomotion~\cite{diedam2008online}.
Linear Inverted Pendulum (LIP) based MPC coupled with a low-level quadratic program has been widely used for legged locomotion on flat surfaces \cite{wieber2006trajectory, herdt2010online, Kumbhar2025Finite}. 
This method regulates the CoM and ZMP, placing steps to counteract pushes without solving full-order dynamics~\cite{wieber2006trajectory}. 
However, a limitation of this approach, especially for pHHI, is contact myopia: arm and interaction forces, as well as human-induced wrenches, are not perfectly modeled, which can lead to abrupt interactions. 
The authors explored extending LIP to account for interaction and combined impedance control in~\cite{KumbharArtemiadis2025MPCQP}, which allows minor human-robot mismatches to be absorbed compliantly. 
Another reduced-order model that has been widely used for humanoid walking is the Single Rigid Body (SRB)~\cite{dai2014whole} and the Spring-Loaded Inverted Pendulum (SLIP)~\cite{geyer2006compliant, xiong2020dynamic}.

Most reduced-order models are obtained by making assumptions about the centroidal dynamics~\cite{orin2013centroidal}. 
Approaches that use centroidal dynamics with MPC optimize CoM and global momentum while explicitly considering contact schedules, friction cones, and wrench bounds, eventually realizing the plan through whole-body control. 
This methodology has been successfully applied for rough terrain navigation and contact-rich maneuvers~\cite{kuindersma2016optimization}. 
Additionally, it has been extended with multi-contact previews~\cite{audren2014model} to coordinate hand and foot movements, effectively leveraging centroidal dynamics with full kinematic consistency~\cite{dai2014whole}. 
For pHHI, centroidal MPC could co-optimize foot placement. 
Nonetheless, there are challenges related to latency (due to large quadratic programs/nonlinear programs) and model fidelity under human variability. 

\paragraph{Hierarchical Approaches}
Stack-of-Tasks (SoT) methods organize whole-body control into a hierarchy of tasks, where each task is assigned a strict priority level.
High-priority constraints (e.g., contacts, balance, joint/torque limits) are enforced first, while lower-priority tasks are projected into the corresponding null spaces, resulting in predictable behavior in the event of task conflicts. 
Foundational ideas trace back to operational-space control and task-priority Inverse Kinematics (IK) for redundant robots \cite{Khatib1987,Baerlocher1998,Mansard2007}. 
In modern humanoids, SoT is often realized as hierarchical QPs that solve a cascade of convex problems at each cycle, enabling real-time multi-contact behaviors while preserving hard constraints \cite{Escande2014,SentisKhatib2006}.
These formulations have been demonstrated for dynamic whole-body motion with unilateral contact/friction constraints, as well as on torque-controlled humanoids with momentum regulation \cite{Saab2013,Herzog2016}.

\paragraph{Sampling-based MPC (MPPI)}
The Model Predictive Path Integral (MPPI) method bypasses large, nonlinear programs by sampling multiple rollouts each cycle and averaging the best results. 
This produces a gradient-free controller capable of tolerating nonconvex costs and contact discontinuities, making MPPI particularly suitable for contact-rich situations and rapid ``what-if" re-planning during cooperative tasks~\cite{alvarez2024real, xue2024full}. 
For pHHI, key considerations include the trade-off between sample budget and latency, cost shaping to account for human comfort and safety, and addressing stochastic variability. 
The sampling-based nature of MPPI could also be leveraged with learned coupled dynamics of the humanoid and the human. This could lead to robust motion planning and efficient collaboration.  

\paragraph{Differential Dynamic Programming (DDP)}
Differential Dynamic Programming (DDP) is a method for solving optimization that alternates forward rollouts with a backward quadratic expansion of the value function, yielding both feedforward control updates and time-varying Linear Quadratic Regulator (LQR) feedback gains. 
For legged and humanoid systems, DDP (and first-order variants such as iterative Linear Quadratic Gaussian (iLQG) / iterative Linear Quadratic Regulator (iLQR)) has been used to generate and stabilize whole-body motions online (e.g., get-up, disturbance recovery), and to respect input bounds via control-limited updates \cite{Tassa2012,Tassa2014}. 
Recent formulations address contact-rich locomotion by optimizing multi-phase rigid-contact dynamics and even hybrid switching/impact events while handling friction and torque limits through augmented-Lagrangian and barrier techniques~\cite{Budhiraja2018,LiWensing2020}.

\subsubsection{Learning-based control}
With the rise of machine learning, many researchers are using these techniques for humanoid control, aiming to achieve more general or adaptive behaviors that are difficult to design manually. 
These approaches are particularly appealing for pHHI because human behavior is complex to model. 
In theory, a learning robot could adapt to individual human partners or acquire complex interactive skills through examples or trial and error.

\paragraph{Reinforcement learning (RL)}
Deep Reinforcement Learning (RL) has enabled the development of whole-body locomotion and emerging loco-manipulation skills through extensive simulation training. 
Recent advancements include transformer-based locomotion policies for humanoid robots~\cite{radosavovic2023learning}, sim-to-real curricula that enhance the robustness of hardware~\cite{lai2023sim}, and contact-rich loco-manipulation controllers for tasks like pushing and carrying boxes~\cite{dao2024sim, xie2023hierarchical}.
The appeal of RL for pHHI lies in its ability to personalize actions and discover strategies; a policy can learn when to step, brace with the arms, or adjust grip and effort based on cues from a partner, which behaviors can enhance model-based designs~\cite{schwarke2023curiosity}.

We would like to direct readers to \cite{bao2025deepreinforcementlearningbipedal} for a detailed review of learning-based methods for humanoids. 
The authors categorize these approaches into two primary types: end-to-end frameworks \cite{Siekmann2020Learning,Siekmann2021Sim,Xie2018Feedback,xie2020learning, Li2021Reinforcement}, which directly map sensory inputs to control actions, and hierarchical frameworks \cite{fu2024humanplus, jiayi_2024,Singh2022Learning}, which decompose locomotion into layers such as high-level planning and low-level control. 
Importantly, they highlight how future research can leverage the complementary strengths of optimization- and learning-based methods within unified control frameworks. 
For example, learning can be made more sample-efficient by incorporating guided policy learning \cite{Siekmann2020Learning, levine2013guided, gangapurwala2020guided}, where expert demonstrations or planner-generated trajectories are used to guide policy training in both end-to-end and hierarchical settings. 
Likewise, optimization methods can be made more robust to uncertainty and unmodeled dynamics by incorporating residual learning, where learned corrections augment model-based baselines. 
These strategies for fusing optimization and learning not only advance locomotion control but also provide inspiration for pHHI, particularly in developing coupled models, human-aware control, and improved robustness in collaborative tasks.

In practice, on-policy training methods (such as Proximal Policy Optimization (PPO) and Trust Region Policy Optimization (TRPO)) offer stability but require a substantial amount of data, while off-policy methods (like Soft Actor Critic (SAC) and Deep Deterministic Policy Gradient (DDPG) can be more efficient. 
However, they are often challenging to fine-tune. 
This has led to a reliance on domain randomization and structured curricula to facilitate transfer from simulation to real-world applications.
The main challenges for pHHI include safety, as unsafe exploration is unacceptable when humans are involved, and the sim-to-real gap. 
Possible solutions involve exposing policies only as high-level references, while simultaneously implementing impedance and walking behavior control (WBC) safety layers, using conservative action spaces, and shaping rewards to prioritize comfort, for example, by minimizing sudden movements and maintaining bounded forces.

\paragraph{Imitation learning (IL) \& inverse RL}
Instead of starting from scratch, Imitation Learning (IL) utilizes demonstrations, such as teleoperation or motion capture, to acquire natural, human-like whole-body behaviors. 
This can then be optionally refined using RL for greater robustness. For pHHI, IL can capture socially appropriate timing, effective handovers, or partner-following strategies~\cite{cheng2024expressive,ji2024exbody2,liu2024human,fu2024humanplus}. 
Meanwhile, Inverse Reinforcement Learning (IRL) learns reward functions that align with human preferences, such as ensuring stability while minimizing abrupt movements, which enhances comfort during assistance~\cite{wu2024infer}.
However, there are significant challenges, including covariate shift (when the robot's experiences deviate from those observed during demonstrations) and limited exposure to edge cases.
Possible solutions involve aggregating datasets and employing IL-to-RL pipelines to enhance robustness. 
Additionally, using uncertainty-aware execution can help; if the robot's confidence is low, it can revert to compliant tracking to maintain safety and reliability.

\paragraph{Behavior/Policy foundations, language \& teleoperation}
Recent research has investigated language-guided and ``foundation-style" policy models that can compile and adapt whole-body skills from extensive datasets. 
This approach allows for rapid adaptation to new tasks with minimal prompts or few examples~\cite{kumar2023words, jiang2024harmon, yao2024anybipe, sun2024prompt}. 
When combined with universal whole-body teleoperation, these systems can gather high-quality demonstrations and facilitate clear shared autonomy during contact-rich collaborations~\cite{he2024omnih2o, fu2024humanplus}. 
In pHHI, such models help align robot behavior with human intent, utilizing either language or operator cues, while also preserving the subtleties of contact learned from diverse prior data. 
The trade-offs involved include scale (data and computational resources), assurance, and interpretability. 
In practice, these models are deployed alongside certified balance and safety layers, as well as hierarchical whole-body control, ensuring that high-level instructions are executed smoothly and predictably.
Although this approach shows great potential for achieving outstanding performance in complex tasks, significant work remains to be done, particularly in avoiding hallucination, ensuring safety guarantees, and preventing Large Language Model (LLM)-Jailbreaking~\cite{robey2024jailbreaking, zhang2024badrobot}.

\subsubsection{Stability/safety certificates: ZMP / CoP, Capture-Point / DCM, HZD, CBF-QP}
When considering pHHI, a humanoid must ensure balance and safety in real-time, even if higher-level MPC or learning may hesitate or err. 
At the low-levels, techniques such as ZMP/CoP feasibility and DCM/Capture Point stepping provide rapid corrections in milliseconds, keeping the CoM in a recoverable state and enabling smooth push recovery.
For continuous locomotion, Hybrid Zero Dynamics (HZD) guarantees stable gaits for underactuated bipeds, offering a solid foundation upon which regulation and disturbance rejection can be built~\cite{westervelt2003hybrid, westervelt2018feedback, grizzle2014models}. 
As a runtime safety measure, Control Barrier Function Quadratic Programs (CBF-QPs) enforce forward-invariant sets, such as no-slip cones, CoM envelopes, and joint/torque limits, regardless of the nominal controller, filtering out unsafe regions during execution~\cite{hsu2015control}.
The main challenges include certifying the composition across multi-phase loco-manipulation, designing uncertainty-aware sets considering unknown human forces and sensor delays, and maintaining comfort by avoiding abrupt or overly cautious reactions. 
Promising avenues for improvement include: 
i) personalizing barrier sets with partner-specific stiffness and bounded force models, 
ii) implementing rate-limited barriers to control jerk and force rate, 
iii) co-designing with impedance and centroidal/WBC to minimize interventions from barriers, 
and iv) integrating barriers with learned uncertainty, tightening sets when models are weak and relaxing them when confidence is high, all while maintaining certified dynamics at the core.

\begin{figure*}[t]
    \centering
    \includegraphics[width=\linewidth]{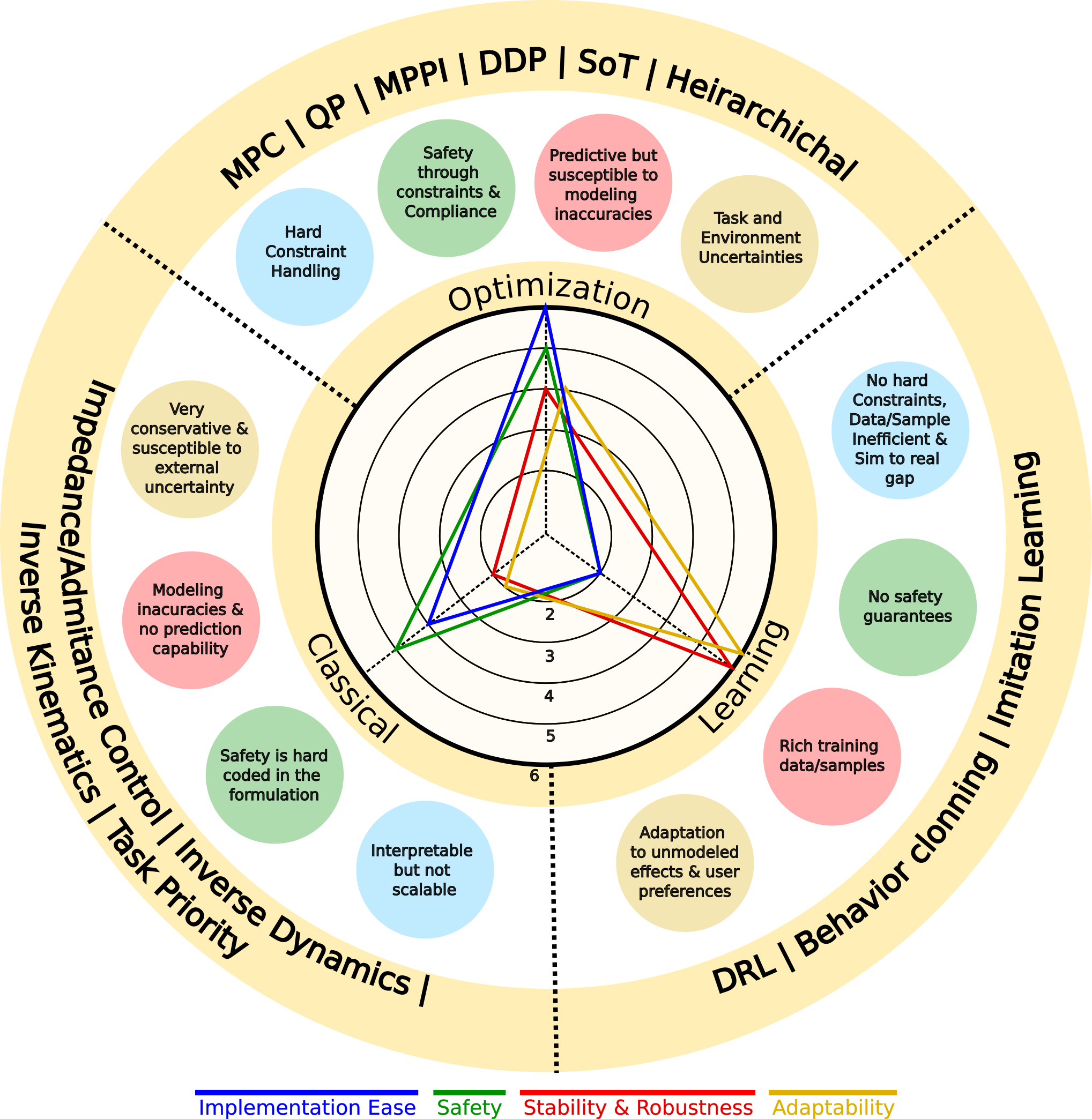}
    \caption{Conceptual Framework of Pillar I: Humanoid Modeling \& Control. Classical, optimization-based, and learning-based control paradigms are compared using a radar chart that highlights safety, stability and robustness, adaptability, and implementation effort. The scores presented are qualitative and comparative rather than absolute measurements.}
    \label{fig:pillar_I}
\end{figure*}

\subsection{Pillar I Insights and Discussion}
A humanoid controller designed for pHHI should exhibit four key features: \emph{safety}, to ensure human well-being during close physical contact;
\emph{stability and robustness}, to guarantee reliable performance under external disturbances and varying loads;
\emph{adaptability}, to accommodate diverse human partners, tasks, and environments;
and \emph{ease of implementation}, to enable practical deployment on complex humanoid systems. 
Guided by the humanoid control architectures discussed in Pillar~I, we examine how three major control paradigms, classical, optimization-based, and learning-based methods, perform with respect to these key requirements for pHHI, as illustrated in Figure~\ref{fig:pillar_I}.

The first feature considered is safety, defined as the protection of the human during controller operation. 
Safety encompasses the robot's compliance during human contact, robustness under disturbances or model mismatches, and the level of trust that the human perceives. 
Classical and optimization-based controllers generally provide comparable safety, as safety can be enforced by design and, in optimization-based methods, modulated through hard and soft constraints. 
Both approaches, however, remain vulnerable to model inaccuracies and external disturbances. Incorporating impedance or admittance models can mitigate these vulnerabilities. 
Classical controllers may produce more aggressive motions, while optimization-based controllers can address this issue. 
Learning-based controllers offer smoother motion but currently lack full explainability and interpretability, which reduces their perceived safety due to uncertainty regarding potential failures that could endanger humans.
Additionally, it is challenging to add explicit safety constraints during policy learning. 
Constrained Proximal Policy Optimization (CPPO)~\cite{gangapurwala2020guided} enables adding a threshold on the cumulative violation of all constraints, but lacks individual thresholds and hard constraints. 
Safety can also be ensured to some extent by guiding policy learning to avoid unsafe actions.

The second feature considered is stability and robustness. 
This accounts for the robot's balance and its robustness in compensating for or anticipating disturbances that could lead to a misstep or fall.
In this category, we consider that learning-based methods outperform classical and optimization-based methods. 
The primary reason is that learning-based methods are more effective at adapting and compensating for unmodeled effects, switching states, and disturbances. 
An optimization-based approach could explicitly trade off balance, tracking, and effort, yielding whole-body responses or employing predictive methods that decide to step early (specifying where/when/how far) in order to maintain stability. 
Classical methods can be myopic, reacting only after errors have grown, often too late, and lacking predictive capability.

The third feature considered is adaptability, defined as the controller's ability to personalize to user preferences, manage changing arbitration among objectives, and generalize across tasks. 
Learning-based methods generally demonstrate superior adaptability, as learned policies can be conditioned on context and preference signals such as embeddings, demonstrations, or reward shaping. 
These policies can be updated online through residuals, enabling rapid personalization without redesign. 
Learning-based controllers also adapt effectively to changes in partners, payloads, or goals. 
Optimization-based controllers can adapt by re-solving with updated costs, priorities, and constraints or by incorporating online parameter estimates. 
However, each modification requires problem re-specification and incurs solver latency, while generalization is constrained by the selected model and task encoding. 
Classical controllers encode preferences implicitly through gains and heuristics, and significant changes in arbitration or task necessitate manual retuning and often re-architecting of the control loops.

The final feature considered is ease of implementation. 
Optimization-based approaches are generally preferred because they enable the formulation of tasks and safety requirements as costs and hard constraints, and benefit from the use of established commercial solvers. 
While tuning of weights and models remains necessary, these methods scale effectively as interaction complexity increases. 
Classical controllers provide transparency for individual behaviors, but their implementation becomes more complex as the number of contacts and arbitration scenarios increases, requiring additional gains, heuristics, or operational modes.
Learning-based methods present greater implementation challenges, as they demand extensive safety-validated datasets or high-fidelity simulations, must address simulation-to-reality transfer and exploration risks, and often require supplementary safety filters because constraints are not inherently integrated.

\section{Pillar II: Human Intent Estimation}
\label{sec:pillar2: intent}
A robust humanoid controller, as presented in the previous section, allows the robot to manage its dynamics and remain resilient to external disturbances. 
However, proactive physical collaboration with a humanoid robot depends on the robot's ability to anticipate its partner's next actions and the intensity with which they will act. 
In this section, we first define human intent, which encompasses everything from low-level desired trajectories and force profiles to higher-level goals, roles, and cognitive states. 
This clarification helps us understand what needs to be estimated and the timeframes involved.
Next, we examine various sensing modalities for detecting intent, including kinematic and pose data, force-torque and tactile feedback, physiological signals, and gaze cues. 
We emphasize what each modality reveals and conceals in contact-rich settings.
Based on these signals, we review models and algorithms for predicting intent. 
These include probabilistic filters, movement primitives, inverse optimal control/inverse reinforcement learning for inferring objectives, and Partially Observable Markov Decision Process (POMDP) formulations for actions taken under uncertainty. 
We also explore modern learning methods, including Recurrent Neural Networks (RNNs), Transformers, and generative models, with a focus on real-time applications, personalization, and calibrated uncertainty.
Throughout our discussion, we highlight the strengths and limitations of these approaches in the context of pHHI. 
We conclude by indicating how intent estimates should be integrated with whole-body impedance, preview planning, and safety layers, setting the stage for the integration pathways developed in later sections.

\subsection{Definition of Human Intent: From Desired Trajectories to Cognitive States}
In the context of this review, ``intent" refers to the underlying variables that help to explain and predict a partner's future movements and force profile. 
This includes understanding where they are heading, what they intend to do with shared objects, and the strength of their actions. 
At the most basic operational level, intent can be represented as a desired trajectory, essentially, the desired positions, velocity profiles, or force profiles (wrench envelopes). 
If a robot knows these trajectories, it can adjust its interactions and anticipate movements, such as steps or grasps. 
At a higher level of task planning, intent may represent specific goals (e.g., turning left, stopping, or handing over an object) or role assignments (such as leader or follower) that influence how individuals collaborate. 
At the cognitive level, intent is associated with plans, preferences, and social cues, such as gaze and timing, which are critical to building trust and enhancing legibility in interactions. 
Research on shared control and intention-based systems shows that making intent explicit can improve interaction fluency and reduce effort. However, it is crucial to maintain transparency and calibrated autonomy for user acceptance~\cite{losey2018review, zhang2023integrating, hoffman2023inferring}.

It is important to note that in robotics, intent must often be inferred in situations of partial observability, under tight deadlines, and in the presence of disturbances caused by contact. 
This intent then needs to be translated into whole-body behaviors that ensure comfort and safety, a requirement that presents challenges for both perception and control~\cite{farajtabar2024path}.

\subsection{Sensing Modalities for Human Intent Estimation}
A humanoid's ability to understand intent is limited by its sensory capabilities and the latency of its sensory input. 
Different sensory channels reveal various aspects of a person's underlying plan. 
For example, body posture can provide information about geometry and timing, but it may become unclear if the view is obstructed. 
Additionally, the forces exerted during contact can indicate immediate goals but may be confused by environmental factors that affect human intent. 
Physiological signals and gaze direction can offer cognitive-level clues that appear earlier than the corresponding motion, but they require careful calibration and pose challenges in terms of reliability.

\subsubsection{Kinematic and Motion-Based Sensing: Leveraging Body Pose and Skeletal Data}
Kinematics is fundamental for intent inference in pHHI because an individual's pose and its changes over time provide strong indicators of near-future actions. 
Recent advances in markerless, multi-person tracking tailored for Human-Robot Interaction (HRI) have minimized jitter and occlusion artifacts, resulting in stable pose streams that downstream predictors can reliably use~\cite{martini2024robust}. 
When direct contact sensing is not available, dependable 3D pose baselines and monocular human dynamics estimation provide practical inputs for early intent decoding~\cite{martinez2017simple, yu2021human}. 
Using these pose streams, lightweight predictors can quickly identify imminent actions (e.g., a pivot during co-carrying) to facilitate timely adjustments in impedance or planning. 
In walking scenarios, intent recognition has been framed as a multi-class problem (continue, turn, stop), incorporating phase-aware features to trigger balance-sensitive assistance~\cite{lanini2018human}. 
Building on this foundation, vision-based skeleton tracking with RGB-D (i.e., images with both color (RGB) and depth (D) information) or monocular cameras offers a valuable, non-contact means of understanding posture, velocity, and configuration, supporting proactive behaviors. 
By observing the early stages of motion, the robot can forecast short-horizon trajectories~\cite{lyu20223d, usman2022skeleton}, allowing it to anticipate user needs before physical contact occurs. 
This capability enables proactive handovers and preemptive collision avoidance~\cite{strabala2013toward, admoni2014deliberate, mainprice2013human}. 
In pHHI, these predictions can be directly integrated into whole-body or impedance controllers, helping to shape interaction forces while maintaining balance.

However, limitations exist; kinematic data can be ambiguous without context related to objects or forces (the same motion may have different underlying goals), and close-quarter interactions can heighten issues like occlusions and sensitivity to lighting~\cite{lyu20223d, usman2022skeleton, townsend2017estimating}. 
To achieve cohesion and robustness in practice, several strategies could be implemented, such as tracking techniques (including temporal priors, kinematic constraints, and multi-view/depth systems) to stabilize pose and contact-aware features that align predictions with the timing of tasks.

\subsubsection{Force and Torque-Based Sensing: Direct Physical Cues}
Haptic measurements, specifically the interaction forces and torques at the contact point, provide the clearest insight into a partner's short-term intentions, particularly in co-manipulation scenarios. 
End-effector and joint force/torque sensors, or estimators derived from motor currents, effectively capture human push/pull patterns, desired direction changes, and load-sharing behaviors with high fidelity and low latency. 
This generates an immediately actionable signal for impedance/admittance and whole-body control systems~\cite{farajtabar2024path, al2021improving}. 
Learning from these signals enhances shared object manipulation by mapping force histories to necessary motion corrections and role allocations. 
In co-transport situations, combining recent kinematics with wrench data enables rapid estimation of desired velocity and force direction. 
Additionally, constraint-aware formulations ensure that inferred goals remain physically feasible for the human–object–robot system~\cite{shao2024constraint}. 
When dedicated sensors are not available, several studies demonstrate the recovery of virtual interaction wrenches from the robot's dynamics (specifically, torque residuals), retaining much of the functionality for intent decoding and compliance shaping~\cite{erden2010human, li2013human}.

While the strengths of this modality, such as immediacy and actionability, are significant, they also highlight its limitations. 
Pure force sensing is inherently reactive; the robot can only learn from the effort applied, which can result in sluggish responses or require unnecessary human exertion. 
Practical challenges include sensor placement, bias and noise, contact switches that complicate estimation, and the difficulty of separating human intent from environmental disturbances.
Cohesive design approaches help mitigate these issues by implementing filtering and wrench observers linked to the contact state or object state data to clarify intent.

\subsubsection{Physiological and Gaze-Based Sensing} 
Physiological signals and gaze patterns provide anticipatory evidence of human intent, often preceding overt motion.
Eye tracking reveals where a person's visual attention is focused and which targets they are selecting, strong predictors of the next reach, route, or grasp. 
This information enables actions such as pre-positioning, safer handovers, and earlier collision avoidance~\cite{belcamino2024gaze}. 
Research on handover timing and gaze indicates that well-timed cues can enhance compliance and perceived fluidity during interactions. 
These cues can be integrated into intent estimators to better coordinate role exchanges and initiatives during close contact~\cite{strabala2013toward, admoni2014deliberate}.
On the physiological side, ElectroMyoGraphy (EMG) captures muscle activation before any measurable force is generated. 
Additionally, wearable Inertial Measurement Units (IMUs) can segment activities, detect gait phases, and trigger short-term motion primitives that are generally applicable across users~\cite{taborri2016gait,xue2021using}. 
In pHHI, these early signals can guide whole-body and impedance controllers to anticipate the most likely action.

The advantages of these channels lie in their ability to reduce lead time and clarify tasks. 
However, there are practical and user-dependent challenges: eye trackers and EMG can be intrusive and require calibration. 
At the same time, signals can vary significantly between users, depending on factors such as placement and context.
Additionally, performance can degrade due to fatigue, perspiration, or changes in lighting and occlusions~\cite{su2023recent, qiu2022multi, wang2024multimodal}.
Consequently, cohesive designs treat these modalities as supporting evidence rather than definitive measurements. 
They fuse data from physiological signals, gaze, kinematic, and force information using probabilistic filters. 
 
\subsubsection{Multi-Modal Fusion}
No single sensor is reliable across the wide range of pHHI scenarios, so effective intent understanding and safe assistance increasingly depend on sensor fusion. 
Combining visual data with force and tactile cues consistently enhances continuous intent recognition and attention estimation compared to using a single modality, thereby reducing false triggers in tasks such as hand-guiding and shared carrying~\cite{wong2023vision}. 
Markerless tracking, tailored for HRI, which incorporates temporal priors, multi-view perspectives, and kinematic constraints, stabilizes the visual component in cluttered environments~\cite{martini2024robust}. 
Meanwhile, gaze information helps clarify which target a reach is directed at, while wearables like EMG sensors and IMUs provide phase and early activation cues. 
The outcome is the gathering of complementary evidence at various latencies and levels of abstraction: for instance, gaze can clarify the intended target of a reach, vision and kinematics may predict motion, and force measurements can confirm physical intent upon contact.


In practical applications, early cues (such as gaze, EMG, and IMU signals) serve as priors that inform predictions and control set points, while haptic feedback and kinematics provide quick posteriors to effectively close the control loop. 
The potential benefits of this approach are improved disambiguation and fault tolerance. 
However, challenges include managing latency and computational demands, as well as the necessity for reliability-aware attention and online calibration/personalization. 
Key open issues encompass adapting fusion weights based on different users and contexts, addressing sensor bias and drift during contact transitions, and ensuring comfort and legibility (for example, through rate-limiting updates to stiffness or wrenches) as beliefs evolve. 
A practical approach involves stabilizing vision, fusing gaze, EMG, and IMU data as intent priors, confirming with force and torque evidence, enforcing feasibility, and communicating confidence levels to the controller. 
This allows the system to balance between reactive compliance and proactive assistance smoothly~\cite{su2023recent, qiu2022multi, wang2024multimodal}.

\subsection{Models and Algorithms for Human Intent Prediction}
Once sensing collects information to expose parts of a human's plan, the main challenge is to transform these insights into actionable forecasts. 
Specifically, predicting what the human will do next, the timing and intensity of their actions, and our level of confidence in these predictions. 
Here, we discuss some of the models and algorithms most frequently used to capture human intent.

\subsubsection{Probabilistic Models: Hidden Markov Models (HMMs) and Bayesian Filtering}
Probabilistic intent modeling views human goals as latent stochastic processes that generate observed motion and force.
Classical HMMs and recursive Bayesian filters maintain a belief over candidate intentions or phases and update this belief online as new sensory evidence arrives (such as from vision, wrenches, gaze, and EMG). 
This process yields an interpretable posterior that controllers can use for risk-aware decisions~\cite{callens2020framework, townsend2017estimating}. 
On the trajectory side, probabilistic movement primitives (ProMPs) and Gaussian mixture or regression models encode distributions over motions. 
These models enable early action recognition and short-horizon forecasts from partial observations. 
Phase and speed estimators condition a motion library online for collaborative tasks, even with limited data~\cite{wang2013probabilistic,maeda2017probabilistic, callens2020framework}.

The advantages of these models are threefold: they are data-efficient, provide explicit uncertainty (calibrated confidence for safe decision-making), and facilitate reflex-rate inference that interacts with impedance, WBC, or MPC.
However, they also have some weaknesses: they depend on task- or library-specific motion sets, have limited prediction horizons, and can be sensitive to personalization, contact reconfiguration, and high-dimensional continuous dynamics.

\subsubsection{Partially Observable Markov Decision Processes (POMDPs)}
When intent is hidden and observations are noisy, the model naturally fits a Partially Observable Markov Decision Process (POMDP). 
In this framework, the human's intent is considered a latent state, while the robot maintains a belief over hypotheses derived from multimodal evidence. 
It chooses actions that both advance the task and reduce ambiguity. 
In scenarios like collaborative transport or handovers, this approach results in policies that gently probe the human's intent, such as slight object reorientation or compliant micro-motions, to encourage clarifying responses from the human while adhering to strict safety limits. 
This reflects findings on implicit communication through motion~\cite{yang2025implicit}.
POMDPs also formalize the concept of belief-dependent controllers, which can adjust their impedance, being softer when uncertainty is high and firmer when confidence is high. 
However, exact POMDP solutions are complex for humanoid robots, as beliefs are often high-dimensional and continuous. 
Therefore, systems often rely on approximations that retain many advantages, such as fast Bayesian filters to generate intent or phase beliefs~\cite{maeda2017probabilistic}. 
These filters are paired with belief-aware receding-horizon controllers that encode intent as constraints, costs, or scenarios~\cite{losey2018review}.

\subsubsection{Inverse Optimal Control (IOC) and Inverse Reinforcement Learning (IRL): Inferring Human Objectives}
Humans tend to act approximately optimally with respect to a hidden objective, and researchers in IRL seek to recover that objective from demonstrations.
This approach focuses on translating intent into a cost landscape rather than simply making kinematic predictions~\cite{shao2024constraint}. 
In the area of collaborative loco-manipulation, IRL has been effectively used to derive rewards that align with human preferences, such as stability, smoothness, and effort~\cite{wu2024infer}. 
This results in robot behaviors that users perceive as more natural and comfortable~\cite{huang2023conditional}.
Another perspective views physical interaction as a form of communication, where the robot updates its objective in real-time based on human feedback. 
This co-active learning process helps the robot adapt its strategies to reflect the user's implicit goals during the task~\cite{losey2022physical}. 
The semantics-first approach to intent in pHHI facilitates generalization to new contexts, whether involving unfamiliar objects, paths, or role distributions. 

However, this approach presents challenges. 
Many different objectives can explain the same motion (a problem known as a lack of identifiability), and the features involved are important. 
Additionally, data collection can be limited, and full IOC and IRL can be computationally intensive, particularly under conditions of partial observability and complex, contact-rich, whole-body dynamics. 
To address these challenges, practical strategies include: structuring rewards using physically meaningful features, such as ergonomics, effort, and contact safety, accompanied by regularization constraints.
Combining offline IOC with online, co-adaptive updates based on corrections made during tasks to enhance personalization.

\subsubsection{Machine Learning and Deep Learning Approaches}
More recently, deep learning has become the dominant approach, offering powerful tools for learning complex patterns from large datasets without the need for extensive feature engineering.

\paragraph{Recurrent Neural Networks (RNNs, LSTMs, GRUs) for Temporal Dynamics}
Recurrent Neural Networks (RNNs), particularly Long Short-Term Memory (LSTM) networks and Gated Recurrent Unit (GRUs), effectively utilize the sequential structure in pose, wrench, and gaze data streams~\cite{liu2021robot}. 
This enables them to recognize intent and forecast short-term movements more accurately than traditional methods~\cite{gao2021hybrid}. 
In the context of pHHI, these models support early intent detection (such as turning, stopping, or continuing while walking) by employing gait-phase-aware classifiers. 
These classifiers can trigger balance-sensitive assistance and control preview based on detected intent~\cite{lanini2018human}.
RNNs are trained on sliding windows of data and deployed in real-time, providing low-latency predictions. 
These predictions can be integrated into impedance, momentum, or WBC and QP controllers, allowing for pre-alignment of frames, biasing of stiffness in the predicted direction, and scheduling of stepping or handover timings. 
This results in proactive yet easily interpretable assistance.

Their advantages include strong temporal expressiveness, moderate computational requirements, and ease of integration into real-time systems. 
However, they also exhibit some weaknesses, such as sensitivity to dataset biases and domain shifts, performance drift over longer prediction horizons, and instability during phase or contact transitions when there is no explicit structural guidance.
To create cohesive designs, the following strategies are recommended: 
First, incorporate phase and contact information (e.g., phase channels or event flags) to stabilize transitions. 
Second, pair RNNs with physics-based priors (e.g., Probabilistic or Dynamic Movement Primitives), which allows the network to learn residuals. 
Third, implement multi-task heads (for classification and trajectory prediction) with curriculum or scheduled sampling to reduce compounding errors.

\paragraph{Transformer Models: Conditioning on Robot Actions for Enhanced Prediction}
Transformers use attention mechanisms to capture long-range dependencies and make robust early predictions based on partial histories. 
They have been employed to anticipate human actions in collaborative settings~\cite{zhang2024early}, predict forces and velocities from video during co-transport, allowing robots to proactively adapt to load changes~\cite{dominguez2024exploring,dominguez2024force}, and tokenize wearable IMU streams into motion primitives that generalize across different users, yielding latent ``intent tokens" that interface smoothly with control systems~\cite{zhang2025mopformer}. 
Generative models provide a way to achieve more human-like and diverse motion prediction. 
Rather than predicting a single future trajectory, models such as Variational Autoencoders (VAEs) and Generative Adversarial Networks (GANs) can learn a distribution of plausible future motions~\cite{xiang2024socialcvae,barsoum2018hp}. 
This capability enables robots to consider multiple potential human actions. 
The strength of these models lies in their expressiveness and their ability to capture the multi-modal nature of human behavior.
However, they face challenges related to controlling the generation process to align with specific task constraints and ensuring the stability of their training. 
Compared to recurrent models, the attention mechanism connects distant cues (for example, an early gaze or pose shift to a later force peak). 
Thus, enhancing short-horizon forecasts that can be utilized by whole-body or impedance controllers for timing, stiffness adjustments, and safe anticipatory actions. 
A crucial area of development is conditioning on the robot's intended actions and compliance state. 
By inputting the robot's future motions (or candidate plans) alongside human signals, the predictor can address the question, ``\textit{What will the human do in response to this robot motion?}''. 
This approach reduces ambiguity and produces consistent forecasts, which are better suited for shared autonomy, such as anticipating the timing of handovers, adjusting foot placements during co-carrying, or facilitating proactive load sharing. 

The practical challenges of this approach are well recognized: transformers are data-hungry and compute-intensive, and conditioning on high-dimensional robot states while ensuring real-time performance and calibrated uncertainty remains complex.
Emerging strategies include pairing lightweight transformer heads (such as streaming or sparse attention and sliding windows) with distilled or frozen backbones. 
Additionally, pretraining on simulations or logs with domain adaptation, exposing uncertainty for assistance gating, and implementing a certified execution layer underneath can help manage residual model errors.


\paragraph{Generative Models (VAEs, GANs, Diffusion Models) for Expressive Motion and Scene Understanding}
Generative models provide a way to achieve more human-like and diverse motion prediction. 
Rather than predicting a single future trajectory, models such as Variational Autoencoders (VAEs) and Generative Adversarial Networks (GANs) can learn a distribution of plausible future motions~\cite{dermy2018prediction}. 
This capability enables robots to consider multiple potential human actions~\cite{lyu20223d}. 
The strength of these models lies in their expressiveness and their ability to capture the multi-modal nature of human behavior. 
However, they face challenges related to controlling the generation process to align with specific task constraints and ensuring the stability of their training.

\begin{figure}
    \centering
    \includegraphics[width=0.7\linewidth]{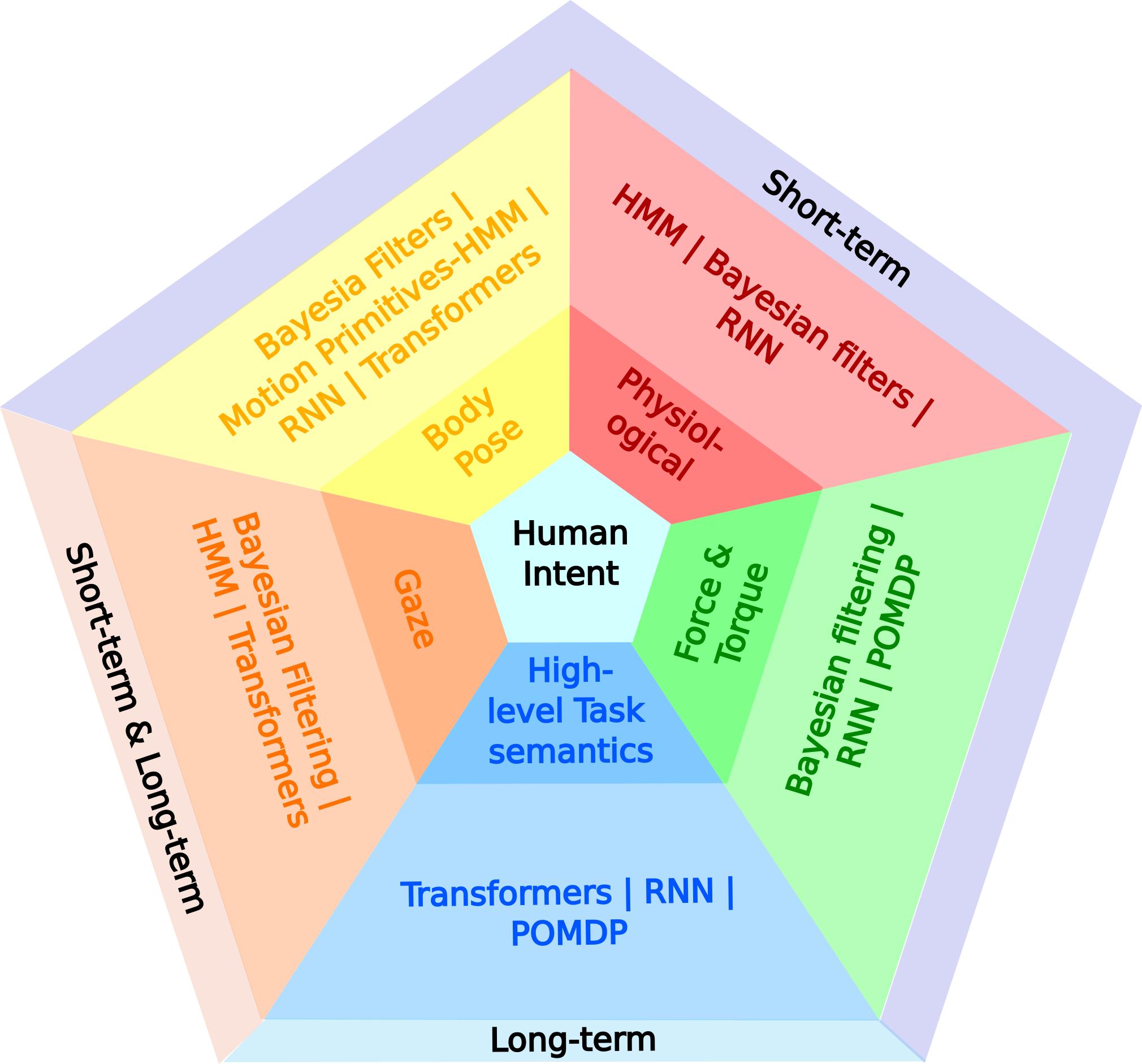}
    \caption{A map of human intent modalities, where the outer layer lists the most commonly used methods for each signal type, organized by their effective time horizon. Short-term physical cues (e.g., body pose, force/torque) are matched with reactive models, while long-term semantic cues (e.g., gaze, task goals) are paired with predictive, goal-aware algorithms.
    }
    \label{fig:pillar2_st-lt}
\end{figure}


\begin{table*}[htbp]
\centering
\footnotesize
\setlength{\tabcolsep}{2.pt}
\caption{A comparative analysis of intent estimation algorithms, highlighting their primary sensing requirements, output types, and key trade-offs.}
\label{tab:pillar2}
\begin{tabularx}{\textwidth}{
  >{\RaggedRight\arraybackslash}p{2.2cm}   
  >{\RaggedRight\arraybackslash}X  
  >{\RaggedRight\arraybackslash}X  
  >{\RaggedRight\arraybackslash}X  
  >{\RaggedRight\arraybackslash}X  
  >{\RaggedRight\arraybackslash}p{1.3cm}  
}
\toprule
\textbf{Sensing / Scope} & \textbf{Model / Algorithm} & \textbf{What is predicted/estimated} & \textbf{Advantages} & \textbf{Limitations / Risks} & \textbf{Rep. Refs.} \\
\midrule

\multicolumn{6}{>{\RaggedRight\arraybackslash}X}{\bfseries\itshape Sensoring and Fusion} \\
Multi-modal (vision, wrench/FT, tactile, gaze, EMG/IMU) &
Fusion (stabilized vision, gaze/EMG/IMU priors; force/kinematics posteriors) &
Intent priors/posteriors; contact events; complementary cues at different latencies &
Improves disambiguation and fault tolerance; clarifies targets and timing &
Latency/compute management; sensor bias/drift at contact transitions; need personalization &
\cite{hoffman2023inferring, belcamino2024gaze, su2023recent, qiu2022multi, wang2024multimodal}\\
\midrule

\multicolumn{6}{l}{\bfseries\itshape Probabilistic and Decision-Theoretic Models}\\
Pose / force / gaze / EMG streams &
HMMs \& Bayesian filtering &
Belief over goals/phases; early action recognition; short-horizon traj./force forecasts &
Data-efficient; explicit uncertainty for safe decisions; reflex-rate inference &
Library/task-specific motion sets; limited horizons; sensitive w.r.t. personalization/contact reconfig/high-dim dynamics &
\cite{townsend2017estimating,wong2023vision, callens2020framework, wang2013probabilistic, maeda2017probabilistic}\\

Belief over hidden intent under uncertainty &
POMDPs &
Belief state + probing actions to reduce ambiguity; impedance adjusted by confidence &
Formalizes belief-aware control; gentle probing (micro-motions) consistent with implicit communication &
Exact solutions intractable (continuous, high-dim beliefs); use approximations (fast filters + belief-aware MPC) &
\cite{yang2025implicit, losey2018review}\\
\midrule

\multicolumn{6}{l}{\bfseries\itshape Objective Inference and Learning} \\
Demonstrations / human feedback &
IOC / IRL &
Reward/cost aligned with preferences (stability, smoothness, effort); objective updated online &
Behaviors perceived as natural/comfortable; semantics-first aids generalization &
Identifiability; limited data; computational cost under partial observability \& contact-rich dynamics &
\cite{shao2024constraint, wu2024infer, huang2023conditional, losey2022physical}\\

Sequential multimodal data &
RNNs / LSTMs / GRUs &
Early intent detection; short-term movement forecasts; gait-phase-aware triggers &
Better accuracy than classical baselines; low-latency sliding-window inference &
Data hungry; distribution shift; needs calibrated uncertainty &
\cite{liu2021robot, gao2021hybrid, lanini2018human} \\

Sequences, video, IMUs (robot-conditioned) &
Transformers (action-conditioned); intent tokens &
Anticipate actions; predict forces/velocities from video; tokenize IMUs into motion/“intent” tokens &
Capture long-range dependencies; improved short-horizon forecasts when conditioned on robot plans &
Compute/data intensive; real-time + calibrated uncertainty is challenging &
\cite{zhang2024early, dominguez2024exploring, dominguez2024force}\\

Trajectory/scene futures &
Generative models (VAEs, GANs, Diffusion) &
Multi-modal future samples (plausible human actions) &
Expressive; captures multi-modal human behavior &
Training stability; aligning generation with task constraints &
\cite{zhang2025mopformer, lanini2018human} \\

Demonstrations; large datasets; teleop &
Imitation Learning / RL; foundation policies; language + teleoperation &
Policies that compile/adapt whole-body skills; few-shot adaptation &
Rapid adaptation; clear shared autonomy with teleop; leverage diverse prior data &
Scale / interpretability; avoid hallucination; LLM safety (jailbreaking) &
\cite{cheng2024expressive, ji2024exbody2, fu2024humanplus, liu2024human, kumar2023words, jiang2024harmon, yao2024anybipe, sun2024prompt, he2024omnih2o, robey2024jailbreaking, zhang2024badrobot}\\
\bottomrule
\end{tabularx}
\end{table*}

\subsection{Pillar II: Insights and Discussion}
Practical human intent estimation requires a careful matching of sensing modalities and computational models to the specific demands of the interaction.
Figure~\ref{fig:pillar2_st-lt} considers the discussed intent modalities (Body pose, physiological, Force and torque, high-level task semantics, and gaze) and classifies them for both short-term and long-term intentions. 
Additionally, it highlights the more common methods and algorithms that can be utilized more effectively for these modalities. 
The core challenge lies in bridging different time horizons: from immediate, reflex-rate predictions that guide reactive control to long-term, goal-oriented forecasts that enable proactive assistance.
Then, Table~\ref{tab:pillar2} classifies the sensing methods or scopes discussed in this section, describes what is estimated, and outlines their advantages and limitations. 

The selection of sensing modality is closely related to the temporal characteristics of intent. 
Short-horizon intent, including next-step prediction or minor force adjustments, is effectively captured by high-bandwidth modalities such as body pose and force or torque measurements. 
These data streams deliver phase-specific information suitable for lightweight, uncertainty-aware models, including HMMs, Bayesian filters, and Probabilistic Movement Primitives (ProMPs). 
These approaches provide rapid controller updates, supporting impedance modulation and whole-body controller previews at reflex rates. 
For low-latency sequence prediction, RNNs are mainly preferred since they efficiently generate short-term forecasts from these dense data streams.

Long-horizon intent estimation focuses on inferring task-level goals such as target selection or role arbitration and relies on modalities that provide rich semantic information. 
Gaze serves as an early indicator of human focus and anticipated objectives, while high-level task semantics establish the context for interaction. 
Sophisticated computational models are required to interpret these cues. Methods, including IOC and IRL, are designed to infer underlying objectives, while POMDPs explicitly represent beliefs and uncertainty. 
Recently, Transformer architectures have demonstrated strong performance by using attention mechanisms to learn long-term, goal-oriented dependencies from sequential data. 
These semantic-based models are effective for generalizing to novel tasks because they capture the rationale behind actions rather than only surface-level features. 
As a result, they generalize more reliably across different objects and environmental configurations than approaches based solely on trajectory matching.

Robust performance across a diverse user population requires system customization. 
Physiological signals, such as EMG and IMU data, exhibit significant user specificity in intent predictions, emphasizing the need for personalization. 
For these signals and for gaze data, per-user calibration is typically necessary. 
Techniques such as few-shot normalization or online adaptation can further improve predictive accuracy. 
Personalization in modeling may involve adding lightweight residual adapter layers to RNN or Transformer models to reflect individual timing and stiffness preferences. 
An effective and generalizable system should integrate multiple sensing modalities, utilizing early indicators from gaze or EMG as priors and confirming predictions with late-arriving posteriors from pose and force measurements.

\section{Pillar III: Human Models}
\label{sec:pillar3:human_model}
The first two pillars enable the robot to control its body and infer its partner's goals. 
However, to elevate interactions from merely functional to truly safe, efficient, and intuitive, the robot needs a deeper comprehension of the entity it is interacting with. 
This requires a computational model of the human, essentially a framework for understanding the physical principles that govern human movement, capabilities, and limitations~\cite{fang2023human}. 
This third pillar goes beyond treating humans as generic sources of intent signals. 
Instead, it models them as complex cognitive biomechanical systems~\cite{hiatt2017human}. 
The remainder of this section reviews human models across various categories, including musculoskeletal and physics-based bodies, reflex and motor-control descriptions, data-driven motion priors, and cognitive/decision abstractions.
\subsection{Musculoskeletal and physics-based models}
To effectively reason about a human partner beyond mere surface kinematics, pHHI could benefit from models that expose the body's internal mechanics, ranging from muscles to whole-body dynamics~\cite{maurice2019digital, magistris2015dynamic}. 
At the high-fidelity level, OpenSim offers reusable musculoskeletal models, estimation tools, and validation pipelines for neuromuscular control and movement analysis~\cite{seth2018opensim}. 
Notably, comprehensive full-body models, such as a gait model, incorporate numerous skeletal Degrees of Freedom (DOFs) and extensive sets of muscle actuators. 
These models generate joint kinetics and muscle activations that align with experimental data, establishing them as standard references for analysis and controller-in-loop studies~\cite{rajagopal2016full}.

Conversely, at a more abstract but faster level, rigid-body human models with joint limits enable real-time whole-body simulation for virtual reality, teleoperation, and control-theoretic analysis~\cite{su2023recent, prendergast2021biomechanics}. 
This makes them practical interfaces for robot planners and safety filters~\cite{liu2021computational}. 
These physics frameworks seamlessly connect to humanoid balance. 
Empirical evidence shows that humans regulate whole-body angular momentum during walking, which provides principled targets and constraints for momentum-aware collaboration and shared load handling~\cite{herr2008angular}. 
Complementary interactive locomotion studies, where two people physically pair while walking, illustrate how coupling forces, cadence, and phase coordination develop, offering datasets and dynamical patterns that can be emulated or anticipated in human-robot interactions~\cite{lanini2017interactive}.

For pHHI, the benefits of these models are twofold. 
First, musculoskeletal models can predict unmeasured internal variables, such as muscle forces and joint contact loads. 
This enables robots to plan contact behaviors that minimize discomfort or risk and to incorporate ergonomic costs directly into their assistance policies. 
Second, physics-based human dynamics offer physically plausible motion priors under perturbations, which enhance short-horizon posture forecasts and inform balance and momentum controllers with human-consistent constraints, such as allowable angular momentum changes. 
The main challenges in this area are personalization and latency. 
Accurately identifying an individual's bone geometry, soft tissue, and muscle properties requires a labor-intensive process, making it challenging to tailor models efficiently. 
Even when such models are developed, forward musculoskeletal simulations are rarely fast enough to run in real time, further limiting their practical use.

\subsection{Reflex and motor-control models}
Pure biomechanics alone cannot fully explain how people adapt during contact. 
However, neuromechanical models help bridge that gap by linking reflex pathways and task-level feedback to observed motion. 
Studies have demonstrated that compact muscle-reflex loops, which are tuned to legged mechanics, can accurately reproduce human walking kinematics and EMG patterns. 
This provides evidence that reflexes not only stabilize gait but also redistribute effort across different phases~\cite{geyer2010muscle}.
Reflex gains are also dependent on the task at hand. 
For instance, humans tend to enhance ``position-like'' feedback when focusing on positional goals and shift toward ``force-like'' feedback for tasks involving force tracking~\cite{mugge2010rigorous}. 
This suggests that our reflexes can adapt flexibly in response to context. 
In the upper limb, endpoint stiffness is described by a stiffness ellipsoid, whose orientation depends on posture and whose size is influenced by co-contraction. 
This creates reduced-order maps that link configuration and activation to interactive compliance~\cite{burdet2000method, ajoudani2018reduced, patel2014effect}. 
Together, these models allow us to interpret posture, phase, and activation as actionable insights into how a partner might respond during contact.

In pHHI, these insights translate directly into compliance scheduling. 
If the posture and phase suggest a low-stiffness human limb or whole body, the robot should absorb more disturbances and adjust its admittance accordingly. 
Conversely, if the human is bracing (indicating higher co-contraction), the robot can share more of the load or increase its impedance without causing unexpected reactions.
These cues can also guide momentum and whole-body controllers, ensuring that interaction forces are predictable and maintaining stable balance.
However, the challenges are expected: there is significant variability both between different individuals and within the same individual, and estimating internal states (like co-contraction) in real-time without wearables is difficult.

\begin{table*}[htbp]
\centering
\footnotesize
\setlength{\tabcolsep}{2pt}
\caption{Computational Human Models for pHHI (Pillar III): exports to control, required inputs, strengths, limitations, and representative references.}
\label{tab:pillar3}
\begin{tabularx}{\textwidth}{
  >{\RaggedRight\arraybackslash}p{3.cm}  
  >{\RaggedRight\arraybackslash}X         
  >{\RaggedRight\arraybackslash}X         
  >{\RaggedRight\arraybackslash}X         
  >{\RaggedRight\arraybackslash}X         
  >{\RaggedRight\arraybackslash}p{1.2cm}  
}
\toprule
\textbf{Model family (Pillar III category)} & \textbf{Exports to controller (constraints / costs / targets)} & \textbf{Required inputs \& parameters} & \textbf{Strengths / when to use} & \textbf{Limitations \& risks} & \textbf{Rep. refs} \\
\midrule
Musculoskeletal \& physics-based bodies 
& Joint-load limits; feasible wrench/contact sets; centroidal momentum targets; effort/comfort costs 
& Link properties; contact assumptions; human morphology; physics model configuration 
& Physically grounded limits / targets; informative for load sharing and balance-aware co-manipulation 
& Personalization and latency; model mismatch; forward simulations can be slow 
& \cite{seth2018opensim, rajagopal2016full, prendergast2021biomechanics, herr2008angular, lanini2017interactive} \\
\addlinespace
Reflex \& motor-control (neuromechanical) 
& Endpoint stiffness ellipsoids; impedance / feedback targets; task-dependent “position-like/force-like” gains 
& Posture / phase; activation/co-contraction cues; reduced-order stiffness maps 
& Directly informs compliance scheduling and whole-body shaping for safe, predictable interaction 
& Inter-/intra-person variability; challenging real-time estimation without wearables 
& \cite{geyer2010muscle, mugge2010rigorous, burdet2000method, ajoudani2018reduced} \\
\addlinespace
Data-driven motion priors 
& Fast predictors of human motion/forces; priors for costs/constraints and short-horizon targets 
& Datasets/logs (pose, wrench, video); trained generative/sequence models 
& Real-time surrogates that compress complex dynamics; multi-modal forecasts 
& Distribution shift; need calibrated uncertainty for safe use 
& \cite{bowden2000learning, tevet2022human, peng2018deepmimic, hassan2023synthesizing}\\
\addlinespace
Cognitive/decision abstractions 
& Preference-weighted costs; “fairness” and comfort/effort-sharing objectives 
& User preferences, task goals, comfort envelopes, and summaries passed as constraints/weights 
& Aligns robot behavior with user comfort and task goals; supports arbitration at the planning level 
& Ambiguous trade-offs; requires careful calibration of preferences 
& \cite{fang2023human, hiatt2017human, liu2021humans, li2024large} \\
\bottomrule
\end{tabularx}
\end{table*}

\subsection{Data-driven human motion models}
A complementary approach to understanding human motion involves learning directly from data, evolving from early statistical models to modern deep generative methods. 
Classic techniques fittingly employ low-dimensional manifolds and Markovian transitions based on motion capture data, resulting in straightforward predictors and synthesizers of likely motion patterns~\cite{bowden2000learning}. 
Contemporary models, however, enhance both expressivity and robustness. 
For instance, diffusion models can generate diverse and natural motion sequences based on text or action prompts~\cite{tevet2022human}. 
Additionally, HuMoR~\cite{rempe2021humor}, a conditional variational prior, effectively stabilizes pose tracking and prediction even in the presence of visual disturbances and occlusions.
DeepMimic~\cite{peng2018deepmimic}, for example, learns controllers that replicate a broad spectrum of human skills in simulation while maintaining credibility despite disturbances. 
Beyond isolated movements, human-scene interaction models incorporate contacts and affordances, enabling the learning of behaviors limited by environmental constraints that respect object dynamics~\cite{xiao2023unified, wang2023physhoi, hassan2023synthesizing}.

Recent research has begun to merge statistical methods with physics-based approaches to enhance both temporal consistency and contact realism. 
This integration is particularly promising for interaction forecasting, where contact plays a critical role in determining dynamics. 
Finally, behavioral studies show that humans often seek to minimize task-specific costs, which supports the use of learned or handcrafted objective structures to improve prediction and planning~\cite{liu2021humans}.

\subsection{Cognitive and decision models}
Beyond the physical execution of movement, a comprehensive human model must also account for the cognitive processes that drive behavior. 
Cognitive and decision models aim to capture how humans perceive, reason about, and make choices within collaborative tasks~\cite{hiatt2017human}. 
This includes modeling how a human mentally represents a task, often as a hierarchy of goals and sub-goals. 
By maintaining a belief over the human's current position in this task plan, the robot can better predict not just the next movement, but the next logical step in the overall process.

Furthermore, these models can incorporate principles from decision theory to understand how humans make choices under uncertainty or trade off between competing objectives, such as speed and accuracy. 
The emerging frontier in this domain is the use of LLMs as a proxy for human cognitive reasoning~\cite{li2024large}. 
By leveraging the world knowledge and common-sense reasoning embedded in LLMs, a robot can infer a human's high-level plan from sparse instructions or contextual cues, enabling a more proactive and semantically aware form of collaboration. 
The primary challenge is grounding these abstract cognitive models in the robot's real-time sensory data and ensuring their reasoning is both safe and contextually appropriate.

\subsection{Pillar III: Insights and Discussion}

Pillar III translates human factors into controller-ready constraints and costs, turning ergonomics into a live interface the robot can trust. 
It exports feasibility sets, such as joint-load limits and admissible wrenches, along with comfort/effort costs and task targets, so the robot can share work while staying within human limits. 
In our framework, the pillar is anchored by ergonomics, fidelity, and personalization, and it complements control (Pillar I) and intent (Pillar II) by supplying constraints and objectives that remain consistent across the stack.
Table~\ref{tab:pillar3} summarizes the section organization by relevant model families, what they export to control, the inputs they require, and the strengths and limitations.

Choosing the right level of fidelity is central to making this interface useful. 
High-fidelity musculoskeletal models (e.g., OpenSim-style bodies) expose internal variables, such as muscle forces and joint contact loads, which are invaluable for ergonomic costs and safe load sharing. 
However, they can be slow and sensitive to modeling assumptions.
Equally important, the human model must compute in real-time so that the exported feasibility sets and costs accurately reflect the person's shifting state. 
Relying on outdated estimates reduces the quality of assistance and can violate comfort or safety limits, so the human model must update at control-relevant rates while intensive inference operates in the background.
In contrast, rigid-body human models run in real time and pair naturally with balance-aware planners, although with less physiological detail. 
A pragmatic strategy is to ``budget” fidelity: deploy physics-grounded limits and targets online (e.g., feasible wrench envelopes, centroidal momentum targets), while reserving full musculoskeletal inference for offline calibration or slower background updates. 
This division captures the benefits of load sharing while acknowledging the risks associated with model mismatch and forward-simulation latency.

Because people vary both across users and over time, personalization must shift from population averages to individual profiles. 
To support this, per-user calibration of upstream signals (EMG, IMU, gaze) enhances downstream predictions and should be a routine practice, whether achieved through few-shot normalization or lightweight online adaptation.
In turn, structural personalization exposes individualized feasibility and cost parameters to the controller: endpoint stiffness ellipsoids, joint-load limits, and effort/comfort terms that reflect each partner's capabilities. 
With these in place, Pillar I controllers can re-weight assistance and impedance on the fly, aligning help with the person's current state rather than an average model.

\section{Discussion, Challenges, and Future Directions for pHHI}
\label{sec:discuss_future}
This review has examined the state of the art in pHHI through three foundational pillars: 
(i) humanoid modeling and control, 
(ii) human intent estimation and prediction, and 
(iii) computational models of human behavior.
This approach attempts to tackle three challenging problems. First, the control of a floating, hybrid, multi-contact body in real time. Second, the inference of a human goal that is inherently latent and time-varying, all under conditions of partial observability. Finally, the reasoning about the human partner not merely as a source of signals, but as a biomechanical system with its own limitations, preferences, and sensitivities to risk.
While significant advancements have been made within each of these pillars, the field has yet to achieve the seamless, intuitive, and robust interaction envisioned for the widespread deployment of humanoid robots in human-centric environments. 
This section discusses the primary challenges hindering progress, explores the reasons behind the difficulties in creating a common framework, and outlines future directions and research paths that could lead to breakthroughs in the field.

\subsection{Principal Identified Challenges in pHHI}

The development of seamless and robust pHHI is hindered by a set of deeply interconnected challenges that span multiple temporal, abstract, and interaction scales. 
These are not isolated algorithmic gaps but rather a reflection of the fundamental difficulty of creating predictable, safe, and effective physical collaboration between a robot and an adaptive, uncertain human partner. 
The following sections elaborate on the principal identified challenges that the field should currently overcome.

\subsubsection{Time-Scale Management for Real-Time Operations}
pHHI requires the integration of subsystems that operate on significantly different time scales. 
High-frequency control loops for torque and impedance must operate at rates of thousands of hertz, while mid-level planners for locomotion and multi-contact maneuvers operate at rates ranging from tens to a few hundreds of hertz. 
In contrast, intent estimators typically provide information at slower rates, such as those from video or wearable sensors, and important human biomechanical factors, such as fatigue and comfort, change over timescales of seconds to minutes.
A significant challenge is designing multi-rate architectures that can maintain stability and effectiveness when high-level estimates are received intermittently or with varying latencies. 
An effective system must intelligently determine which variables can be reliably controlled at specific rates and implement well-considered fallback strategies. 
These strategies may include tightening safety constraints or softening impedance in response to delayed predictions, ensuring that the robot's actions remain safe and predictable for the human partner.

\subsubsection{Dynamic Abstraction and the Curse of Dimensionality}

The current model-based frameworks for humanoids involved in pHHI follow a hierarchical approach, dividing it into two tasks: high-level planning using abstracted models and low-level controllers that execute these high-level plans on the humanoid. 
The motivation for such abstracted models comes from the computationally vast state-decision space, which includes body pose, contact modes, footstep sequences, and grasp wrenches. 
These abstractions are often overly simplified. 
For example, authors in~\cite{KumbharArtemiadis2025MPCQP} extend the LIP models to account for interaction, but this typically comes at the cost of fidelity. 
As a result, notable discrepancies arise between the abstracted model and the actual system dynamics.
This often results in motion plans that are inefficient for the task. 
For instance, they may struggle to achieve compliance with the co-transported object and fail to fully exploit the humanoid's capabilities. 
For model-based controllers, there is thus a pressing need for abstractions of the coupled dynamics in pHHI that are both computationally efficient and more faithful to the underlying system.
Data-driven and learning-based approaches offer promising avenues for developing such abstractions often referred to as \emph{latent representations}, but they remain largely unexplored. 
Once established, these abstractions can be integrated into model-based methods, such as MPPI, or used within model-based reinforcement learning frameworks.


\subsubsection{Heterogeneity of Interactions and Autonomy Modes}
The scope of pHHI is extensive, encompassing both direct and indirect interactions, as well as shared autonomy in complex tasks. 
Each type of interaction presents its own unique challenges regarding observability, potential failure modes, and social expectations. 
Therefore, a common framework should focus on standardizing the interfaces between components, defining terms such as ``intent belief," ``human feasibility model," and ``safety invariant", while also allowing for mode-specific implementations of sensors, predictors, and constraints.

\subsubsection{Assurance, Interpretability, and the Sim-to-Real Gap for Learned Components}
Data-driven methods have become essential in all three pillars of pHHI. 
However, interactions that involve contact can be challenging, and infrequent prediction errors can lead to significant consequences. 
Providing end-to-end safety assurances when learned modules are involved remains an unresolved issue. 
The sim-to-real gap further complicates this problem, as simulators often struggle to accurately model the intricacies of contact physics and the variability of human behavior. 
Finally, the interpretability of these learned modules is crucial for humans to trust and effectively collaborate with robots.

\subsubsection{Missing datasets for collaborative tasks}
Progress in pHHI is significantly limited by the absence of large, standardized datasets that capture whole-body, bidirectional human–robot interaction. 
Most existing datasets focus on vision or kinematics for individual agents, providing limited contact information, sparse or noisy force and torque signals, minimal intent annotation, restricted task diversity, and homogeneous participant groups. 
These limitations hinder fair benchmarking and impede the development of learning methods that require comprehensive supervision to model intent, predict contact transitions, ensure safety and comfort, and generalize across different partners and tasks. 
An effective dataset should be multimodal and time-synchronized, incorporating motion capture or IMUs, joint states, interaction wrenches, tactile or pressure data, gaze, voice, EMG (where relevant), and environment geometry. 
It should also include explicit task and goal labels, partner intent (such as scripts, self-reports, or latent annotations), and cover core pHHI tasks, including handover, co-manipulation, balance assistance, locomotion support, and collaborative assembly. 
Additionally, the dataset should record safety and comfort outcomes as well as rare events such as near-collisions and slips to facilitate risk-aware learning. 
Without these comprehensive datasets, methods remain tailored to specific cases and are difficult to compare.

\subsubsection{Metric for collaboration: Stability, Efficiency, and Safety}
Most pHHI papers compare systems using task-specific surrogates, such as stability via ZMP/CoP or capture-point margins, and ``time in contact constraints''. 
Efficiency metrics include task time, path/pose error, success rate, interaction effort, assistance ratio, and sometimes human cost. 
Safety and comfort are measured by contact force peaks/rates, slip or over-torque counts, minimum separation, collision counts, jerk, and user studies. 
These metrics allow basic comparisons but are not standardized, often mask trade-offs, and rarely assess intent alignment, tail risks, co-adaptation, or user diversity. 
Stronger evaluation should: 
i) report multi-objective fronts (stability, efficiency, safety); 
ii) include intent-aware metrics (intent-prediction accuracy or lag, mutual information between robot and human goal); 
iii) use risk-sensitive statistics (CVaR of forces, minimum stability margin, barrier-function violations); 
iv) quantify interaction work and role-allocation smoothness; 
and v) standardize tasks/datasets with disturbances, and anthropometric diversity.
This approach will help clarify where each method provides added value.

\subsection{Pathways Towards Seamless pHHI: A Modular Approach}
Addressing the challenges outlined earlier requires a shift in approach from developing individual components in isolation. 
Achieving seamless interaction depends on creating unified frameworks that integrate the pillars. 
We propose an interaction-centric architecture that clearly defines the roles and interfaces, or ``contracts,'' between these pillars, enabling them to work together as a cohesive whole. 
This architecture is conceptualized as a multi-layered system, where each layer builds upon the guarantees and information provided by the layer beneath it.

\subsubsection{The Foundation: Safety-Certified Controller (Pillar I)}
At the lowest level, the framework should be built upon a fast controller that guarantees both the robot's physical integrity and the safety of humans at all times. 
This layer must be more than just a trajectory tracking module, it should serve as the ultimate authority for ensuring safe physical interactions.
\paragraph{Core Functionality} 
The foundational controller should comprise a hierarchical WBC that respects physical constraints such as friction cones and actuation limits. 
This must be augmented with adaptive impedance or admittance control at the interaction contact points, allowing the robot to be both firm and compliant as needed~\cite{caron2018capturabilitybased, caron2018balance, dai2014whole, fu2024humanplus, dominguez2024exploring,dominguez2024force}.
\paragraph{Interface Contract}
This layer is essential since it enforces a set of formally verified invariant sets that operate independently of high-level commands. 
This ``safety shield'' utilizes tools such as CBFs and stability criteria based on the ZMP or the DCM. 
These mechanisms ensure that the robot never enters an unrecoverable or dangerous state~\cite{sartore2024automatic,kumar2023words, jiang2024harmon, lanini2018human, li2024large, dermy2018prediction}.

\subsubsection{The Core: Dynamic Human Feasibility and Cost Modeling (Pillar III)}
At the core of the architecture should be a dynamic and continuously updated computational model of the human partner. 
This model should not be static but provide a real-time representation that personalizes the interaction.
\paragraph{Core Functionality}
This layer estimates and represents the human's physical state, including biomechanical properties, physiological status (e.g., fatigue), and ergonomic comfort.
\paragraph{Interface Contract (Human Feasibility \& Cost)}
The system provides the higher-level planner with specific limits and preferences tailored to each individual. 
This includes estimated endpoint stiffness ellipsoids, joint load limits, and cost functions that assess effort or discomfort. 
These parameters are utilized to align the robot's behavior with the capabilities and comfort level of the human user~\cite{monari2024physical, tevet2022human, rempe2021humor, xiao2023unified, geyer2010muscle, rajagopal2016full}.

\subsubsection{The Brain: Predictive and Uncertainty-Aware Intent Estimation (Pillar II)}
This cognitive layer interprets human actions and predicts future intentions, aiming to transition the robot from a reactive to a proactive stance.
\paragraph{Core Functionality}
It should integrate various sensory data, such as human pose, gaze, and interaction tools, to infer human goals. 
This can be accomplished using models like RNNs, Transformers, or Probabilistic Movement Primitives~\cite{keemink2018admittance, koppula2015anticipating, lanini2017interactive, seth2018opensim, liu2021computational}
\paragraph{Interface Contract (Intent Belief)}
It generates a streaming, calibrated belief about human intent. 
This is not a single point estimate, but rather a probability distribution over future trajectories, desired force envelopes, and discrete goals. 
This calibrated uncertainty is crucial for making informed, robust decisions.

\subsubsection{The Apex: Belief-Aware Shared Autonomy and Planning}
At the highest level, a planner or arbiter integrates information from the lower layers to make decisions. 
It orchestrates collaboration, striking a balance between task efficiency and human comfort and safety.
\paragraph{Core Functionality}
This layer utilizes belief-aware planners, such as MPC or MPPI control, to co-design robot actions, including footsteps and grasp forces~\cite{ponton2016convex, zhang2025mopformer, zhang2024early}.
\paragraph{Interface Contract (Action Policy)}
The process begins with the Intent Belief and Human Feasibility and Cost as inputs, which are used to generate an optimal action policy. 
This policy is then relayed to the Foundation layer, serving as a reference for execution while adhering to safety standards at all times. 
Additionally, this layer is responsible for managing shared autonomy, negotiating control, and ensuring that the robot's actions are understandable to its human partner.

\subsection{Pathways Towards Seamless pHHI: A Unified Approach}
The most effective collaboration in pHHI will emerge from combining the three pillars into a unified framework. 
While modular approaches offer advantages such as interpretability and reduced computational cost, they also suffer from coupling issues and a lack of mutual influence between components. 
To achieve seamless and efficient collaboration, a unified framework is needed that integrates these elements holistically. 
Such unifications could be directly or indirectly inspired by the methods presented in the three pillars. 
We propose dividing this effort into two major areas of focus: Intent-Aware Human Models and Human-Aware Humanoid Control.

\subsubsection{Intent-Aware Human Models}

There have been primarily two approaches to integrating human intent with human models.  
(a) Embedding intent within the human model: In this paradigm, intent is directly incorporated into the model of human dynamics, resulting in a unified framework that predicts human states based on knowledge of the underlying goals. 
This leads to more coherent predictions, as motion, force, and intent are generated from a single representation. 
(b) Treating intent as a separate module: Here, intent is estimated independently and then used to condition the human model, or alternatively, separate human models are constructed for different intents or task categories. 
This modular design offers flexibility, enabling the combination of different inference and modeling methods. 
However, it introduces challenges of \emph{temporal misalignment} (intent updates and biomechanical predictions may operate on different timescales) and \emph{coupling issues} (conflicts or inconsistencies can arise when the outputs of intent inference do not align with the physical predictions of the human model).
For example, a human may transiently generate forces inconsistent with their long-term intent, leading to contradictory signals across the two modules.  
A more unified approach, where intent and human modeling are integrated into a single predictive framework, would require richer datasets and more complex models, but would avoid these integration pitfalls. 
A high-fidelity model would be able to capture not only human intended goals and biomechanical aspects but also the interdependence between them. 
A key requirement, however, is that such models maintain both high fidelity and real-time implementability. 
Advancing both approaches, while maintaining real-time implementability and high-fidelity human representation, remains an important direction for future work.

\subsubsection{Human-Aware Humanoid Control}
Current literature on pHHI often treats the human either as a disturbance or as an oversimplified controller~\cite{KumbharArtemiadis2025MPCQP, stasse2009fast}. 
Such formulations fail to capture the full state and behavior of the human, which restricts the humanoid's ability to optimize for human efficiency or adapt to individual interaction patterns. 
This can lead to inefficient and less natural collaboration.
To address this gap, future work should focus on integrating intent-aware human models that strike a balance between fidelity and real-time applicability into humanoid control paradigms. 
Embedding a state–action model of the human directly within the control loop would enable human-aware control, allowing the humanoid to anticipate human responses to its actions. 
In turn, this would support more compliant, effective, and efficient collaborations. 
Moreover, optimization costs or reward functions could be shaped to explicitly account for human efficiency and comfort. 
These models could either embed intention directly within the model or rely on a separate intent-prediction module. 
Promising approaches in this direction include deep reinforcement learning (DRL) and model-based methods such as MPPI that incorporate learned models of the coupled human–humanoid system. 
Approaches analogous to combining terrain representations with locomotion policies in deep reinforcement learning \cite{gangapurwala2022rloc} could similarly be employed to integrate human models with humanoid control. 

\subsection{Pathways Towards Seamless pHHI: Hybrid Approach}
While fully modular pipelines provide simplicity and computational tractability, they suffer from coupling issues, temporal misalignment, and the inability to capture mutual influences between intent, human state, and control. 
Conversely, fully unified frameworks promise seamless integration by embedding intent, biomechanics, and control within a single predictive–optimization structure, but are often computationally prohibitive for real-time human–robot collaboration.

At the lower level, humanoid control is integrated with a lightweight human model that captures only fast-changing aspects of interaction. 
These include instantaneous interaction forces, short-horizon velocity intentions, and small postural corrections. 
The lightweight nature of the human model ensures that updates remain computationally feasible. 
This level is designed to react quickly to transient changes, prevent unsafe forces, and maintain smooth collaboration. 

At the higher level, a higher-fidelity human model is employed to estimate slow-changing intents and states. Examples include overall task goals, fatigue, strength asymmetries, ergonomic preferences, and comfort envelopes. 
The higher layer produces long-horizon intent beliefs and ergonomic summaries, which are passed down as references, constraints, and weights to the lower level for enforcement. 
In long-duration co-carrying tasks, fatigue can reduce a person’s ability to communicate intent through force. 
A low-level model might misinterpret these weaker or noisy signals as a change in intent, causing the robot to overreact. 
With a two-layer hybrid framework, the high-level model tracks fatigue and provides corrections, allowing the robot to recognize that reduced force is due to fatigue rather than new intent. 
This enables the robot to remain efficient and aligned with the task goal while reducing human strain. 

More generally, the framework could be extended to as many layers as needed, depending on how fast or slow different human states evolve. 
Such a hybrid framework retains the benefits of unification while remaining computationally and resource efficient, making it a practical pathway for robust human–robot collaboration in resource-constrained settings.
 
\subsection{Future Directions for pHHI}
A significant advancement in pHHI will arise from systems that can anticipate needs, personalize experiences, and generalize knowledge, all while ensuring safety and respecting human privacy. 
These systems should be able to reason about the consequences of their actions, adapt in real-time to unique partners, and transfer their skills seamlessly across different tasks and environments.

\subsubsection{Proactive and predictive interaction beyond intent labels}
Tomorrow's pHHI should not only anticipate human intentions but also predict the outcomes of a robot executing a proposed plan. 
This requires action-conditioned prediction models that integrate pose, gaze, and force data to forecast short-term human motion and forces under hypothetical robot actions. 
These models should provide both trajectories and uncertainty estimates to support belief-aware planners.
Two key capabilities are essential for this approach. 
First, consequence-aware forecasting helps distinguish whether a planned acceleration, footstep, or grip change will lead to assistance or resistance. 
Second, intentional versus accidental motion discrimination is necessary.
This involves utilizing multimodal signals, such as gaze fixation, pre-activation patterns, and sudden transitions in force to differentiate purposeful movements from slips, startle reflexes, and to modulate assistance based on these cues appropriately.

\subsubsection{Personalization and co-adaptation: physical and psychological}
A robot must be able to customize its assistance based on the individual's morphology, strength, stiffness, fatigue, and preferences. 
High-fidelity musculoskeletal and neuromechanical models provide essential variables, such as joint loads, endpoint stiffness ellipsoids, and momentum heuristics. 
However, these models need to be simplified into fast, person-specific surrogates that can be created through brief, non-intrusive calibration methods.
Online adaptation is necessary to update ergonomic parameters, such as reachable and comfortable postures and wrench-rate limits, as fatigue changes over time. 
Additionally, personalization has a psychological component; trust and comfort depend on a robot's legibility and predictability. 
Future systems should aim to encode user-defined preferences (including pace, assertiveness, and haptic firmness) along with implicit indicators (such as micro-withdrawals and grip variations) as constraints and costs. 
Furthermore, they should be capable of managing role exchange (i.e., deciding who leads or follows) when confidence in their intent or comfort diminishes.

\subsubsection{Privacy by design in contact-rich interaction}
The rich multimodal sensing required for advanced pHHI, including video, gaze tracking, and inertial wearables, raises legitimate concerns about privacy. 
A forward-looking approach must prioritize privacy by design, emphasizing data minimization and on-device inference. 
This includes favoring robot-side force and tactile sensing over high-fidelity imagery when in close contact, performing intent estimation at the edge, and ensuring that protocols are transparent, giving users explicit control over what data is stored and shared. 

\section{Conclusion}
\label{sec:conclusions}
This review has explored the current state of pHHI through the key pillars of modeling and control of humanoids, human intent estimation and prediction, and human models. 
While significant advancements have been made in each of these areas, the main obstacle to achieving seamless and robust interaction is the ongoing fragmentation among them. 
The challenges related to safety, predictability, and adaptation are not confined to a single pillar; rather, they are systemic issues that require an integrated solution.

We propose that the way forward involves creating unified architectural frameworks that closely connect predictive intent estimation, personalized human models, and certifiably safe whole-body control. 
By designing systems where these components inform and constrain one another in real-time, we can transition from reactive robots to proactive and efficient partners. 
Future research focused on this integration will be essential for developing humanoids that can anticipate human needs, provide personalized assistance, and generalize their skills, ultimately enabling them to become true collaborators in our daily lives.

\backmatter


\section*{Declarations}

\textbf{Author Contributions}
G.A.C. and S.S.K. performed the literature search and data analysis and drafted the paper. P.A. had the idea for the article and critically revised the work.

\noindent
\textbf{Funding}
This material is based upon work supported by the National Science
Foundation under Grants No. 2018905, 250047, 2415093. 

\noindent
\textbf{Competing Interests}
The authors have no conflicts of interest to declare that are relevant to the content of this article.

\noindent
\textbf{Ethical Approval} 
Not applicable



\bibliography{references}

@ARTICLE{Gangapurwala2020Guided,
  author={Gangapurwala, Siddhant and Mitchell, Alexander and Havoutis, Ioannis},
  journal={IEEE Robotics and Automation Letters}, 
  title={Guided Constrained Policy Optimization for Dynamic Quadrupedal Robot Locomotion}, 
  year={2020},
  volume={5},
  number={2},
  pages={3642-3649},
  doi={10.1109/LRA.2020.2979656}}

@article{gangapurwala2022rloc,
  title={Rloc: Terrain-aware legged locomotion using reinforcement learning and optimal control},
  author={Gangapurwala, Siddhant and Geisert, Mathieu and Orsolino, Romeo and Fallon, Maurice and Havoutis, Ioannis},
  journal={IEEE Transactions on Robotics},
  volume={38},
  number={5},
  pages={2908--2927},
  year={2022},
  publisher={IEEE}
}

@inproceedings{levine2013guided,
  title={Guided policy search},
  author={Levine, Sergey and Koltun, Vladlen},
  booktitle={International conference on machine learning},
  pages={1--9},
  year={2013},
  organization={PMLR}
}

@article{jiayi_2024,
  title={Agile and versatile bipedal robot tracking control through reinforcement learning},
  author={Li, Jiayi and Ye, Linqi and Cheng, Yi and Liu, Houde and Liang, Bin},
  journal={arXiv preprint arXiv:2404.08246},
  year={2024}
}

@INPROCEEDINGS{Siekmann2021Sim,
  author={Siekmann, Jonah and Godse, Yesh and Fern, Alan and Hurst, Jonathan},
  booktitle={2021 IEEE International Conference on Robotics and Automation (ICRA)}, 
  title={Sim-to-Real Learning of All Common Bipedal Gaits via Periodic Reward Composition}, 
  year={2021},
  volume={},
  number={},
  pages={7309-7315},
  doi={10.1109/ICRA48506.2021.9561814}}

@INPROCEEDINGS{Singh2022Learning,
  author={Singh, Rohan P. and Benallegue, Mehdi and Morisawa, Mitsuharu and Cisneros, Rafael and Kanehiro, Fumio},
  booktitle={2022 IEEE-RAS 21st International Conference on Humanoid Robots (Humanoids)}, 
  title={Learning Bipedal Walking On Planned Footsteps For Humanoid Robots}, 
  year={2022},
  volume={},
  number={},
  pages={686-693},
  doi={10.1109/Humanoids53995.2022.10000067}}

@inproceedings{Siekmann2020Learning,
  title={Learning Memory-Based Control for Human-Scale Bipedal Locomotion},
  author={Siekmann, Jonah and Valluri, Srikar and Dao, Jeremy and Bermillo, Lorenzo and Duan, Helei and Fern, Alan and Hurst, Jonathan},
  booktitle={Robotics science and systems},
  year={2020}
}

@INPROCEEDINGS{Li2021Reinforcement,
  author={Li, Zhongyu and Cheng, Xuxin and Peng, Xue Bin and Abbeel, Pieter and Levine, Sergey and Berseth, Glen and Sreenath, Koushil},
  booktitle={2021 IEEE International Conference on Robotics and Automation (ICRA)}, 
  title={Reinforcement Learning for Robust Parameterized Locomotion Control of Bipedal Robots}, 
  year={2021},
  volume={},
  number={},
  pages={2811-2817},
  doi={10.1109/ICRA48506.2021.9560769}}

@inproceedings{xie2020learning,
  title={Learning locomotion skills for cassie: Iterative design and sim-to-real},
  author={Xie, Zhaoming and Clary, Patrick and Dao, Jeremy and Morais, Pedro and Hurst, Jonanthan and Panne, Michiel},
  booktitle={Conference on Robot Learning},
  pages={317--329},
  year={2020},
  organization={PMLR}
}

@INPROCEEDINGS{Xie2018Feedback,
  author={Xie, Zhaoming and Berseth, Glen and Clary, Patrick and Hurst, Jonathan and van de Panne, Michiel},
  booktitle={2018 IEEE/RSJ International Conference on Intelligent Robots and Systems (IROS)}, 
  title={Feedback Control For Cassie With Deep Reinforcement Learning}, 
  year={2018},
  volume={},
  number={},
  pages={1241-1246},
  doi={10.1109/IROS.2018.8593722}}

@misc{bao2025deepreinforcementlearningbipedal,
      title={Deep Reinforcement Learning for Bipedal Locomotion: A Brief Survey}, 
      author={Lingfan Bao and Joseph Humphreys and Tianhu Peng and Chengxu Zhou},
      year={2025},
      eprint={2404.17070},
      archivePrefix={arXiv},
      primaryClass={cs.RO},
      url={https://arxiv.org/abs/2404.17070}, 
}

@article{tong2024advancements,
  title={Advancements in humanoid robots: A comprehensive review and future prospects},
  author={Tong, Yuchuang and Liu, Haotian and Zhang, Zhengtao},
  journal={IEEE/CAA Journal of Automatica Sinica},
  volume={11},
  number={2},
  pages={301--328},
  year={2024},
  publisher={IEEE}
}

@article{vianello2021human,
  title={Human-humanoid interaction and cooperation: a review},
  author={Vianello, Lorenzo and Penco, Luigi and Gomes, Waldez and You, Yang and Anzalone, Salvatore Maria and Maurice, Pauline and Thomas, Vincent and Ivaldi, Serena},
  journal={Current Robotics Reports},
  volume={2},
  number={4},
  pages={441--454},
  year={2021},
  publisher={Springer}
}

@article{agravante2019human,
  title={Human-humanoid collaborative carrying},
  author={Agravante, Don Joven and Cherubini, Andrea and Sherikov, Alexander and Wieber, Pierre-Brice and Kheddar, Abderrahmane},
  journal={IEEE Transactions on Robotics},
  volume={35},
  number={4},
  pages={833--846},
  year={2019},
  publisher={IEEE}
}

@inproceedings{agravante2016walking,
  title={Walking pattern generators designed for physical collaboration},
  author={Agravante, Don Joven and Sherikov, Alexander and Wieber, Pierre-Brice and Cherubini, Andrea and Kheddar, Abderrahmane},
  booktitle={2016 IEEE International conference on Robotics and Automation (ICRA)},
  pages={1573--1578},
  year={2016},
  organization={IEEE}
}

@inproceedings{bussy2012proactive,
  title={Proactive behavior of a humanoid robot in a haptic transportation task with a human partner},
  author={Bussy, Antoine and Gergondet, Pierre and Kheddar, Abderrahmane and Keith, Fran{\c{c}}ois and Crosnier, Andr{\'e}},
  booktitle={2012 IEEE RO-MAN: The 21st IEEE International Symposium on Robot and Human Interactive Communication},
  pages={962--967},
  year={2012},
  organization={IEEE}
}

@article{lefevre2024humanoid,
  title={Humanoid-human sit-to-stand-to-sit assistance},
  author={Lef{\`e}vre, Hugo and Chaki, Tomohiro and Kawakami, Tomohiro and Tanguy, Arnaud and Yoshiike, Takahide and Kheddar, Abderrahmane},
  journal={IEEE Robotics and Automation Letters},
  year={2024},
  publisher={IEEE}
}

@inproceedings{rapetti2023control,
  title={A control approach for human-robot ergonomic payload lifting},
  author={Rapetti, Lorenzo and Sartore, Carlotta and Elobaid, Mohamed and Tirupachuri, Yeshasvi and Draicchio, Francesco and Kawakami, Tomohiro and Yoshiike, Takahide and Pucci, Daniele},
  booktitle={2023 IEEE International Conference on Robotics and Automation (ICRA)},
  pages={7504--7510},
  year={2023},
  organization={IEEE}
}

@inproceedings{brecelj2023utilizing,
  title={Utilizing a Phase State System for Reliable Physical Assistance in Human-Humanoid Robot Collaboration},
  author={Brecelj, Tilen and Petri{\v{c}}, Tadej},
  booktitle={2023 21st International Conference on Advanced Robotics (ICAR)},
  pages={258--263},
  year={2023},
  organization={IEEE}
}

@inproceedings{otani2018generating,
  title={Generating assistive humanoid motions for co-manipulation tasks with a multi-robot quadratic program controller},
  author={Otani, Kazuya and Bouyarmane, Karim and Ivaldi, Serena},
  booktitle={2018 IEEE International Conference on Robotics and Automation (ICRA)},
  pages={3107--3113},
  year={2018},
  organization={IEEE}
}

@inproceedings{yang2022centaur,
  title={A centaur system for assisting human walking with load carriage},
  author={Yang, Ping and Yan, Haoyun and Yang, Bowen and Li, Jianquan and Li, Kailin and Leng, Yuquan and Fu, Chenglong},
  booktitle={2022 IEEE/RSJ International Conference on Intelligent Robots and Systems (IROS)},
  pages={5242--5248},
  year={2022},
  organization={IEEE}
}

@inproceedings{agravante2014collaborative,
  title={Collaborative human-humanoid carrying using vision and haptic sensing},
  author={Agravante, Don Joven and Cherubini, Andrea and Bussy, Antoine and Gergondet, Pierre and Kheddar, Abderrahmane},
  booktitle={2014 IEEE international conference on robotics and automation (ICRA)},
  pages={607--612},
  year={2014},
  organization={IEEE}
}

@inproceedings{stasse2009fast,
  title={Fast foot prints re-planning and motion generation during walking in physical human-humanoid interaction},
  author={Stasse, Olivier and Evrard, Paul and Perrin, Nicolas and Mansard, Nicolas and Kheddar, Abderrahmane},
  booktitle={2009 9th IEEE-RAS International Conference on Humanoid Robots},
  pages={284--289},
  year={2009},
  organization={IEEE}
}

@inproceedings{bussy2012human,
  title={Human-humanoid haptic joint object transportation case study},
  author={Bussy, Antoine and Kheddar, Abderrahmane and Crosnier, Andr{\'e} and Keith, Fran{\c{c}}ois},
  booktitle={2012 IEEE/RSJ International Conference on Intelligent Robots and Systems},
  pages={3633--3638},
  year={2012},
  organization={IEEE}
}

@inproceedings{rahem2022human,
  title={Human-humanoid robot cooperative load transportation: model-based control approach},
  author={Rahem, R{\'e}my and Wong, Christopher Yee and Suleiman, Wael},
  booktitle={2022 IEEE/RSJ International Conference on Intelligent Robots and Systems (IROS)},
  pages={8306--8312},
  year={2022},
  organization={IEEE}
}

@inproceedings{evrard2009teaching,
  title={Teaching physical collaborative tasks: object-lifting case study with a humanoid},
  author={Evrard, Paul and Gribovskaya, Elena and Calinon, Sylvain and Billard, Aude and Kheddar, Abderrahmane},
  booktitle={2009 9th IEEE-RAS International Conference on Humanoid Robots},
  pages={399--404},
  year={2009},
  organization={IEEE}
}

@inproceedings{lopez2014compliant,
  title={Compliant control of a humanoid robot helping a person stand up from a seated position},
  author={L{\'o}pez, Alfonso Montellano and Vaillant, Joris and Keith, Fran{\c{c}}ois and Fraisse, Philippe and Kheddar, Abderrahmane},
  booktitle={2014 IEEE-RAS International Conference on Humanoid Robots},
  pages={817--822},
  year={2014},
  organization={IEEE}
}

@inproceedings{maroger2022study,
  title={From the Study of Table Trajectories during Collaborative Carriages toward Pro-active Human-Robot Table Handling Tasks},
  author={Maroger, Isabelle and Stasse, Olivier and Watier, Bruno},
  booktitle={2022 IEEE-RAS 21st International Conference on Humanoid Robots (Humanoids)},
  pages={911--918},
  year={2022},
  organization={IEEE}
}

@inproceedings{evrard2009homotopy,
  title={Homotopy-based controller for physical human-robot interaction},
  author={Evrard, Paul and Kheddar, Abderrahmane},
  booktitle={RO-MAN 2009-The 18th IEEE International Symposium on Robot and Human Interactive Communication},
  pages={1--6},
  year={2009},
  organization={IEEE}
}

@article{romano2017codyco,
  title={The codyco project achievements and beyond: Toward human aware whole-body controllers for physical human robot interaction},
  author={Romano, Francesco and Nava, Gabriele and Azad, Morteza and {\v{C}}amernik, Jernej and Dafarra, Stefano and Dermy, Oriane and Latella, Claudia and Lazzaroni, Maria and Lober, Ryan and Lorenzini, Marta and others},
  journal={IEEE Robotics and Automation Letters},
  volume={3},
  number={1},
  pages={516--523},
  year={2017},
  publisher={IEEE}
}

@article{monje2011new,
  title={A new approach on human--robot collaboration with humanoid robot RH-2},
  author={Monje, CA and Pierro, Paolo and Balaguer, Carlos},
  journal={Robotica},
  volume={29},
  number={6},
  pages={949--957},
  year={2011},
  publisher={Cambridge University Press}
}

@inproceedings{ikemoto2009physical,
  title={Physical interaction learning: Behavior adaptation in cooperative human-robot tasks involving physical contact},
  author={Ikemoto, Shuhei and Amor, Heni Ben and Minato, Takashi and Ishiguro, Hiroshi and Jung, Bernhard},
  booktitle={RO-MAN 2009-The 18th IEEE International Symposium on Robot and Human Interactive Communication},
  pages={504--509},
  year={2009},
  organization={IEEE}
}

@inproceedings{stuckler2011following,
  title={Following human guidance to cooperatively carry a large object},
  author={St{\"u}ckler, J{\"o}rg and Behnke, Sven},
  booktitle={2011 11th IEEE-RAS International Conference on Humanoid Robots},
  pages={218--223},
  year={2011},
  organization={IEEE}
}

@inproceedings{lee2011physical,
  title={Physical human robot interaction in imitation learning},
  author={Lee, Dongheui and Ott, Christian and Nakamura, Yoshihiko and Hirzinger, Gerd},
  booktitle={2011 IEEE International Conference on Robotics and Automation},
  pages={3439--3440},
  year={2011},
  organization={IEEE}
}

@article{fang2023human,
  title={Human modeling in physical human-robot interaction: A brief survey},
  author={Fang, Cheng and Peternel, Luka and Seth, Ajay and Sartori, Massimo and Mombaur, Katja and Yoshida, Eiichi},
  journal={IEEE Robotics and Automation Letters},
  volume={8},
  number={9},
  pages={5799--5806},
  year={2023},
  publisher={IEEE}
}

@inproceedings{dominguez2024exploring,
  title={Exploring transformers and visual transformers for force prediction in human-robot collaborative transportation tasks},
  author={Dom{\'\i}nguez-Vidal, Jos{\'e} Enrique and Sanfeliu, Alberto},
  booktitle={2024 IEEE International Conference on Robotics and Automation (ICRA)},
  pages={3191--3197},
  year={2024},
  organization={IEEE}
}

@inproceedings{dominguez2024force,
  title={Force and Velocity Prediction in Human-Robot Collaborative Transportation Tasks through Video Retentive Networks},
  author={Dom{\'\i}nguez-Vidal, JE and Sanfeliu, Alberto},
  booktitle={2024 IEEE/RSJ International Conference on Intelligent Robots and Systems (IROS)},
  pages={9307--9313},
  year={2024},
  organization={IEEE}
}

@article{lanini2018human,
  title={Human intention detection as a multiclass classification problem: Application in physical human--robot interaction while walking},
  author={Lanini, Jessica and Razavi, Hamed and Urain, Julen and Ijspeert, Auke},
  journal={IEEE Robotics and Automation Letters},
  volume={3},
  number={4},
  pages={4171--4178},
  year={2018},
  publisher={IEEE}
}

@article{liu2024human,
  title={Human-robot collaboration through a multi-scale graph convolution neural network with temporal attention},
  author={Liu, Zhaowei and Lu, Xilang and Liu, Wenzhe and Qi, Wen and Su, Hang},
  journal={IEEE Robotics and Automation Letters},
  volume={9},
  number={3},
  pages={2248--2255},
  year={2024},
  publisher={IEEE}
}

@article{gao2021hybrid,
  title={Hybrid recurrent neural network architecture-based intention recognition for human--robot collaboration},
  author={Gao, Xiaoshan and Yan, Liang and Wang, Gang and Gerada, Chris},
  journal={IEEE Transactions on Cybernetics},
  volume={53},
  number={3},
  pages={1578--1586},
  year={2021},
  publisher={IEEE}
}

@article{zhang2024early,
  title={Early prediction of human intention for human--robot collaboration using transformer network},
  author={Zhang, Xinyao and Tian, Sibo and Liang, Xiao and Zheng, Minghui and Behdad, Sara},
  journal={Journal of Computing and Information Science in Engineering},
  volume={24},
  number={5},
  year={2024},
  publisher={American Society of Mechanical Engineers Digital Collection}
}

@article{callens2020framework,
  title={A framework for recognition and prediction of human motions in human-robot collaboration using probabilistic motion models},
  author={Callens, Thomas and Van der Have, Tuur and Van Rossom, Sam and De Schutter, Joris and Aertbeli{\"e}n, Erwin},
  journal={IEEE Robotics and Automation Letters},
  volume={5},
  number={4},
  pages={5151--5158},
  year={2020},
  publisher={IEEE}
}

@inproceedings{li2024large,
  title={Large Language Model for Humanoid Cognition in Proactive Human-Robot Collaboration},
  author={Li, Shufei and Wang, Zuoxu and Yan, Zhijie and Gao, Yiping and Jiang, Han and Zheng, Pai},
  booktitle={2024 IEEE 20th International Conference on Automation Science and Engineering (CASE)},
  pages={540--545},
  year={2024},
  organization={IEEE}
}

@inproceedings{dermy2018prediction,
  title={Prediction of human whole-body movements with ae-promps},
  author={Dermy, Oriane and Chaveroche, Maxime and Colas, Francis and Charpillet, Fran{\c{c}}ois and Ivaldi, Serena},
  booktitle={2018 IEEE-RAS 18th International Conference on Humanoid Robots (Humanoids)},
  pages={572--579},
  year={2018},
  organization={IEEE}
}

@article{hoffman2023inferring,
  title={Inferring human intent and predicting human action in human--robot collaboration},
  author={Hoffman, Guy and Bhattacharjee, Tapomayukh and Nikolaidis, Stefanos},
  journal={Annual Review of Control, Robotics, and Autonomous Systems},
  volume={7},
  year={2023},
  publisher={Annual Reviews}
}

@article{wong2023vision,
  author    = {Wong, C. Y. and Vergez, L. and Suleiman, W.},
  title     = {Vision-and Tactile-Based Continuous Multimodal Intention and Attention Recognition for Safer Physical Human–Robot Interaction},
  journal   = {IEEE Transactions on Automation Science and Engineering},
  year      = {2023},
  doi       = {10.1109/TASE.2023.3248496}
}

@article{townsend2017estimating,
  title={Estimating human intent for physical human-robot co-manipulation},
  author={Townsend, Eric C and Mielke, Erich A and Wingate, David and Killpack, Marc D},
  journal={arXiv preprint arXiv:1705.10851},
  year={2017}
}

@article{shao2024constraint,
  title={Constraint-Aware Intent Estimation for Dynamic Human-Robot Object Co-Manipulation},
  author={Shao, Yifei Simon and Li, Tianyu and Keyvanian, Shafagh and Chaudhari, Pratik and Kumar, Vijay and Figueroa, Nadia},
  journal={arXiv preprint arXiv:2409.00215},
  year={2024}
}

@article{yang2025implicit,
  title={Implicit Communication in Human-Robot Collaborative Transport},
  author={Yang, Elvin and Mavrogiannis, Christoforos},
  journal={arXiv preprint arXiv:2502.03346},
  year={2025}
}

@article{erden2010human,
  title={Human-intent detection and physically interactive control of a robot without force sensors},
  author={Erden, Mustafa Suphi and Tomiyama, Tetsuo},
  journal={IEEE Transactions on Robotics},
  volume={26},
  number={2},
  pages={370--382},
  year={2010},
  publisher={IEEE}
}

@article{losey2018review,
  title={A review of intent detection, arbitration, and communication aspects of shared control for physical human--robot interaction},
  author={Losey, Dylan P and McDonald, Craig G and Battaglia, Edoardo and O'Malley, Marcia K},
  journal={Applied Mechanics Reviews},
  volume={70},
  number={1},
  pages={010804},
  year={2018},
  publisher={American Society of Mechanical Engineers}
}

@article{lanini2017interactive,
  title={Interactive locomotion: Investigation and modeling of physically-paired humans while walking},
  author={Lanini, Jessica and Duburcq, Alexis and Razavi, Hamed and Le Goff, Camille G and Ijspeert, Auke Jan},
  journal={PLoS One},
  volume={12},
  number={9},
  pages={e0179989},
  year={2017},
  publisher={Public Library of Science San Francisco, CA USA}
}

@article{li2013human,
  title={Human--robot collaboration based on motion intention estimation},
  author={Li, Yanan and Ge, Shuzhi Sam},
  journal={IEEE/ASME Transactions on Mechatronics},
  volume={19},
  number={3},
  pages={1007--1014},
  year={2013},
  publisher={IEEE}
}

@article{wang2013probabilistic,
  title={Probabilistic movement modeling for intention inference in human--robot interaction},
  author={Wang, Zhikun and M{\"u}lling, Katharina and Deisenroth, Marc Peter and Ben Amor, Heni and Vogt, David and Sch{\"o}lkopf, Bernhard and Peters, Jan},
  journal={The International Journal of Robotics Research},
  volume={32},
  number={7},
  pages={841--858},
  year={2013},
  publisher={SAGE Publications Sage UK: London, England}
}

@article{seth2018opensim,
  title={OpenSim: Simulating musculoskeletal dynamics and neuromuscular control to study human and animal movement},
  author={Seth, Ajay and Hicks, Jennifer L and Uchida, Thomas K and Habib, Ayman and Dembia, Christopher L and Dunne, James J and Ong, Carmichael F and DeMers, Matthew S and Rajagopal, Apoorva and Millard, Matthew and others},
  journal={PLoS computational biology},
  volume={14},
  number={7},
  pages={e1006223},
  year={2018},
  publisher={Public Library of Science San Francisco, CA USA}
}

@article{hyon2007full,
  title={Full-body compliant human--humanoid interaction: balancing in the presence of unknown external forces},
  author={Hyon, Sang-Ho and Hale, Joshua G and Cheng, Gordon},
  journal={IEEE transactions on robotics},
  volume={23},
  number={5},
  pages={884--898},
  year={2007},
  publisher={IEEE}
}

@article{liu2021computational,
  title={Computational modeling: Human dynamic model},
  author={Liu, Lijia and Cooper, Joseph L and Ballard, Dana H},
  journal={Frontiers in Neurorobotics},
  volume={15},
  pages={723428},
  year={2021},
  publisher={Frontiers Media SA}
}

@article{liu2021humans,
  title={Humans use minimum cost movements in a whole-body task},
  author={Liu, Lijia and Ballard, Dana},
  journal={Scientific Reports},
  volume={11},
  number={1},
  pages={20081},
  year={2021},
  publisher={Nature Publishing Group UK London}
}

@article{hiatt2017human,
  title={Human modeling for human--robot collaboration},
  author={Hiatt, Laura M and Narber, Cody and Bekele, Esube and Khemlani, Sangeet S and Trafton, J Gregory},
  journal={The International Journal of Robotics Research},
  volume={36},
  number={5-7},
  pages={580--596},
  year={2017},
  publisher={SAGE Publications Sage UK: London, England}
}

@incollection{maurice2019digital,
  TITLE = {{Digital Human Modeling for Collaborative Robotics}},
  AUTHOR = {Maurice, Pauline and Padois, Vincent and Measson, Yvan and Bidaud, Philippe},
  URL = {https://hal.science/hal-02389726},
  BOOKTITLE = {{DHM and Posturography}},
  YEAR = {2019},
  MONTH = Aug,
  HAL_ID = {hal-02389726},
  HAL_VERSION = {v1},
}

@article{magistris2015dynamic,
  title={Dynamic digital human models for ergonomic analysis based on humanoid robotics techniques},
  author={Magistris, Giovanni De and Micaelli, Alain and Savin, Jonathan and Gaudez, Clarisse and Marsot, Jacques},
  journal={International Journal of the Digital Human},
  volume={1},
  number={1},
  pages={81--109},
  year={2015},
  publisher={Inderscience Publishers}
}

@article{tevet2022human,
  title={Human motion diffusion model},
  author={Tevet, Guy and Raab, Sigal and Gordon, Brian and Shafir, Yonatan and Cohen-Or, Daniel and Bermano, Amit H},
  journal={arXiv preprint arXiv:2209.14916},
  year={2022}
}

@inproceedings{bowden2000learning,
  title={Learning statistical models of human motion},
  author={Bowden, Richard},
  booktitle={IEEE Workshop on Human Modeling, Analysis and Synthesis, CVPR},
  volume={2000},
  year={2000}
}

@inproceedings{rempe2021humor,
  title={Humor: 3{D} human motion model for robust pose estimation},
  author={Rempe, Davis and Birdal, Tolga and Hertzmann, Aaron and Yang, Jimei and Sridhar, Srinath and Guibas, Leonidas J},
  booktitle={Proceedings of the IEEE/CVF international conference on computer vision},
  pages={11488--11499},
  year={2021}
}

@inproceedings{kajita2003resolved,
  title={Resolved momentum control: Humanoid motion planning based on the linear and angular momentum},
  author={Kajita, Shuuji and Kanehiro, Fumio and Kaneko, Kenji and Fujiwara, Kiyoshi and Harada, Kensuke and Yokoi, Kazuhito and Hirukawa, Hirohisa},
  booktitle={Proceedings 2003 ieee/rsj international conference on intelligent robots and systems (iros 2003)(cat. no. 03ch37453)},
  volume={2},
  pages={1644--1650},
  year={2003},
  organization={IEEE}
}

@inproceedings{wieber2006trajectory,
  title={Trajectory free linear model predictive control for stable walking in the presence of strong perturbations},
  author={Wieber, Pierre-Brice},
  booktitle={2006 6th IEEE-RAS International Conference on Humanoid Robots},
  pages={137--142},
  year={2006},
  organization={IEEE}
}

@article{westervelt2003hybrid,
  title={Hybrid zero dynamics of planar biped walkers},
  author={Westervelt, Eric R and Grizzle, Jessy W and Koditschek, Daniel E},
  journal={IEEE transactions on automatic control},
  volume={48},
  number={1},
  pages={42--56},
  year={2003},
  publisher={IEEE}
}

@book{westervelt2018feedback,
  title={Feedback control of dynamic bipedal robot locomotion},
  author={Westervelt, Eric R and Grizzle, Jessy W and Chevallereau, Christine and Choi, Jun Ho and Morris, Benjamin},
  year={2018},
  address = {Boca Raton},
  publisher={CRC press}
}

@article{grizzle2014models,
  title={Models, feedback control, and open problems of 3{D} bipedal robotic walking},
  author={Grizzle, Jessy W and Chevallereau, Christine and Sinnet, Ryan W and Ames, Aaron D},
  journal={Automatica},
  volume={50},
  number={8},
  pages={1955--1988},
  year={2014},
  publisher={Elsevier}
}

@inproceedings{hsu2015control,
  title={Control barrier function based quadratic programs with application to bipedal robotic walking},
  author={Hsu, Shao-Chen and Xu, Xiangru and Ames, Aaron D},
  booktitle={2015 American Control Conference (ACC)},
  pages={4542--4548},
  year={2015},
  organization={IEEE}
}

@article{radosavovic2023learning,
  title={Learning humanoid locomotion with transformers},
  author={Radosavovic, Ilija and Xiao, Tete and Zhang, Bike and Darrell, Trevor and Malik, Jitendra and Sreenath, Koushil},
  journal={CoRR},
  year={2023}
}

@inproceedings{lai2023sim,
  title={Sim-to-real transfer for quadrupedal locomotion via terrain transformer},
  author={Lai, Hang and Zhang, Weinan and He, Xialin and Yu, Chen and Tian, Zheng and Yu, Yong and Wang, Jun},
  booktitle={2023 IEEE International Conference on Robotics and Automation (ICRA)},
  pages={5141--5147},
  year={2023},
  organization={IEEE}
}

@article{alvarez2024real,
  title={Real-Time Whole-Body Control of Legged Robots with Model-Predictive Path Integral Control},
  author={Alvarez-Padilla, Juan and Zhang, John Z and Kwok, Sofia and Dolan, John M and Manchester, Zachary},
  journal={arXiv preprint arXiv:2409.10469},
  year={2024}
}

@article{dugar2024learning,
  title={Learning multi-modal whole-body control for real-world humanoid robots},
  author={Dugar, Pranay and Shrestha, Aayam and Yu, Fangzhou and van Marum, Bart and Fern, Alan},
  journal={arXiv preprint arXiv:2408.07295},
  year={2024}
}

@article{orin2013centroidal,
  title={Centroidal dynamics of a humanoid robot},
  author={Orin, David E and Goswami, Ambarish and Lee, Sung-Hee},
  journal={Autonomous robots},
  volume={35},
  pages={161--176},
  year={2013},
  publisher={Springer}
}

@article{kuindersma2016optimization,
  title={Optimization-based locomotion planning, estimation, and control design for the atlas humanoid robot},
  author={Kuindersma, Scott and Deits, Robin and Fallon, Maurice and Valenzuela, Andr{\'e}s and Dai, Hongkai and Permenter, Frank and Koolen, Twan and Marion, Pat and Tedrake, Russ},
  journal={Autonomous robots},
  volume={40},
  pages={429--455},
  year={2016},
  publisher={Springer}
}

@article{koolen2012capturability,
  title={Capturability-based analysis and control of legged locomotion, Part 1: Theory and application to three simple gait models},
  author={Koolen, Twan and De Boer, Tomas and Rebula, John and Goswami, Ambarish and Pratt, Jerry},
  journal={The international journal of robotics research},
  volume={31},
  number={9},
  pages={1094--1113},
  year={2012},
  publisher={SAGE Publications Sage UK: London, England}
}

@article{fahmi2019passive,
  title={Passive whole-body control for quadruped robots: Experimental validation over challenging terrain},
  author={Fahmi, Shamel and Mastalli, Carlos and Focchi, Michele and Semini, Claudio},
  journal={IEEE Robotics and Automation Letters},
  volume={4},
  number={3},
  pages={2553--2560},
  year={2019},
  publisher={IEEE}
}

@article{herr2008angular,
  title={Angular momentum in human walking},
  author={Herr, Hugh and Popovic, Marko},
  journal={Journal of experimental biology},
  volume={211},
  number={4},
  pages={467--481},
  year={2008},
  publisher={Company of Biologists}
}

@inproceedings{kajita1991study,
  title={Study of dynamic biped locomotion on rugged terrain-derivation and application of the linear inverted pendulum mode},
  author={Kajita, Shuuji and Tani, Kazuo},
  booktitle={Proceedings. 1991 IEEE International Conference on Robotics and Automation},
  pages={1405--1406},
  year={1991},
  organization={IEEE Computer Society}
}

@inproceedings{englsberger2017smooth,
  title={Smooth trajectory generation and push-recovery based on divergent component of motion},
  author={Englsberger, Johannes and Mesesan, George and Ott, Christian},
  booktitle={2017 IEEE/RSJ International Conference on Intelligent Robots and Systems (IROS)},
  pages={4560--4567},
  year={2017},
  organization={IEEE}
}

@inproceedings{englsberger2011bipedal,
  title={Bipedal walking control based on capture point dynamics},
  author={Englsberger, Johannes and Ott, Christian and Roa, M{\'a}ximo A and Albu-Sch{\"a}ffer, Alin and Hirzinger, Gerhard},
  booktitle={2011 IEEE/RSJ international conference on intelligent robots and systems},
  pages={4420--4427},
  year={2011},
  organization={IEEE}
}

@article{caron2018capturabilitybased,
  title={Capturability-based analysis, optimization and control of 3d bipedal walking},
  author={Caron, St{\'e}phane and Escande, Adrien and Lanari, Leonardo and Mallein, Bastien},
  journal={arXiv preprint arXiv:1801.07022},
  year={2018}
}

@inproceedings{caron2018balance,
  title={Balance control using both {ZMP} and {COM} height variations: A convex boundedness approach},
  author={Caron, St{\'e}phane and Mallein, Bastien},
  booktitle={2018 IEEE International Conference on Robotics and Automation (ICRA)},
  pages={1779--1784},
  year={2018},
  organization={IEEE}
}

@inproceedings{hirukawa2006universal,
  title={A universal stability criterion of the foot contact of legged robots-adios {ZMP}},
  author={Hirukawa, Hirohisa and Hattori, Shizuko and Harada, Kensuke and Kajita, Shuuji and Kaneko, Kenji and Kanehiro, Fumio and Fujiwara, Kiyoshi and Morisawa, Mitsuharu},
  booktitle={Proceedings 2006 IEEE International Conference on Robotics and Automation, 2006. ICRA 2006.},
  pages={1976--1983},
  year={2006},
  organization={IEEE}
}

@inproceedings{dai2014whole,
  title={Whole-body motion planning with centroidal dynamics and full kinematics},
  author={Dai, Hongkai and Valenzuela, Andr{\'e}s and Tedrake, Russ},
  booktitle={2014 IEEE-RAS International Conference on Humanoid Robots},
  pages={295--302},
  year={2014},
  organization={IEEE}
}

@inproceedings{ponton2016convex,
  title={A convex model of humanoid momentum dynamics for multi-contact motion generation},
  author={Ponton, Brahayam and Herzog, Alexander and Schaal, Stefan and Righetti, Ludovic},
  booktitle={2016 IEEE-RAS 16th International Conference on Humanoid Robots (Humanoids)},
  pages={842--849},
  year={2016},
  organization={IEEE}
}

@misc{neumann2025humanoid,
  author       = {Neumann, Clas},
  title        = {Humanoid robots offer both disruption and promise. Here’s why},
  year         = {2025},
  month        = {June},
  day          = {16},
  url          = {https://www.weforum.org/stories/2025/06/humanoid-robots-offer-disruption-and-promise/},
  organization = {World Economic Forum},
  note         = {Accessed: 2025-08-09}
}

@inproceedings{sakagami2002intelligent,
  title={The intelligent {ASIMO}: System overview and integration},
  author={Sakagami, Yoshiaki and Watanabe, Ryujin and Aoyama, Chiaki and Matsunaga, Shinichi and Higaki, Nobuo and Fujimura, Kikuo},
  booktitle={IEEE/RSJ international conference on intelligent robots and systems},
  volume={3},
  pages={2478--2483},
  year={2002},
  organization={IEEE}
}

@article{guizzo2019leaps,
  title={By leaps and bounds: An exclusive look at how boston dynamics is redefining robot agility},
  author={Guizzo, Erico},
  journal={IEEE Spectrum},
  volume={56},
  number={12},
  pages={34--39},
  year={2019},
  publisher={IEEE}
}

@misc{unitreeG1,
  author       = {{Unitree Robotics}},
  title        = {G1 Humanoid Robot},
  year         = {n.d.},
  url          = {https://www.unitree.com/g1},
  note         = {Accessed: 2025-08-09}
}

@misc{agilityroboticsHomepage,
  author       = {{Agility Robotics}},
  title        = {Agility Robotics},
  year         = {n.d.},
  url          = {https://www.agilityrobotics.com/},
  note         = {Accessed: 2025-08-09}
}

@inproceedings{mukherjee2022humanoid,
  title={Humanoid robot in healthcare: a systematic review and future research directions},
  author={Mukherjee, Subhodeep and Baral, Manish Mohan and Pal, Surya Kant and Chittipaka, Venkataiah and Roy, Rita and Alam, Khursheed},
  booktitle={2022 International conference on machine learning, big data, cloud and parallel computing (COM-IT-CON)},
  volume={1},
  pages={822--826},
  year={2022},
  organization={IEEE}
}

@inproceedings{mcginn2014towards,
  title={Towards the design of a new humanoid robot for domestic applications},
  author={McGinn, Conor and Cullinan, Michael and Holland, Donal and Kelly, Kevin},
  booktitle={2014 IEEE International Conference on Technologies for Practical Robot Applications (TePRA)},
  pages={1--6},
  year={2014},
  organization={IEEE}
}

@inproceedings{settimi2014modular,
  title={A modular approach for remote operation of humanoid robots in search and rescue scenarios},
  author={Settimi, Alessandro and Pavan, Corrado and Varricchio, Valerio and Ferrati, Mirko and Mingo Hoffman, Enrico and Rocchi, Alessio and Melo, Kamilo and Tsagarakis, Nikos G and Bicchi, Antonio},
  booktitle={International Workshop on Modelling and simulation for autonomous systems},
  pages={192--205},
  year={2014},
  organization={Springer}
}

@inproceedings{diftler2011robonaut,
  title={Robonaut 2-the first humanoid robot in space},
  author={Diftler, Myron A and Mehling, Joshua S and Abdallah, Muhammad E and Radford, Nicolaus A and Bridgwater, Lyndon B and Sanders, Adam M and Askew, Roger Scott and Linn, D Marty and Yamokoski, John D and Permenter, FA and others},
  booktitle={2011 IEEE international conference on robotics and automation},
  pages={2178--2183},
  year={2011},
  organization={IEEE}
}

@article{mikolajczyk2022recent,
  title={Recent advances in bipedal walking robots: Review of gait, drive, sensors and control systems},
  author={Mikolajczyk, Tadeusz and Miko{\l}ajewska, Emilia and Al-Shuka, Hayder FN and Malinowski, Tomasz and K{\l}odowski, Adam and Pimenov, Danil Yurievich and Paczkowski, Tomasz and Hu, Fuwen and Giasin, Khaled and Miko{\l}ajewski, Dariusz and others},
  journal={Sensors},
  volume={22},
  number={12},
  pages={4440},
  year={2022},
  publisher={MDPI}
}

@article{zhang2025review,
  title={A Review of Fall Coping Strategies for Humanoid Robots},
  author={Zhang, Haoyan and Wu, Jiaqi and Fan, Jiarong and An, Yang and Jin, Xingze and Cui, Da and Yang, YiRu},
  journal={Journal of Bionic Engineering},
  volume={22},
  number={2},
  pages={480--512},
  year={2025},
  publisher={Springer}
}

@article{laschi2000grasping,
  title={Grasping and manipulation in humanoid robotics},
  author={Laschi, Cecilia and Dario, Paolo and Carrozza, Maria Chiara and Guglielmelli, Eugenio and Teti, Giancarlo and Taddeucci, Davide and Leoni, Fabio and Massa, Bruno and Zecca, Massimiliano and Lazzarini, Roberto and others},
  journal={Scuola Superiore Sant Anna, Italia},
  year={2000}
}

@article{farajtabar2024path,
  title={The path towards contact-based physical human--robot interaction},
  author={Farajtabar, Mohammad and Charbonneau, Marie},
  journal={Robotics and Autonomous Systems},
  volume={182},
  pages={104829},
  year={2024},
  publisher={Elsevier}
}

@article{darvish2023teleoperation,
  title={Teleoperation of humanoid robots: A survey},
  author={Darvish, Kourosh and Penco, Luigi and Ramos, Joao and Cisneros, Rafael and Pratt, Jerry and Yoshida, Eiichi and Ivaldi, Serena and Pucci, Daniele},
  journal={IEEE Transactions on Robotics},
  volume={39},
  number={3},
  pages={1706--1727},
  year={2023},
  publisher={IEEE}
}

@article{li2024variable,
  title={Variable stiffness methods for robots: A review},
  author={Li, Zhang and Chu, Xiaoyu and Hu, Xinye and Zhang, Zhiyi and Li, Nanpei and Li, Junfeng},
  journal={Smart Materials and Structures},
  volume={33},
  number={6},
  pages={063002},
  year={2024},
  publisher={IOP Publishing}
}

@article{song2019tutorial,
  title={A tutorial survey and comparison of impedance control on robotic manipulation},
  author={Song, Peng and Yu, Yueqing and Zhang, Xuping},
  journal={Robotica},
  volume={37},
  number={5},
  pages={801--836},
  year={2019},
  publisher={Cambridge University Press}
}

@INPROCEEDINGS{Kumbhar2025Finite,
  author={Kumbhar, Shubham S. and Kulkarni, Abhijeet M. and Poulakakis, Ioannis},
  booktitle={2025 IEEE International Conference on Robotics and Automation (ICRA)}, 
  title={Finite-Step Capturability and Recursive Feasibility for Bipedal Walking in Constrained Regions}, 
  year={2025},
  volume={},
  number={},
  pages={397-403},
  doi={10.1109/ICRA55743.2025.11128831}}

@inproceedings{KumbharArtemiadis2025MPCQP,
  author={Kumbhar, Shubham S. and Artemiadis, Panagiotis},
  booktitle={2025 IEEE International Conference on Robotics and Automation (ICRA)}, 
  title={{MPC-QP}-Based Control Framework for Compliant Behavior of Humanoid Robots in Physical Collaboration with Humans}, 
  year={2025},
  volume={},
  number={},
  pages={14714-14720},
  doi={10.1109/ICRA55743.2025.11128037}}

@inproceedings{martini2024robust,
  title={A robust filter for marker-less multi-person tracking in human-robot interaction scenarios},
  author={Martini, Enrico and Parekh, Harshil and Peng, Shaoting and Bombieri, Nicola and Figueroa, Nadia},
  booktitle={2024 33rd IEEE International Conference on Robot and Human Interactive Communication (ROMAN)},
  pages={424--429},
  year={2024},
  organization={IEEE}
}

@article{xiao2023unified,
  title={Unified human-scene interaction via prompted chain-of-contacts},
  author={Xiao, Zeqi and Wang, Tai and Wang, Jingbo and Cao, Jinkun and Zhang, Wenwei and Dai, Bo and Lin, Dahua and Pang, Jiangmiao},
  journal={arXiv preprint arXiv:2309.07918},
  year={2023}
}

@inproceedings{dao2024sim,
  title={Sim-to-real learning for humanoid box loco-manipulation},
  author={Dao, Jeremy and Duan, Helei and Fern, Alan},
  booktitle={2024 IEEE International Conference on Robotics and Automation (ICRA)},
  pages={16930--16936},
  year={2024},
  organization={IEEE}
}

@inproceedings{paredes2022resolved,
  title={Resolved motion control for 3d underactuated bipedal walking using linear inverted pendulum dynamics and neural adaptation},
  author={Paredes, Victor C and Hereid, Ayonga},
  booktitle={2022 IEEE/RSJ International Conference on Intelligent Robots and Systems (IROS)},
  pages={6761--6767},
  year={2022},
  organization={IEEE}
}

@article{wang2023physhoi,
  title={Physhoi: Physics-based imitation of dynamic human-object interaction},
  author={Wang, Yinhuai and Lin, Jing and Zeng, Ailing and Luo, Zhengyi and Zhang, Jian and Zhang, Lei},
  journal={arXiv preprint arXiv:2312.04393},
  year={2023}
}

@inproceedings{hassan2023synthesizing,
  title={Synthesizing physical character-scene interactions},
  author={Hassan, Mohamed and Guo, Yunrong and Wang, Tingwu and Black, Michael and Fidler, Sanja and Peng, Xue Bin},
  booktitle={ACM SIGGRAPH 2023 Conference Proceedings},
  pages={1--9},
  year={2023}
}

@article{xie2023hierarchical,
  title={Hierarchical planning and control for box loco-manipulation},
  author={Xie, Zhaoming and Tseng, Jonathan and Starke, Sebastian and van de Panne, Michiel and Liu, C Karen},
  journal={Proceedings of the ACM on Computer Graphics and Interactive Techniques},
  volume={6},
  number={3},
  pages={1--18},
  year={2023},
  publisher={ACM New York, NY, USA}
}

@article{yu2021human,
  title={Human dynamics from monocular video with dynamic camera movements},
  author={Yu, Ri and Park, Hwangpil and Lee, Jehee},
  journal={ACM Transactions on Graphics (TOG)},
  volume={40},
  number={6},
  pages={1--14},
  year={2021},
  publisher={ACM New York, NY, USA}
}

@article{peng2018deepmimic,
  title={Deepmimic: Example-guided deep reinforcement learning of physics-based character skills},
  author={Peng, Xue Bin and Abbeel, Pieter and Levine, Sergey and Van de Panne, Michiel},
  journal={ACM Transactions On Graphics (TOG)},
  volume={37},
  number={4},
  pages={1--14},
  year={2018},
  publisher={ACM New York, NY, USA}
}

@article{cheng2024expressive,
  title={Expressive whole-body control for humanoid robots},
  author={Cheng, Xuxin and Ji, Yandong and Chen, Junming and Yang, Ruihan and Yang, Ge and Wang, Xiaolong},
  journal={arXiv preprint arXiv:2402.16796},
  year={2024}
}

@inproceedings{wu2024infer,
  title={Infer and adapt: Bipedal locomotion reward learning from demonstrations via inverse reinforcement learning},
  author={Wu, Feiyang and Gu, Zhaoyuan and Wu, Hanran and Wu, Anqi and Zhao, Ye},
  booktitle={2024 IEEE International Conference on Robotics and Automation (ICRA)},
  pages={16243--16250},
  year={2024},
  organization={IEEE}
}

@inproceedings{schwarke2023curiosity,
  title={Curiosity-driven learning of joint locomotion and manipulation tasks},
  author={Schwarke, Clemens and Klemm, Victor and Van der Boon, Matthijs and Bjelonic, Marko and Hutter, Marco},
  booktitle={Proceedings of the 7th Conference on Robot Learning},
  volume={229},
  pages={2594--2610},
  year={2023},
  organization={PMLR}
}

@inproceedings{sartore2024automatic,
  title={Automatic gain tuning for humanoid robots walking architectures using gradient-free optimization techniques},
  author={Sartore, Carlotta and Rando, Marco and Romualdi, Giulio and Molinari, Cesare and Rosasco, Lorenzo and Pucci, Daniele},
  booktitle={2024 IEEE-RAS 23rd International Conference on Humanoid Robots (Humanoids)},
  pages={996--1003},
  year={2024},
  organization={IEEE}
}

@article{kumar2023words,
  title={Words into action: Learning diverse humanoid robot behaviors using language guided iterative motion refinement},
  author={Kumar, K Niranjan and Essa, Irfan and Ha, Sehoon},
  journal={arXiv preprint arXiv:2310.06226},
  year={2023}
}

@article{jiang2024harmon,
  title={Harmon: Whole-body motion generation of humanoid robots from language descriptions},
  author={Jiang, Zhenyu and Xie, Yuqi and Li, Jinhan and Yuan, Ye and Zhu, Yifeng and Zhu, Yuke},
  journal={arXiv preprint arXiv:2410.12773},
  year={2024}
}

@article{yao2024anybipe,
  title={Anybipe: An end-to-end framework for training and deploying bipedal robots guided by large language models},
  author={Yao, Yifei and He, Wentao and Gu, Chenyu and Du, Jiaheng and Tan, Fuwei and Zhu, Zhen and Lu, Junguo},
  journal={arXiv preprint arXiv:2409.08904},
  year={2024}
}

@article{he2024omnih2o,
  title={Omnih2o: Universal and dexterous human-to-humanoid whole-body teleoperation and learning},
  author={He, Tairan and Luo, Zhengyi and He, Xialin and Xiao, Wenli and Zhang, Chong and Zhang, Weinan and Kitani, Kris and Liu, Changliu and Shi, Guanya},
  journal={arXiv preprint arXiv:2406.08858},
  year={2024}
}

@inproceedings{sun2024prompt,
  title={Prompt, plan, perform: {LLM}-based humanoid control via quantized imitation learning},
  author={Sun, Jingkai and Zhang, Qiang and Duan, Yiqun and Jiang, Xiaoyang and Cheng, Chong and Xu, Renjing},
  booktitle={2024 IEEE International Conference on Robotics and Automation (ICRA)},
  pages={16236--16242},
  year={2024},
  organization={IEEE}
}

@article{ji2024exbody2,
  title={Exbody2: Advanced expressive humanoid whole-body control},
  author={Ji, Mazeyu and Peng, Xuanbin and Liu, Fangchen and Li, Jialong and Yang, Ge and Cheng, Xuxin and Wang, Xiaolong},
  journal={arXiv preprint arXiv:2412.13196},
  year={2024}
}

@article{murooka2021humanoid,
  title={Humanoid loco-manipulations pattern generation and stabilization control},
  author={Murooka, Masaki and Chappellet, Kevin and Tanguy, Arnaud and Benallegue, Mehdi and Kumagai, Iori and Morisawa, Mitsuharu and Kanehiro, Fumio and Kheddar, Abderrahmane},
  journal={IEEE Robotics and Automation Letters},
  volume={6},
  number={3},
  pages={5597--5604},
  year={2021},
  publisher={IEEE}
}

@inproceedings{audren2014model,
  title={Model preview control in multi-contact motion-application to a humanoid robot},
  author={Audren, Herv{\'e} and Vaillant, Joris and Kheddar, Abderrahmane and Escande, Adrien and Kaneko, Kenji and Yoshida, Eiichi},
  booktitle={2014 IEEE/RSJ International Conference on Intelligent Robots and Systems},
  pages={4030--4035},
  year={2014},
  organization={IEEE}
}

@inproceedings{nozawa2010full,
  title={A full-body motion control method for a humanoid robot based on on-line estimation of the operational force of an object with an unknown weight},
  author={Nozawa, Shunichi and Ueda, Ryohei and Kakiuchi, Youhei and Okada, Kei and Inaba, Masayuki},
  booktitle={2010 IEEE/RSJ International Conference on Intelligent Robots and Systems},
  pages={2684--2691},
  year={2010},
  organization={IEEE}
}

@article{harada2007real,
  title={Real-time planning of humanoid robot's gait for force-controlled manipulation},
  author={Harada, Kensuke and Kajita, Shuuji and Kanehiro, Fumio and Fujiwara, Kiyoshi and Kaneko, Kenji and Yokoi, Kazuhito and Hirukawa, Hirohisa},
  journal={IEEE/ASME Transactions on Mechatronics},
  volume={12},
  number={1},
  pages={53--62},
  year={2007},
  publisher={IEEE}
}

@article{liu2024opt2skill,
  title={Opt2skill: Imitating dynamically-feasible whole-body trajectories for versatile humanoid loco-manipulation},
  author={Liu, Fukang and Gu, Zhaoyuan and Cai, Yilin and Zhou, Ziyi and Jung, Hyunyoung and Jang, Jaehwi and Zhao, Shijie and Ha, Sehoon and Chen, Yue and Xu, Danfei and others},
  journal={arXiv preprint arXiv:2409.20514},
  year={2024}
}

@article{fu2024humanplus,
  title={Humanplus: Humanoid shadowing and imitation from humans},
  author={Fu, Zipeng and Zhao, Qingqing and Wu, Qi and Wetzstein, Gordon and Finn, Chelsea},
  journal={arXiv preprint arXiv:2406.10454},
  year={2024}
}

@article{shahbazi2016unified,
  title={Unified modeling and control of walking and running on the spring-loaded inverted pendulum},
  author={Shahbazi, Mohammad and Babu{\v{s}}ka, Robert and Lopes, Gabriel AD},
  journal={IEEE Transactions on Robotics},
  volume={32},
  number={5},
  pages={1178--1195},
  year={2016},
  publisher={IEEE}
}

@incollection{geyer2019gait,
  title={Gait based on the spring-loaded inverted pendulum},
  author={Geyer, Hartmut and Saranli, Uluc},
  booktitle={Humanoid Robotics: A Reference},
  pages={923--947},
  year={2019},
  address="Dordrecht",
  publisher={Springer}
}

@article{robey2024jailbreaking,
  title={Jailbreaking {LLM}-controlled robots},
  author={Robey, Alexander and Ravichandran, Zachary and Kumar, Vijay and Hassani, Hamed and Pappas, George J},
  journal={arXiv preprint arXiv:2410.13691},
  year={2024}
}

@article{zhang2024badrobot,
  title={Badrobot: Jailbreaking {LLM}-based embodied {AI} in the physical world},
  author={Zhang, Hangtao and Zhu, Chenyu and Wang, Xianlong and Zhou, Ziqi and Hu, Shengshan and Zhang, Leo Yu},
  journal={arXiv preprint arXiv:2407.20242},
  volume={3},
  year={2024}
}

@article{zhang2023integrating,
  title={Integrating intention-based systems in human-robot interaction: {A} scoping review of sensors, algorithms, and trust},
  author={Zhang, Yifei and Doyle, Thomas},
  journal={Frontiers in Robotics and AI},
  volume={10},
  pages={1233328},
  year={2023},
  publisher={Frontiers Media SA}
}

@article{keemink2018admittance,
  title={Admittance control for physical human--robot interaction},
  author={Keemink, Arvid QL and Van der Kooij, Herman and Stienen, Arno HA},
  journal={The International Journal of Robotics Research},
  volume={37},
  number={11},
  pages={1421--1444},
  year={2018},
  publisher={SAGE Publications Sage UK: London, England}
}

@article{su2023recent,
  title={Recent advancements in multimodal human--robot interaction},
  author={Su, Hang and Qi, Wen and Chen, Jiahao and Yang, Chenguang and Sandoval, Juan and Laribi, Med Amine},
  journal={Frontiers in Neurorobotics},
  volume={17},
  pages={1084000},
  year={2023},
  publisher={Frontiers Media SA}
}

@inproceedings{martinez2017simple,
  title={A simple yet effective baseline for 3{D} human pose estimation},
  author={Martinez, Julieta and Hossain, Rayat and Romero, Javier and Little, James J},
  booktitle={Proceedings of the IEEE international conference on computer vision},
  pages={2640--2649},
  year={2017}
}

@article{lyu20223d,
  title={3{D} human motion prediction: A survey},
  author={Lyu, Kedi and Chen, Haipeng and Liu, Zhenguang and Zhang, Beiqi and Wang, Ruili},
  journal={Neurocomputing},
  volume={489},
  pages={345--365},
  year={2022},
  publisher={Elsevier}
}

@article{usman2022skeleton,
  title={Skeleton-based motion prediction: A survey},
  author={Usman, Muhammad and Zhong, Jianqi},
  journal={Frontiers in Computational Neuroscience},
  volume={16},
  pages={1051222},
  year={2022},
  publisher={Frontiers Media SA}
}

@inproceedings{belcamino2024gaze,
  title={Gaze-based intention recognition for human-robot collaboration},
  author={Belcamino, Valerio and Takase, Miwa and Kilina, Mariya and Carf{\`\i}, Alessandro and Mastrogiovanni, Fulvio and Shimada, Akira and Shimizu, Sota},
  booktitle={Proceedings of the 2024 International Conference on Advanced Visual Interfaces},
  pages={1--5},
  year={2024}
}

@article{monari2024physical,
  title={Physical ergonomics monitoring in human--robot collaboration: A standard-based approach for hand-guiding applications},
  author={Monari, Eugenio and Avallone, Giulia and Valori, Marcello and Agostini, Lorenzo and Chen, Yi and Palazzi, Emanuele and Vertechy, Rocco},
  journal={Machines},
  volume={12},
  number={4},
  pages={231},
  year={2024},
  publisher={MDPI}
}

@article{qiu2022multi,
  title={Multi-sensor information fusion based on machine learning for real applications in human activity recognition: State-of-the-art and research challenges},
  author={Qiu, Sen and Zhao, Hongkai and Jiang, Nan and Wang, Zhelong and Liu, Long and An, Yi and Zhao, Hongyu and Miao, Xin and Liu, Ruichen and Fortino, Giancarlo},
  journal={Information Fusion},
  volume={80},
  pages={241--265},
  year={2022},
  publisher={Elsevier}
}

@article{wang2024multimodal,
  title={Multimodal human--robot interaction for human-centric smart manufacturing: a survey},
  author={Wang, Tian and Zheng, Pai and Li, Shufei and Wang, Lihui},
  journal={Advanced Intelligent Systems},
  volume={6},
  number={3},
  pages={2300359},
  year={2024},
  publisher={Wiley Online Library}
}

@article{strabala2013toward,
  title={Toward seamless human-robot handovers},
  author={Strabala, Kyle and Lee, Min Kyung and Dragan, Anca and Forlizzi, Jodi and Srinivasa, Siddhartha S and Cakmak, Maya and Micelli, Vincenzo},
  journal={Journal of Human-Robot Interaction},
  volume={2},
  number={1},
  pages={112--132},
  year={2013},
  publisher={Journal of Human-Robot Interaction Steering Committee}
}

@inproceedings{admoni2014deliberate,
  title={Deliberate delays during robot-to-human handovers improve compliance with gaze communication},
  author={Admoni, Henny and Dragan, Anca and Srinivasa, Siddhartha S and Scassellati, Brian},
  booktitle={Proceedings of the 2014 ACM/IEEE international conference on Human-robot interaction},
  pages={49--56},
  year={2014}
}

@article{koppula2015anticipating,
  title={Anticipating human activities using object affordances for reactive robotic response},
  author={Koppula, Hema S and Saxena, Ashutosh},
  journal={IEEE transactions on pattern analysis and machine intelligence},
  volume={38},
  number={1},
  pages={14--29},
  year={2015},
  publisher={IEEE}
}

@inproceedings{mainprice2013human,
  title={Human-robot collaborative manipulation planning using early prediction of human motion},
  author={Mainprice, Jim and Berenson, Dmitry},
  booktitle={2013 IEEE/RSJ International Conference on Intelligent Robots and Systems},
  pages={299--306},
  year={2013},
  organization={IEEE}
}

@article{al2021improving,
  title={Improving human robot collaboration through Force/Torque based learning for object manipulation},
  author={Al-Yacoub, Ali and Zhao, YC and Eaton, William and Goh, Yee Mey and Lohse, Niels},
  journal={Robotics and Computer-Integrated Manufacturing},
  volume={69},
  pages={102111},
  year={2021},
  publisher={Elsevier}
}

@article{maeda2017probabilistic,
  title={Probabilistic movement primitives for coordination of multiple human--robot collaborative tasks},
  author={Maeda, Guilherme J and Neumann, Gerhard and Ewerton, Marco and Lioutikov, Rudolf and Kroemer, Oliver and Peters, Jan},
  journal={Autonomous Robots},
  volume={41},
  number={3},
  pages={593--612},
  year={2017},
  publisher={Springer}
}

@article{losey2022physical,
  title={Physical interaction as communication: Learning robot objectives online from human corrections},
  author={Losey, Dylan P and Bajcsy, Andrea and O’Malley, Marcia K and Dragan, Anca D},
  journal={The International Journal of Robotics Research},
  volume={41},
  number={1},
  pages={20--44},
  year={2022},
  publisher={SAGE Publications Sage UK: London, England}
}

@article{zhang2025mopformer,
  title={MoPFormer: Motion-Primitive Transformer for Wearable-Sensor Activity Recognition},
  author={Zhang, Hao and Zhuang, Zhan and Wang, Xuehao and Yang, Xiaodong and Zhang, Yu},
  journal={arXiv preprint arXiv:2505.20744},
  year={2025}
}

@article{huang2023conditional,
  title={Conditional predictive behavior planning with inverse reinforcement learning for human-like autonomous driving},
  author={Huang, Zhiyu and Liu, Haochen and Wu, Jingda and Lv, Chen},
  journal={IEEE Transactions on Intelligent Transportation Systems},
  volume={24},
  number={7},
  pages={7244--7258},
  year={2023},
  publisher={IEEE}
}

@article{liu2021robot,
  title={Robot recognizing humans intention and interacting with humans based on a multi-task model combining {ST-GCN-LSTM} model and {YOLO} model},
  author={Liu, Chunfang and Li, Xiaoli and Li, Qing and Xue, Yaxin and Liu, Huijun and Gao, Yize},
  journal={Neurocomputing},
  volume={430},
  pages={174--184},
  year={2021},
  publisher={Elsevier}
}

@article{geyer2010muscle,
  title={A muscle-reflex model that encodes principles of legged mechanics produces human walking dynamics and muscle activities},
  author={Geyer, Hartmut and Herr, Hugh},
  journal={IEEE Transactions on neural systems and rehabilitation engineering},
  volume={18},
  number={3},
  pages={263--273},
  year={2010},
  publisher={IEEE}
}

@article{rajagopal2016full,
  title={Full-body musculoskeletal model for muscle-driven simulation of human gait},
  author={Rajagopal, Apoorva and Dembia, Christopher L and DeMers, Matthew S and Delp, Denny D and Hicks, Jennifer L and Delp, Scott L},
  journal={IEEE transactions on biomedical engineering},
  volume={63},
  number={10},
  pages={2068--2079},
  year={2016},
  publisher={IEEE}
}

@article{mugge2010rigorous,
  title={A rigorous model of reflex function indicates that position and force feedback are flexibly tuned to position and force tasks},
  author={Mugge, Winfred and Abbink, David A and Schouten, Alfred C and Dewald, Julius PA and Van Der Helm, Frans CT},
  journal={Experimental brain research},
  volume={200},
  number={3},
  pages={325--340},
  year={2010},
  publisher={Springer}
}

@article{burdet2000method,
  title={A method for measuring endpoint stiffness during multi-joint arm movements},
  author={Burdet, Etienne and Osu, R and Franklin, DW and Yoshioka, T and Milner, TE and Kawato, M},
  journal={Journal of biomechanics},
  volume={33},
  number={12},
  pages={1705--1709},
  year={2000},
  publisher={Elsevier}
}

@article{ajoudani2018reduced,
  title={Reduced-complexity representation of the human arm active endpoint stiffness for supervisory control of remote manipulation},
  author={Ajoudani, Arash and Fang, Cheng and Tsagarakis, Nikos and Bicchi, Antonio},
  journal={The International Journal of Robotics Research},
  volume={37},
  number={1},
  pages={155--167},
  year={2018},
  publisher={SAGE Publications Sage UK: London, England}
}

@article{prendergast2021biomechanics,
  title={Biomechanics aware collaborative robot system for delivery of safe physical therapy in shoulder rehabilitation},
  author={Prendergast, J Micah and Balvert, Stephan and Driessen, Tom and Seth, Ajay and Peternel, Luka},
  journal={IEEE Robotics and Automation Letters},
  volume={6},
  number={4},
  pages={7177--7184},
  year={2021},
  publisher={IEEE}
}

@article{bethala2025h2,
  title={{H2-COMPACT}: Human-Humanoid Co-Manipulation via Adaptive Contact Trajectory Policies},
  author={Bethala, Geeta Chandra Raju and Huang, Hao and Pudasaini, Niraj and Ali, Abdullah Mohamed and Yuan, Shuaihang and Wen, Congcong and Tzes, Anthony and Fang, Yi},
  journal={arXiv preprint arXiv:2505.17627},
  year={2025}
}

@inproceedings{sentis2005freefloating,
  author    = {Luis Sentis and Oussama Khatib},
  title     = {Control of free-floating humanoid robots through task prioritization},
  booktitle = {Proceedings of the IEEE International Conference on Robotics and Automation (ICRA)},
  year      = {2005}
}

@inproceedings{sentis2006wholebody,
  author    = {Luis Sentis and Oussama Khatib},
  title     = {A whole-body control framework for humanoids operating in human environments},
  booktitle = {Proceedings of the IEEE International Conference on Robotics and Automation (ICRA)},
  year      = {2006}
}

@article{park2008multiplecontact,
  author    = {Jaeheung Park and Oussama Khatib},
  title     = {Robot multiple contact control},
  journal   = {Robotica},
  year      = {2008},
  volume    = {26},
  number    = {5},
  pages     = {661--673}
}

@article{sentis2010compliant,
  author    = {Luis Sentis},
  title     = {Compliant control of multicontact and center-of-mass behaviors in humanoid robots},
  journal   = {IEEE Transactions on Robotics},
  year      = {2010}
}

@inproceedings{dietrich2011dynamic,
  author    = {Alexander Dietrich and Thomas Wimb{\"o}ck and Alin Albu-Sch{\"a}ffer},
  title     = {Dynamic whole-body mobile manipulation with a torque-controlled humanoid robot via impedance control laws},
  booktitle = {Proceedings of the IEEE/RSJ International Conference on Intelligent Robots and Systems (IROS)},
  year      = {2011}
}

@article{dietrich2012reactive,
  author    = {Alexander Dietrich and Thomas Wimb{\"o}ck and Alin Albu-Sch{\"a}ffer and Gerd Hirzinger},
  title     = {Reactive whole-body control: Dynamic mobile manipulation using a large number of degrees of freedom},
  journal   = {IEEE Transactions on Robotics},
  year      = {2012}
}

@article{khatib1987operational,
  author    = {Oussama Khatib},
  title     = {A unified approach for motion and force control of robot manipulators: The operational space formulation},
  journal   = {IEEE Journal of Robotics and Automation},
  year      = {1987},
  volume    = {3},
  number    = {1},
  pages     = {43--53}
}

@article{herdt2010online,
  title={Online walking motion generation with automatic footstep placement},
  author={Herdt, Andrei and Diedam, Holger and Wieber, Pierre-Brice and Dimitrov, Dimitar and Mombaur, Katja and Diehl, Moritz},
  journal={Advanced Robotics},
  volume={24},
  number={5-6},
  pages={719--737},
  year={2010},
  publisher={Taylor \& Francis}
}

@article{geyer2006compliant,
  title={Compliant leg behaviour explains basic dynamics of walking and running},
  author={Geyer, Hartmut and Seyfarth, Andre and Blickhan, Reinhard},
  journal={Proceedings of the Royal Society B: Biological Sciences},
  volume={273},
  number={1603},
  pages={2861--2867},
  year={2006},
  publisher={The Royal Society London}
}

@article{xiong2020dynamic,
  title={Dynamic and versatile humanoid walking via embedding 3d actuated slip model with hybrid lip based stepping},
  author={Xiong, Xiaobin and Ames, Aaron D},
  journal={IEEE Robotics and Automation Letters},
  volume={5},
  number={4},
  pages={6286--6293},
  year={2020},
  publisher={IEEE}
}

@inproceedings{diedam2008online,
  title={Online walking gait generation with adaptive foot positioning through linear model predictive control},
  author={Diedam, Holger and Dimitrov, Dimitar and Wieber, Pierre-Brice and Mombaur, Katja and Diehl, Moritz},
  booktitle={2008 IEEE/RSJ international conference on intelligent robots and systems},
  pages={1121--1126},
  year={2008},
  organization={IEEE}
}

@article{Khatib1987,
  author  = {Oussama Khatib},
  title   = {A Unified Approach for Motion and Force Control of Robot Manipulators: The Operational Space Formulation},
  journal = {IEEE Journal of Robotics and Automation},
  year    = {1987},
  volume  = {3},
  number  = {1},
  pages   = {43--53},
  doi     = {10.1109/JRA.1987.1087068}
}

@inproceedings{baerlocher1998,
  title={Task-priority formulations for the kinematic control of highly redundant articulated structures},
  author={Baerlocher, Paolo and Boulic, Ronan},
  booktitle={Proceedings. 1998 IEEE/RSJ International Conference on Intelligent Robots and Systems. Innovations in Theory, Practice and Applications (Cat. No. 98CH36190)},
  volume={1},
  pages={323--329},
  year={1998},
  organization={IEEE}
}

@article{Mansard2007,
  author  = {Nicolas Mansard and Fran{\c c}ois Chaumette},
  title   = {Task Sequencing for High-Level Sensor-Based Control},
  journal = {IEEE Transactions on Robotics},
  year    = {2007},
  volume  = {23},
  number  = {1},
  pages   = {60--72},
  doi     = {10.1109/TRO.2006.889487}
}

@inproceedings{SentisKhatib2006,
  author    = {Luis Sentis and Oussama Khatib},
  title     = {A Whole-Body Control Framework for Humanoids Operating in Human Environments},
  booktitle = {Proceedings 2006 IEEE International Conference on Robotics and Automation, 2006. ICRA 2006.},
  year      = {2006},
  address   = {Orlando, FL, USA}
}

@article{Escande2014,
  author  = {Adrien Escande and Nicolas Mansard and Pierre-Brice Wieber},
  title   = {Hierarchical Quadratic Programming: Fast Online Humanoid-Robot Motion Generation},
  journal = {The International Journal of Robotics Research},
  year    = {2014},
  volume  = {33},
  number  = {7},
  pages   = {1006--1028},
  doi     = {10.1177/0278364914521306}
}

@article{Saab2013,
  author  = {Layale Saab and Oscar E. Ramos and Fran{\c c}ois Keith and Nicolas Mansard and Philippe Sou{\`e}res and Jean-Yves Fourquet},
  title   = {Dynamic Whole-Body Motion Generation Under Rigid Contacts and Other Unilateral Constraints},
  journal = {IEEE Transactions on Robotics},
  year    = {2013},
  volume  = {29},
  number  = {2},
  pages   = {346--362},
  doi     = {10.1109/TRO.2012.2234351}
}

@article{Herzog2016,
  author  = {Alexander Herzog and Nicholas Rotella and Sean Mason and Felix Grimminger and Stefan Schaal and Ludovic Righetti},
  title   = {Momentum Control with Hierarchical Inverse Dynamics on a Torque-Controlled Humanoid},
  journal = {Autonomous Robots},
  year    = {2016},
  volume  = {40},
  number  = {3},
  pages   = {473--491},
  doi     = {10.1007/s10514-015-9476-6}
}

@inproceedings{Tassa2012,
  author    = {Yuval Tassa and Tom Erez and Emanuel Todorov},
  title     = {Synthesis and Stabilization of Complex Behaviors Through Online Trajectory Optimization},
  booktitle = {IEEE/RSJ International Conference on Intelligent Robots and Systems (IROS)},
  year      = {2012},
  pages     = {4906--4913},
  doi       = {10.1109/IROS.2012.6386025}
}

@inproceedings{Tassa2014,
  author    = {Yuval Tassa and Nicolas Mansard and Emanuel Todorov},
  title     = {Control-Limited Differential Dynamic Programming},
  booktitle = {IEEE International Conference on Robotics and Automation (ICRA)},
  year      = {2014},
  pages     = {1168--1175},
  doi       = {10.1109/ICRA.2014.6907001}
}

@inproceedings{Budhiraja2018,
  author    = {Rohan Budhiraja and Justin Carpentier and Carlos Mastalli and Nicolas Mansard},
  title     = {Differential Dynamic Programming for Multi-Phase Rigid Contact Dynamics},
  booktitle = {IEEE-RAS International Conference on Humanoid Robots (Humanoids)},
  year      = {2018},
  pages     = {1--8},
  doi       = {10.1109/HUMANOIDS.2018.8624925}
}

@article{LiWensing2020,
  author  = {He Li and Patrick M. Wensing},
  title   = {Hybrid Systems Differential Dynamic Programming for Whole-Body Motion Planning of Legged Robots},
  journal = {IEEE Robotics and Automation Letters},
  year    = {2020},
  volume  = {5},
  number  = {3},
  pages   = {4556--4563},
  doi     = {10.1109/LRA.2020.3007475}
}

@article{xue2024full,
  title={Full-order sampling-based {MPC} for torque-level locomotion control via diffusion-style annealing},
  author={Xue, Haoru and Pan, Chaoyi and Yi, Zeji and Qu, Guannan and Shi, Guanya},
  journal={arXiv preprint arXiv:2409.15610},
  year={2024}
}

@article{patel2014effect,
  title={On the effect of muscular cocontraction on the 3-{D} human arm impedance},
  author={Patel, Harshil and O’Neill, Gerald and Artemiadis, Panagiotis},
  journal={IEEE Transactions on Biomedical Engineering},
  volume={61},
  number={10},
  pages={2602--2608},
  year={2014},
  publisher={IEEE}
}

@article{taborri2016gait,
  title={Gait partitioning methods: A systematic review},
  author={Taborri, Juri and Palermo, Eduardo and Rossi, Stefano and Cappa, Paolo},
  journal={Sensors},
  volume={16},
  number={1},
  pages={66},
  year={2016},
  publisher={MDPI}
}

@article{xue2021using,
  title={Using probabilistic movement primitives in analyzing human motion differences under transcranial current stimulation},
  author={Xue, Honghu and Herzog, Rebecca and Berger, Till M and B{\"a}umer, Tobias and Weissbach, Anne and Rueckert, Elmar},
  journal={Frontiers in Robotics and AI},
  volume={8},
  pages={721890},
  year={2021},
  publisher={Frontiers Media SA}
}

@inproceedings{xiang2024socialcvae,
  title={Socialcvae: Predicting pedestrian trajectory via interaction conditioned latents},
  author={Xiang, Wei and Haoteng, YIN and Wang, He and Jin, Xiaogang},
  booktitle={Proceedings of the AAAI Conference on Artificial Intelligence},
  volume={38},
  pages={6216--6224},
  year={2024}
}

@inproceedings{barsoum2018hp,
  title={Hp-gan: Probabilistic 3{D} human motion prediction via gan},
  author={Barsoum, Emad and Kender, John and Liu, Zicheng},
  booktitle={Proceedings of the IEEE conference on computer vision and pattern recognition workshops},
  pages={1418--1427},
  year={2018}
}

@article{ficht2021bipedal,
  title={Bipedal humanoid hardware design: A technology review},
  author={Ficht, Grzegorz and Behnke, Sven},
  journal={Current Robotics Reports},
  volume={2},
  number={2},
  pages={201--210},
  year={2021},
  publisher={Springer}
}

\end{document}